\documentclass{ecai}  

\usepackage{graphicx}
\usepackage{latexsym}

\usepackage{amsmath}
\usepackage{amsfonts}
\usepackage{subcaption}
\usepackage{derivative}

\usepackage{url}
\usepackage[hidelinks]{hyperref}
\usepackage{amssymb}

\renewcommand{\>}{{\rightarrow}}

\newcommand{\argmax}{\operatorname{argmax}}

\newtheorem{lemma}[theorem]{Lemma}

\newtheorem{assumption}{Assumption} 

\newcommand{\R}{{\mathbb R}}

\newcommand{\E}{{\mathbf E}}

\newcommand{\1}{{\mathbf 1}}

\newcommand{\bX}{{\mathbf X}}

\newcommand{\ba}{{\mathbf a}}

\newcommand{\e}{{\mathbf e}}
\newcommand{\g}{{\mathbf g}}

\newcommand{\x}{{\mathbf x}}

\newcommand{\z}{{\mathbf z}}

\newcommand{\bbs}{{\boldsymbol s}}
\newcommand{\bbw}{{\boldsymbol w}}
\newcommand{\bbu}{{\boldsymbol u}}

\begin{document}

\begin{frontmatter}

\title{On the Learning Dynamics of Attention Networks}

\author[1]{\fnms{Rahul}~\snm{Vashisht}\thanks{Corresponding Author, email: rahul@cse.iitm.ac.in, accepted work at 26th European Conference on Artificial Intelligence ECAI 2023, September 30 to 5th October 2023  }}
\author[1]{\fnms{Harish G.}~\snm{Ramaswamy}}

\address[1]{Department of CSE, Indian Institute of Technology Madras}

\begin{abstract}
Attention models are typically learned by optimizing one of three standard loss functions that are variously called -- soft attention, hard attention, and latent variable marginal likelihood (LVML) attention. 
All three paradigms are motivated by the same goal of finding two models-- a `focus' model that `selects' the right \textit{segment} of the input and a `classification' model that processes the selected segment into the target label.  
However, they differ significantly in the way the selected segments are aggregated, resulting in distinct dynamics and final results. 
We observe a unique signature of models learned using these paradigms and explain this as a consequence of the evolution of the classification model under gradient descent when the focus model is fixed. 
We also analyze these paradigms in a simple setting and derive closed-form expressions for the parameter trajectory under gradient flow.
With the soft attention loss, the focus model improves quickly at initialization and splutters later on. 
On the other hand, hard attention loss behaves in the opposite fashion.  
Based on our observations, we propose a simple hybrid approach that combines the advantages of the different loss functions and demonstrates it on a collection of semi-synthetic and real-world datasets.
\end{abstract}

\end{frontmatter}

\section{Introduction}
Attention models have emerged as one of the most successful architectures in deep learning \cite{Bahdanau2016,edunov-etal-2018-classical,Attnall17,pmlr-v37-xuc15}. These models offer a natural way to interpret the intermediate outputs by introducing an \textit{attention vector}, which identifies the relevant part of the input responsible for the output. Along with improving the model's performance, attention models also provide a transparent mechanism to study intermediate outputs in neural networks . As a result, attention models have become an essential tool for explainability in downstream tasks, thereby emphasizing the need for a comprehensive understanding of their working mechanisms. 

Recent advances in the field have focused on softly simulating alignments, also known as \textit{soft attention}, where the model uses a convex combination of features based on attention weights calculated using a deterministic function. Soft attention allows the model to focus on multiple segments of the input for downstream tasks, providing better accuracy than traditional neural networks. In contrast, \textit{hard attention} based approaches select one of the input segments based on the attention weight distribution \cite{10.5555/972470.972474,10.5555/2969033.2969073,pmlr-v37-xuc15}. Typically, a hard attention model is trained either by directly maximizing the log-marginal-likelihood or maximizing a lower bound objective obtained using Jensen's inequality. Both these methods are computationally expensive compared to the soft attention model.

\textbf{Contributions:} In this paper, we give insights into the learning dynamics of attention mechanisms under soft and hard attention paradigms. We study the dynamics of attention mechanisms under a ``fixed focus setting'' and identify a distinct property of soft and hard attention that makes them act very differently at different points in the training. This property makes the final learned model through soft attention less interpretable and makes the training with hard attention particularly slow at initialization.  We propose a hybrid approach that addresses some of these limitations. We also derive closed-form expressions for the parameter trajectory under gradient flow in a simple setting. Our work sheds light on the failure modes of different attention paradigms and enables the design of new algorithms with desirable properties.


\textbf{Related Work:} There has been a lot of research in the field of deep learning to understand the behavior of attention models. Translation tasks between sentence pairs have used hard attention \cite{10.5555/92858.92860,10.5555/972470.972474}, while soft attention has been used in other recent applications \cite{ChorowskiBCB14,machine_trns}. Several studies have investigated whether attention mechanisms can offer meaningful insights using empirical methods \cite{Jain2019,wiegreffe-pinter-2019-attention,Vashishth2019AttentionIA}. Even with these studies, a thorough understanding of attention models is still lacking in most research. Some research has focused on understanding attention weights in classification tasks and natural language settings such as visual question answering. In addition, other studies have discovered a mathematical relationship between attention scores and word embedding norms in topic classification \cite{lu2021on}. There have been works that show latent variable attention performs better than soft attention. These works also propose variational inference based attention models \cite{NEURIPS2018_b691334c}. Similar to this, there have been works focused on performing exact hard attention for the monotonic and non-monotonic sequence to sequence character-level transduction tasks \cite{wu-cotterell-2019-exact,wu-etal-2018-hard}. In this paper, we study the learning dynamics of various attention paradigms and explore how they produce different results.

\section{Losses and inference methods for attention}
In this section, we briefly summarise the three known paradigms/loss functions for attention in the context of a simple problem that we term selective dependence classification \cite{pmlr-v189-pandey23a} (SDC). We also analyze the performance of the three paradigms on a semi-synthetic dataset based on CIFAR10 \cite{Krizhevsky09learningmultiple} and identify some characteristic signatures of these paradigms. 
\subsection{Attention and Latent variable alignment (LVA)}

In the latent variable model for attention, we consider an instance $\bX$ that is a mosaic object consisting of multiple segments(or parts or patches), each of which is represented by a fixed dimensional vector. The label variable $y \in \mathcal{Y}$ is generated based on the mosaic instance $\bX$ and a latent variable $\z$. The latent variable $\z$ (often called `alignment'  \cite{NEURIPS2018_b691334c} in literature) indicates which segment (or segments) of $\bX$ generates $y$. Directly maximizing the $\log p (y\lvert \bX;\theta)$ (log marginal likelihood) is complicated in general, and becomes even more intractable in cases where the alignment variable $\z$ can take a large number of values \cite{wu-cotterell-2019-exact,wu-etal-2018-hard}.

\subsection{Selective Dependence Classification}
For the sake of studying the three standard paradigms of attention, we consider the following concrete version of the latent variable alignment (LVA) problem mentioned above -- we call this a selective dependence classification(SDC) problem. Here the instance $\bX = [\x_1, \x_2, \ldots, \x_m] \in \R^{d \times m}$, contains $m$ parts or segments each of which is represented by a vector in $\R^d$. The label $Y$ takes values in $\mathcal Y=\{1,2,\ldots, C\}$. The hidden alignment variable $Z$  takes values in $\{1,2,\ldots,m\}$, indicating that only one of the segments of $\bX$ is responsible for the generation of $y$. The instance $\bX$ is called a mosaic instance, and the segment identified by $Z$ is called a `foreground' segment, while the rest are called `background' segments. The latent variable alignment probabilistic model is given as follows:
\begin{align*}
    P(Y=k|\bX,Z=i) &= \sigma_k \left([\g^*_1(\x_i), \ldots, \g^*_C(\x_i)]\right) \\
    P(Z= i| \bX) &= \sigma_i \left( [f^*(\x_1), \ldots, f^*(\x_m)] \right) 
\end{align*}
where $f^*:\R^d \> \R$ and $\g^*:\R^d \> \R^C$ are the parameters of the data model defined above and $\sigma$ is the softmax operator that transforms an arbitrary vector into probability vector of the same dimension, and $\sigma_j$ represents the $j^{th}$ co-ordinate.

The training data for the task is the collection of pairs $\bX, Y$. The final goal in the SDC problem is simply to learn a model that predicts the label $Y$ correctly from a mosaic instance $\bX$. This toy problem is analogous to an image classification problem where each image is labeled only based on an object occupying only a small (and unknown) portion of the image. The LVA model is a discriminative model and only gives $P(Z,Y|\bX)$. In our synthetic and theoretical arguments, we consider a corresponding full generative model by giving $P(Y,Z)$ (which is simply uniform over $[C]\times [m]$) and $P(\bX|Y,Z)$.  Conditioned on $Z,Y$, the $Z^\text{th}$ segment of $\bX$, denoted by $\x_Z$ is distributed as $D_Y$, a distribution over $\R^d$ which we call the foreground distribution for class $Y$. $\x_j$ for all $j \neq Z$ are drawn independently from a background distribution $D_0$. An illustration of SDC is given in the appendix\footnote{\href{https://arxiv.org/pdf/2307.13421.pdf}{https://arxiv.org/pdf/2307.13421.pdf}} 

 \begin{figure*}
  \centering
    \begin{subfigure}[b]{0.3\textwidth}
    \includegraphics[width=\textwidth]{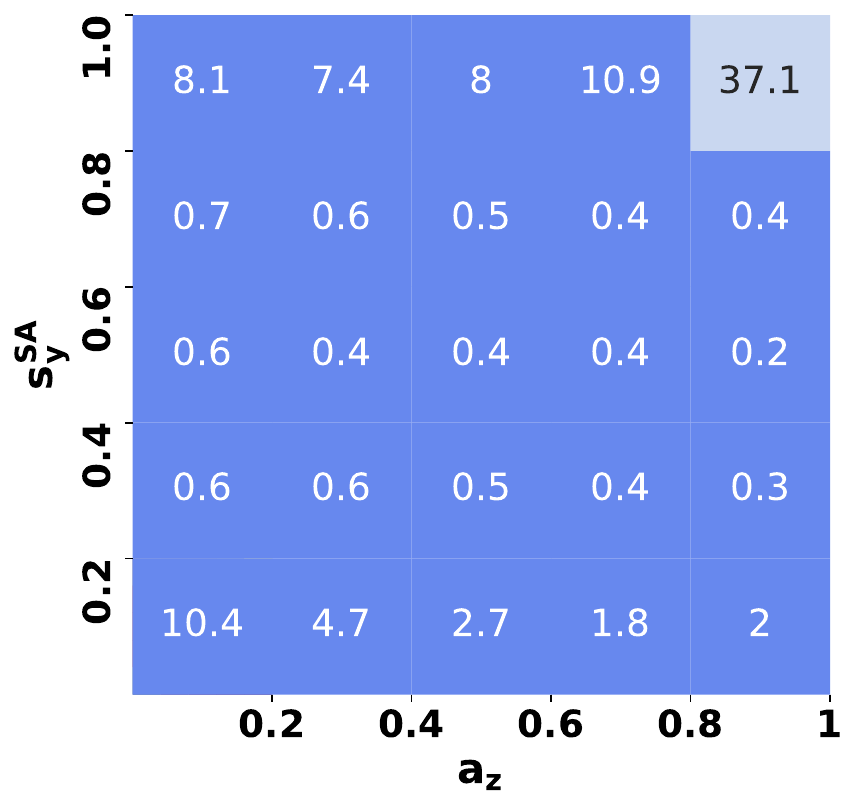}
       \caption{Soft Attention }
        \label{fig1:subfig4}
    \end{subfigure}
    \begin{subfigure}[b]{0.3\textwidth}
    \includegraphics[width=\textwidth]{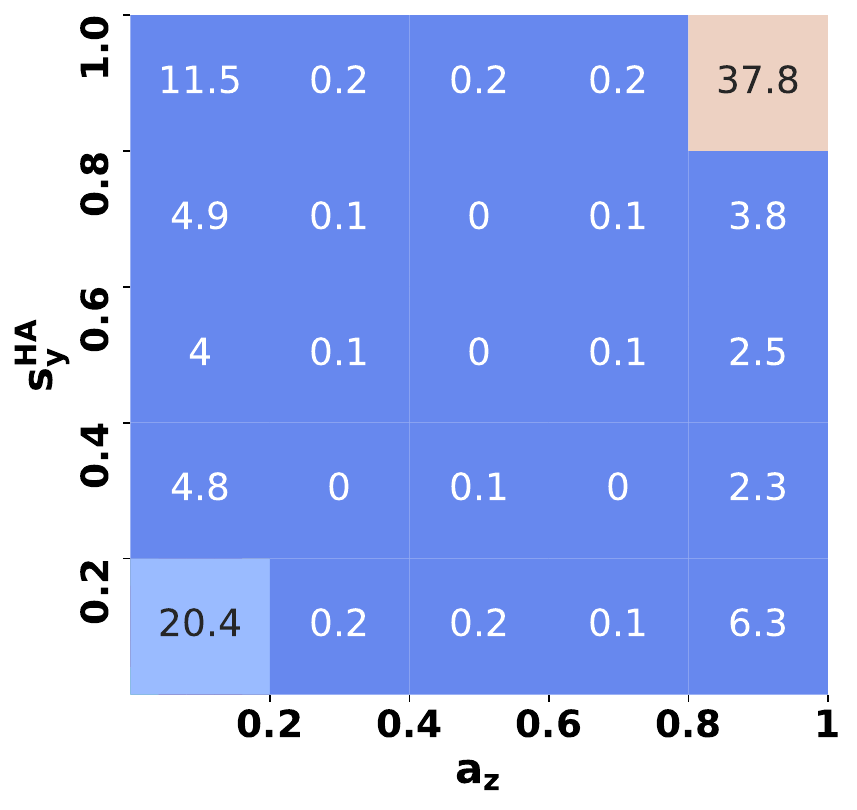}
       \caption{Hard Attention }
        \label{fig1:subfig5}
    \end{subfigure}
    \begin{subfigure}[b]{0.3\textwidth}
    \includegraphics[width=\textwidth]{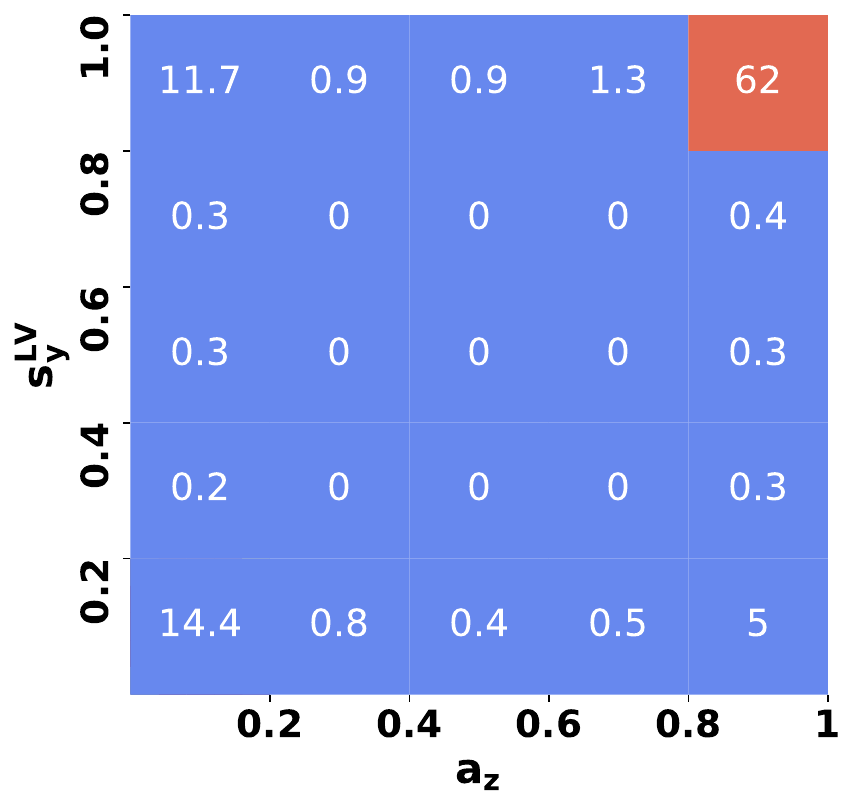}
       \caption{LVML }
        \label{fig1:subfig6}
    \end{subfigure}
    \caption{Focus-Prediction heat maps of test data for the three attention paradigms on the CIFAR10 SDC task with $n=10000$ and $m=5$.}
    \label{fig1:my_label}
\end{figure*}

\subsection{Paradigms of Attention}

Consider the following template attention model for the SDC task that we call a Focus-Classify Attention Model (FCAM). The FCAM is parameterised by a focus function $f:\R^d \> \R$ that scores each segment, and a classification function $\g:\R^d \> \R^C$ that classifies an `aggregated segment'. The different modes of aggregation and penalties for misclassification give rise to the three loss functions used commonly in attention -- soft attention (SA), latent variable marginal likelihood (LVML), and hard attention (HA). 

\begin{align*}
    L^\text{SA}(f, \g, \bX, y) &= -\log(\sigma_y(\g(\sum_j a_j(\bX) \x_j))) \\
    L^\text{LV}(f, \g, \bX, y) &= -\log(\sum_j a_j(\bX) \sigma_y(\g(\x_j))) \\
    L^\text{HA}(f, \g, \bX, y) &= - \sum_j a_j(\bX) \log(\sigma_y(\g(\x_j)))  \\ 
\end{align*}
where $\R^m \ni \ba(\bX) = \sigma(f(\x_1), \ldots, f(\x_m))$ is the normalised score given to the m segments in the mosaic input $\bX$ by the focus model $f$. It can easily be seen that $L^\text{LV}$ corresponds exactly to the negative log-likelihood of the SDC model, and hence the minimizer of $L^\text{LV}$ is the maximum-likelihood estimator for the SDC problem. The soft (hard) attention loss is usually motivated as an approximation (Jensen's inequality upper bound relaxation) of $L^\text{LV}$ \cite{NEURIPS2018_b691334c}. 
Also, it is important to note that in the case when the attention vector $\ba(\bX)$ has only one non-zero entry, all three losses become equal and that is indeed a desirable property when training such models. However, this does not happen in practice as the initial focus network $f$ is usually such that the entries of $\ba$ are all approximately $1/m$.

In the training phase, the focus and classification models $f,\g$ are learnt by optimising one of the loss functions above, and the learnt model is used for making a prediction on a new mosaic instance $\bX$ using the three corresponding inference procedures as (i) $s^\text{SA}_k =  \sigma_k(\g(\sum_j a_j \x_j))$, (ii) $s^\text{LV}_k =  \sum_j a_j \sigma_k(\g( \x_j))$, and (iii) $s^\text{HA}_k =  \sigma_k(\g( \x_{j^*}))$ 
where, \mbox{$j^* = \argmax_{j} f(\x_j)$} and \mbox{$\ba = [\sigma(f(\x_1), \ldots,$} $f(\x_m))]$. 

The vectors $s^\text{SA}, s^\text{LV}$ and $s^\text{HA}$ are the scores for the $C$ classes given by the aggregation methods. The final class prediction is done by simply returning the maximising co-ordinate of the score vector. Given a focus model $f$ and classification model $\g$ any of the above 3 inference procedures can be used on a test point for making a prediction, but the default choice is to use the inference procedure corresponding to the loss function that the model $f,\g$ minimises.

Despite the latent variable marginal likelihood loss having pride of place as the classic ML estimator, it is typically less preferred over the soft attention and hard attention paradigms. The soft attention loss has the advantage of being efficient in the number of calls to the computationally expensive $\g$ function, as the segment aggregation happens \textit{before} passing it to $\g$. This is especially advantageous in situations where the alignment variable $Z$ takes an exponentially large number of values (a typical alignment example that generates a $5$ word caption for an image with 196 segments has $196^5$ values). While the hard attention loss $L^\text{HA}$ has the same issue as $L^\text{LV}$ it can be efficiently approximated as it can be expressed as an expectation. While the computational issues with the three paradigms are well known, the differences in learning dynamics and the final model learnt when using the different loss functions are not as well studied.

\subsection{Comparative Empirical Analysis of the Attention Paradigms} \label{sec:2.4}


In this section, we perform an empirical study on a synthetic SDC task based on the CIFAR10 dataset and identify some key characteristics of the final model learnt in all three paradigms. The dataset for the SDC task was generated as follows. The label $y$ for each mosaic instance takes one of three possible values (\text{car, plane, bird}). The $m-1$ background segments of the corresponding mosaic instance $\bX$ are drawn randomly from images of the other 7 classes, the foreground segment is drawn from images corresponding to the foreground class given by the label $y$. Note that the position (or index) of the foreground segment can be arbitrary. We sample several such pairs (mosaic instances, labels), train an FCAM on a subset of this dataset, and evaluate it on the rest. The architecture of the focus model $f$ and classification model $\g$ are both convolutional neural networks with three convolutional and four fully connected layers.

 The results of the experiment (\href{https://github.com/VASHISHT-RAHUL/On-the-Learning-Dynamics-of-Attention-Networks/tree/main}{github link}) with $m=5$ segments and $n=10000$ training points are given in Figure \ref{fig1:my_label}. (More results in other settings and also with CIFAR100 using large number of classes are in the appendix). The Focus-Prediction heat map of any given FCAM $f,\g$ simply gives the joint histogram of the normalised focus score of the true foreground $a_{z}$, and the model score of the true class $s^\text{AP}_y$ where $\text{AP}$ represents the attention paradigm used and takes values in $\{\text{SA, LV, HA}\}$. Here $z$ corresponds to the true value of the hidden alignment variable $Z$ and $y$ corresponds to the true label $Y$. Note that computing this heat map requires access to the hidden alignment variable $Z$ even though the models $f,\g$ have been trained without access to it. 

  A perfect model would have all instances in the top right corner of the heat map (we define this as the Strongly Accurate Interpretable Fraction, SAIF ). Instances in the top left correspond to mosaic instances being classified correctly despite the focus model $f$ giving a low score to the foreground segment. Instances in the bottom right correspond to mosaic instances where the focus model $f$ scores the foreground patch much higher than the background patches, but the final score for the true class is low.
 
 The Focus-Prediction heat maps in Figure \ref{fig1:my_label} reveal some interesting patterns (that are also present in other settings, see supplementary material). 

  \begin{itemize}
  \setlength\itemsep{1em}
    \item The models trained with soft attention gives confident class label predictions (observe that row sums of the middle 3 rows in Figure \ref{fig1:my_label} (a) are close to zero)
    \item The focus model in the FCAM trained with soft attention loss is often not very confident (observe the column sums of the middle 3 columns are non-negligible in Figure \ref{fig1:my_label}(a)) 
    \item The models trained with hard attention do not give confident class label predictions (observe that row sums of the middle 3 rows in Figure \ref{fig1:my_label}(b)  are non-negligible)
    \item The focus model in the FCAM trained with hard attention loss is often confidently wrong or right (observe the column sums of the middle 3 columns are close to zero in Figure \ref{fig1:my_label}(b) ) 
    \item The FCAM models trained with latent variable marginal likelihood have both confident class label predictions and focus scores (observe that the row (columns) sums for the middle rows (columns) are close to zero in Figure \ref{fig1:my_label}(c))
    \item The top-right number in the heat map -- corresponding to the fraction of instances classified correctly after being focussed correctly -- in the LVML paradigm is often noticeably higher than the SA and HA paradigms.
 \end{itemize}

 \begin{figure*}
  \centering
    \begin{subfigure}[b]{0.32\textwidth}
    \includegraphics[width=\textwidth]{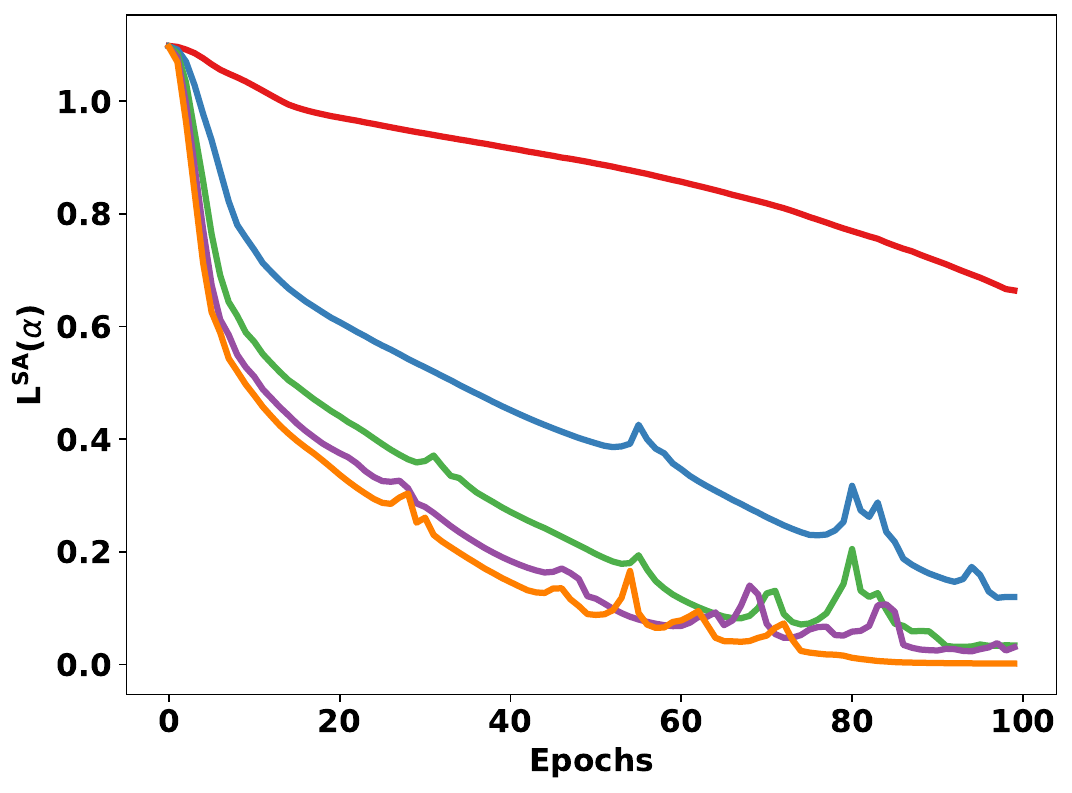}
       \caption{CIFAR10 Soft attention }
        \label{fig2:subfig1}
    \end{subfigure}
    \begin{subfigure}[b]{0.32\textwidth}
    \includegraphics[width=\textwidth]{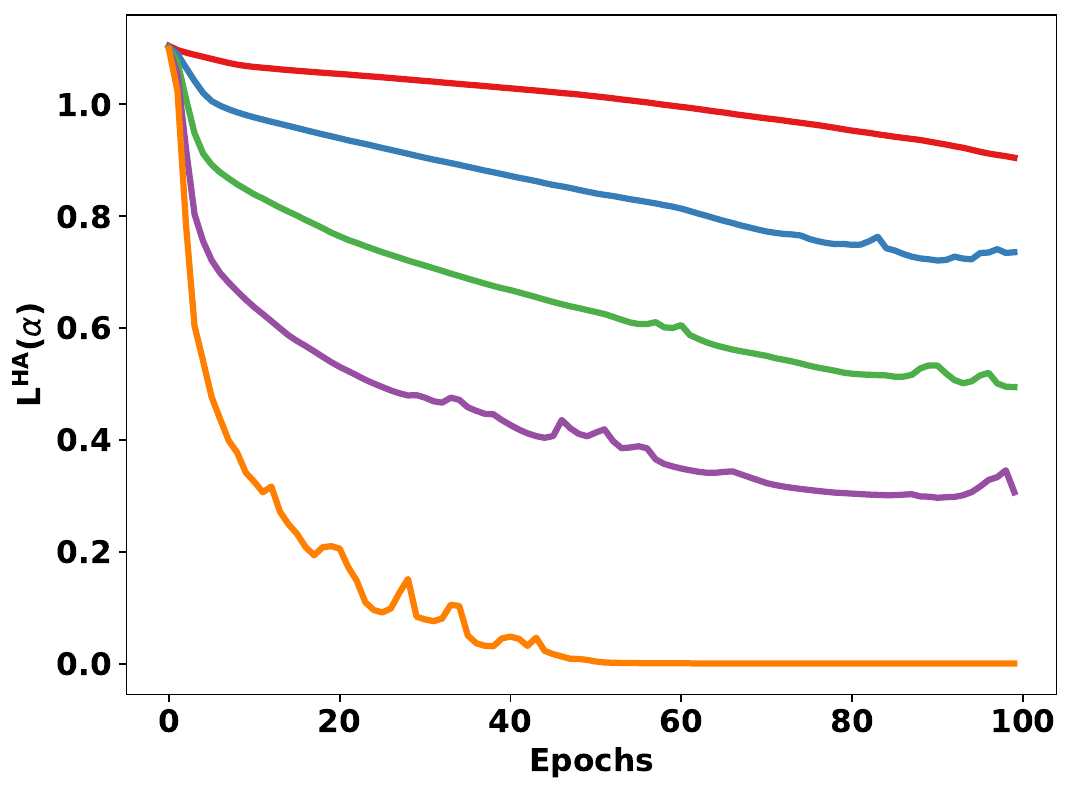}
       \caption{CIFAR10 Hard Attention}
        \label{fig2:subfig2}
    \end{subfigure}
    \begin{subfigure}[b]{0.32\textwidth}
    \includegraphics[width=\textwidth]{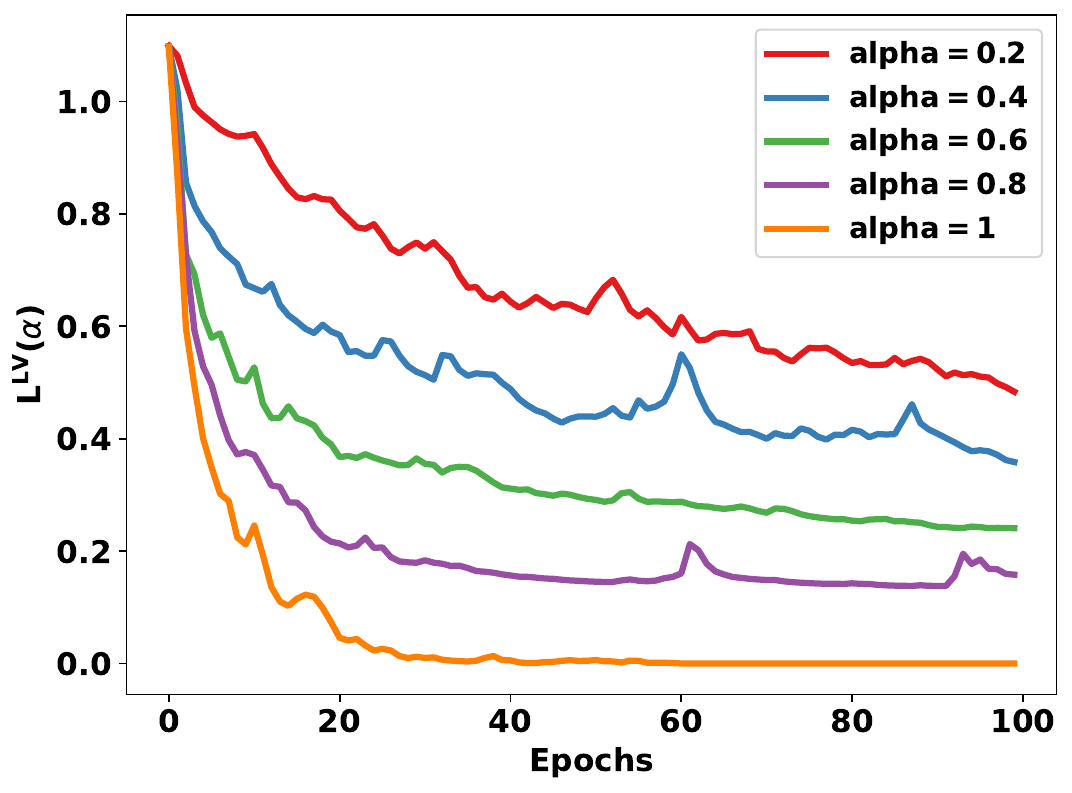}
       \caption{CIFAR10 LVML }
        \label{fig2:subfig3}
    \end{subfigure}
    \begin{subfigure}[b]{0.32\textwidth}
    \includegraphics[width=\textwidth]{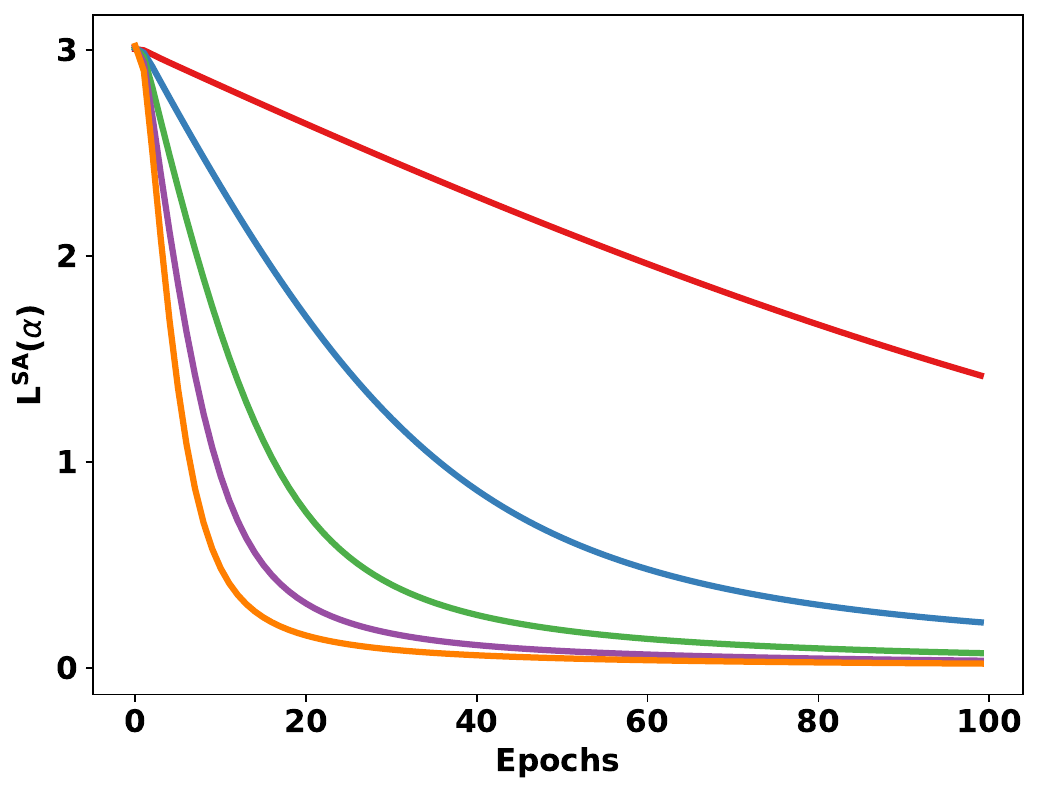}
       \caption{Linear-Orthogonal Soft Attention }
        \label{fig2:subfig4}
    \end{subfigure}
    \begin{subfigure}[b]{0.32\textwidth}
    \includegraphics[width=\textwidth]{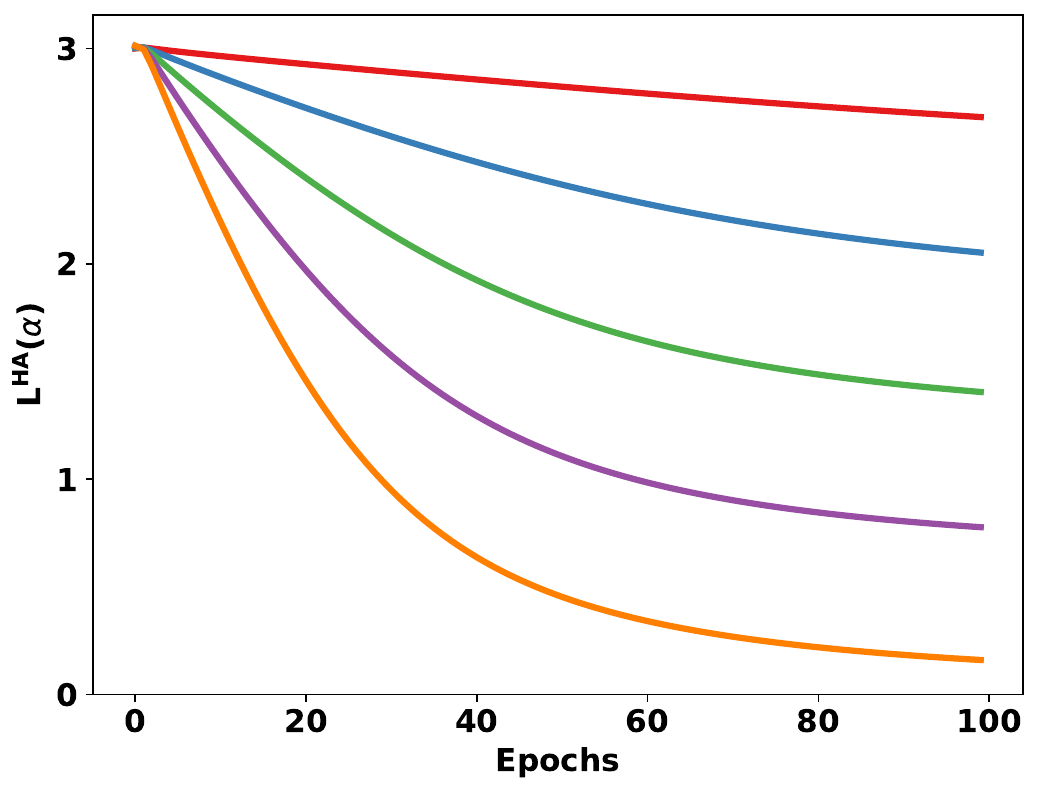}
       \caption{Linear-Orthogonal Hard Attention}
        \label{fig2:subfig5}
    \end{subfigure}
    \begin{subfigure}[b]{0.32\textwidth}
    \includegraphics[width=\textwidth]{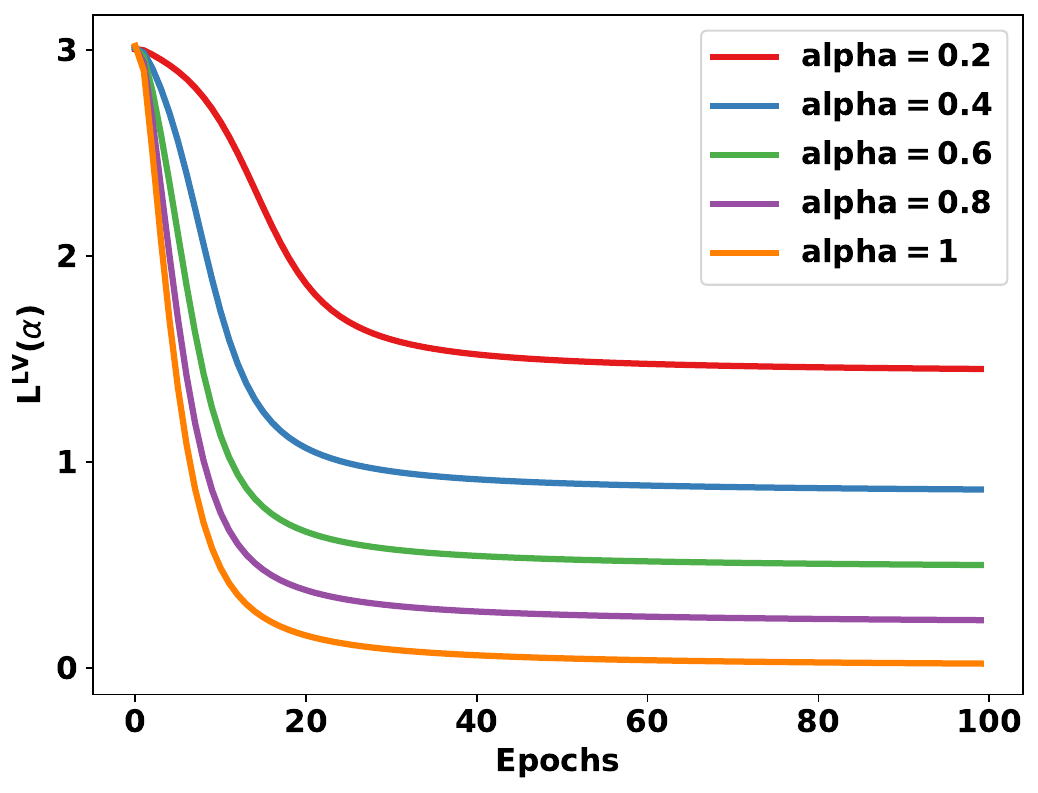}
       \caption{Linear-Orthogonal LVML}
        \label{fig2:subfig6}
    \end{subfigure}
    \caption{Log-Loss curves for train data for SDC task under fixed focus setting. The top row contains CIFAR10 data for $n=10000$ and $m=5$ and the bottom row is for linear orthogonal setting for $C=20$ and $m=20$. }
    \label{fig2:my_label}
\end{figure*}

 The LVML paradigm is clearly superior to the other paradigms, however, computational complexity issues stand in the way of choosing it. The rest of the paper comes up with a simple explanation for why the three paradigms behave the way they do, and tries to come up with an approach that performs as well as the LVML paradigm, but without its computational issues.

\section{Learning Dynamics of Attention Models}

The loss functions for attention are typically simultaneously minimized over $f$ and $\g$. However, it is instructive to analyze the dynamics of the evolution of $\g$ for certain fixed focus models $f$. We consider focus models that give a score of $\alpha \in [0,1]$ to the foreground patch (i.e. $a_{j^*}=\alpha$ where $j^*$ is the index of the foreground patch) and $\frac{1-\alpha}{m-1}$ for all the background patches (i.e $a_j = \frac{1-\alpha}{m-1}$ for all $j\neq j^*$).

\subsection{Fixed Focus Loss Curves}
\label{subsec:fixed-focus-loss-curves}

For a given value of $\alpha$, the problem of optimising for $\g$ is similar to the standard $C$-class classification problem with all three paradigms, with different data properties. As $\alpha$ increases, the optimisation problem over $\g$ becomes simpler for all three paradigms. When $\alpha=1$, the optimisation problem over $\g$ for all three paradigms becomes equal to the classification problem of distinguishing between the foreground classes. However, the three paradigms simplify the optimisation problem over $\g$ in different ways for $\alpha \in [\frac{1}{m},1)$. An increase in $\alpha$ increases the margin between the $C$-classes in the case of soft attention, while it reduces the number of `outlier points' in the case of hard attention, and is a hybrid of these two in the case of latent variable marginal likelihood. 

Concretely, the fixed focus loss values for a given classification model $\g$ and mosaic instance $\bX$ with label $y$, and foreground index $j^*$ are given below for the three attention paradigms.

\begin{align*}
    L^\text{FF,SA} & = ~
     -\log[ \sigma_y(\g(\alpha\x_{j^*} + \tfrac{1-\alpha}{m-1}\sum_{j\neq j^*} \x_j))]  
  \\
    L^\text{FF,LV} & = ~ -\log[ \alpha\sigma_y(\g(\x_{j*})) + \tfrac{1-\alpha}{m-1} \sum_{j\neq j^*} \sigma_y(\g(\x_{j}))] & 
\\
L^\text{FF,HA} &=~ - \alpha \log \left[\sigma_y(\g(\x_{j^*}))\right] + \tfrac{1-\alpha}{m-1}\sum_{j\neq j^*} \log\left[ \sigma_y(\g(\x_j)) \right]
\end{align*}

From the above expressions, one can make the following observation. While it is possible to find a $\g$ that makes any of the above three expressions close to zero for any $\alpha$ and any given $(\bX,y)$ pair, the population expectation of $L^\text{FF, LV}$ and $L^\text{FF, HA}$ is bounded away from $0$ for any $\alpha<1$. In the Hard Attention (HA) and Latent Variable (LV) paradigms, the value of $1$-$\alpha$ represents the proportion of data coming from background segments that can have all possible labels. In the HA paradigm, $\alpha < 1$ indicates that a fraction of 1-$\alpha$ of the total data points are sampled from the set of background segments. Since background segments can appear with any foreground segment, they possess all possible labels, and achieving zero loss is not possible. In particular, this means that it is not possible to achieve low loss with the LVML and HA paradigms (on a large enough dataset) if the focus model $f$ is such that  $f(\x_j)$ is not large for a unique $j$ (we call such a $f$ as non-confident). On the other hand,  low losses are possible in the soft attention paradigm even with non-confident focus models.


Figure \ref{fig2:my_label} shows the evolution of fixed focus loss values when $\g$ is updated through gradient descent for varying values of $\alpha$. Figure \ref{fig2:my_label} gives such curves for both the CIFAR10 SDC data and a purely synthetic SDC setting (that we call the linear orthogonal setting) for which we can derive the trajectory of the parameters when running gradient flow on the population loss. The details of this setting are in section \ref{sec:lin-ortho}. We denote the classifier $\g$ in the $t^{\text{th}}$ epoch when using the attention paradigm $\text{AP}$ with fixed foreground focus score $\alpha$  as $\g^\text{AP}_{\alpha, t}$.

We make the following observations about the fixed focus losses based on Figure \ref{fig2:my_label}
\begin{itemize}
    \setlength\itemsep{1em}
    \item The soft attention loss eventually goes to zero for moderately large values of $\alpha$.
    \item The hard attention and LVML losses flatten at a value above $0$ for any $\alpha<1$.
    \item  The LVML fixed focus loss curves generally have a steeper decline at initialization than the other paradigms, and converges to a lesser loss value than the hard attention paradigm.
    \item The difference in the fixed focus loss curves for different $\alpha$ values is minimal at larger values of $\alpha$ in the case of soft attention, but remains significant for the other two paradigms.
    \item In the hard attention paradigm, the flat loss curve at small values of $\alpha$ indicates the difficulty faced by the classification module to improve when the focus model $f$ is not confident and correct.
\end{itemize}
While  the above trends are obvious in the synthetic linear-orthogonal setting, they are also clearly visible in the case of the CIFAR SDC dataset, indicating that this difference in behaviour is a characteristic of the paradigm used.

\begin{figure*}[!ht]
    \centering
    \begin{subfigure}[b]{0.32\textwidth}
       \centering
       \includegraphics[width=\textwidth]{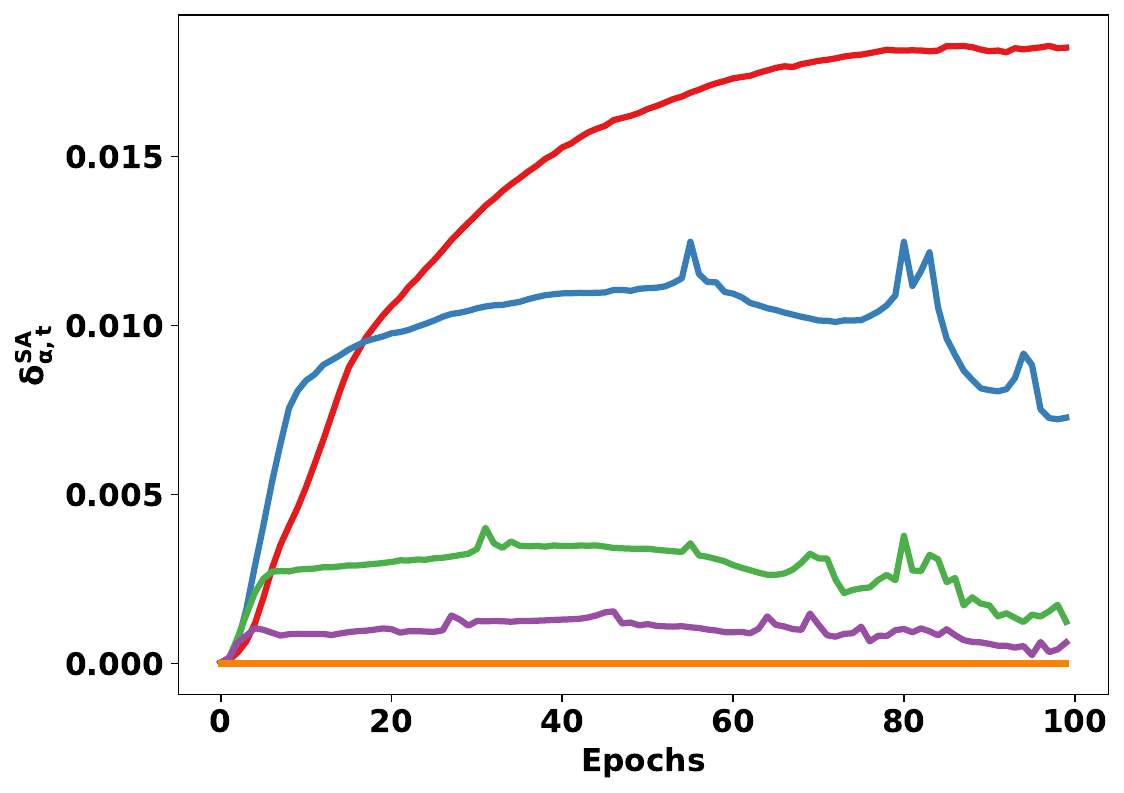}
       \caption{CIFAR10 Soft attention }
        \label{fig3:subfig1}
    \end{subfigure}
    \begin{subfigure}[b]{0.32\textwidth}
       \centering
       \includegraphics[width=\textwidth]{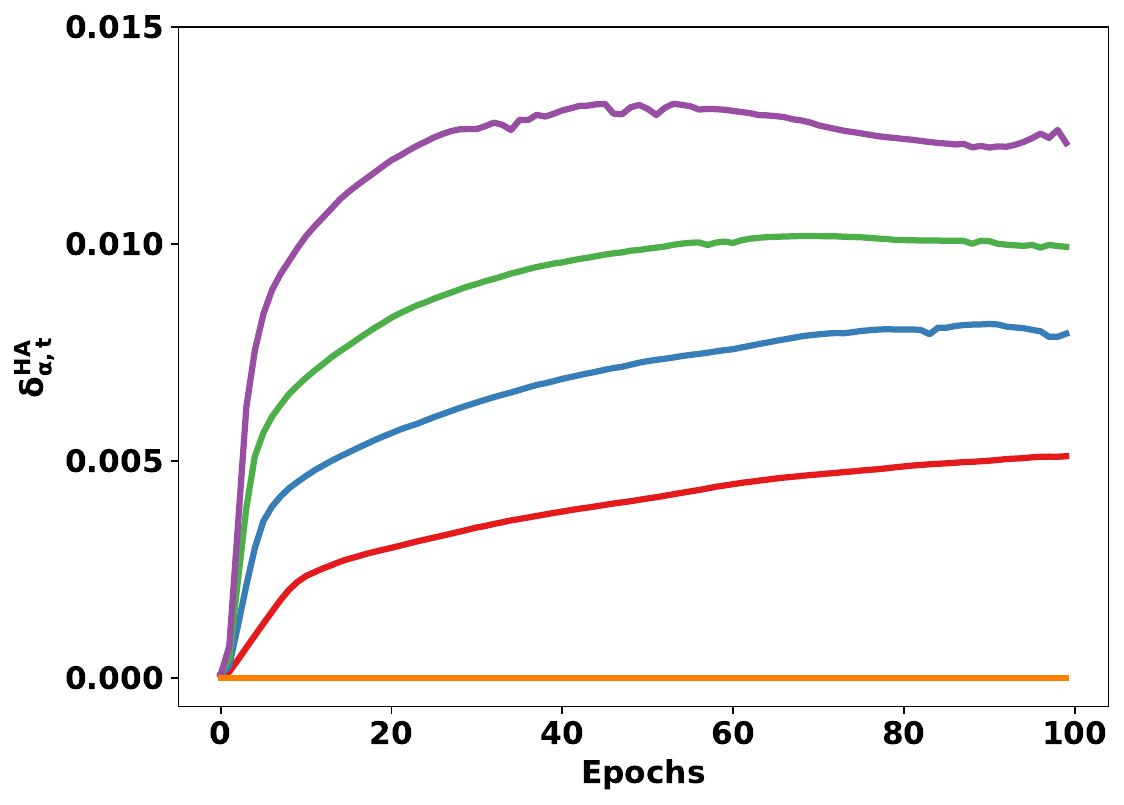}
       \caption{CIFAR10 Hard attention }
        \label{fig3:subfig2}
    \end{subfigure}
    \begin{subfigure}[b]{0.32\textwidth}
       \centering
       \includegraphics[width=\textwidth]{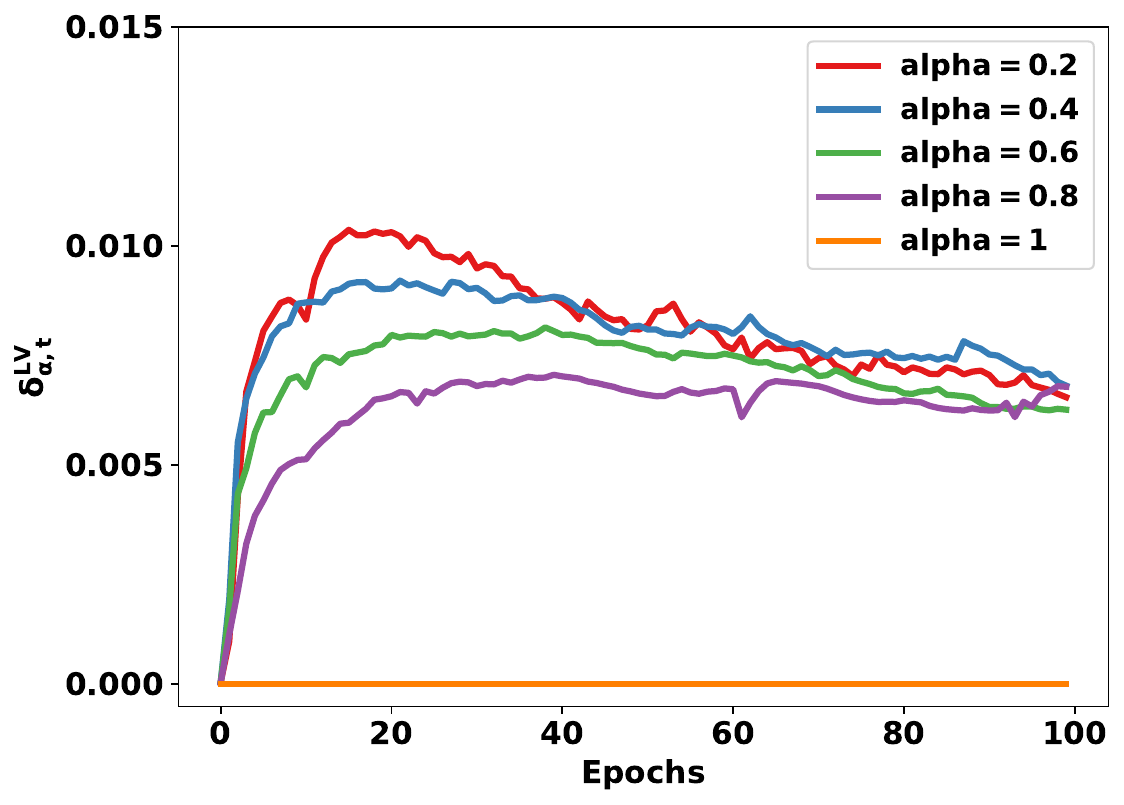}
       \caption{CIFAR10 LVML}
        \label{fig3:subfig3}
    \end{subfigure}
    \begin{subfigure}[b]{0.32\textwidth}
       \centering
       \includegraphics[width=\textwidth]{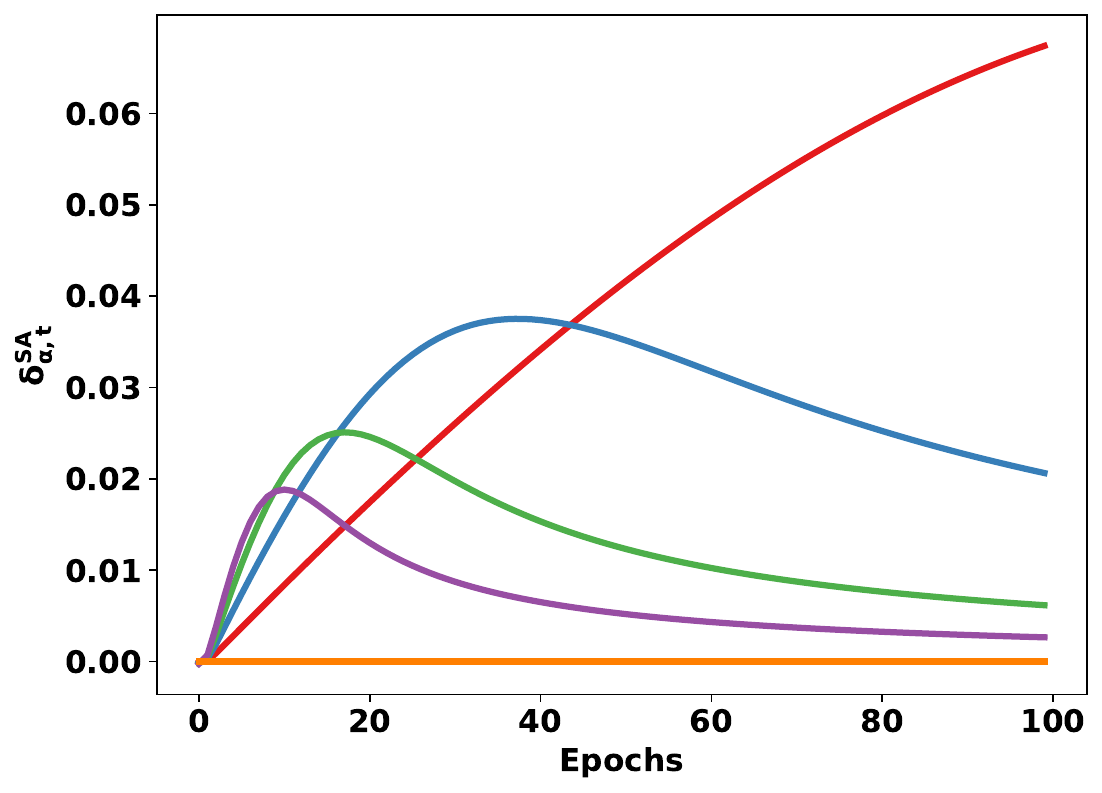}
       \caption{Linear-Orthogonal Soft attention}
        \label{fig3:subfig4}
    \end{subfigure}
    \begin{subfigure}[b]{0.32\textwidth}
       \centering
       \includegraphics[width=\textwidth]{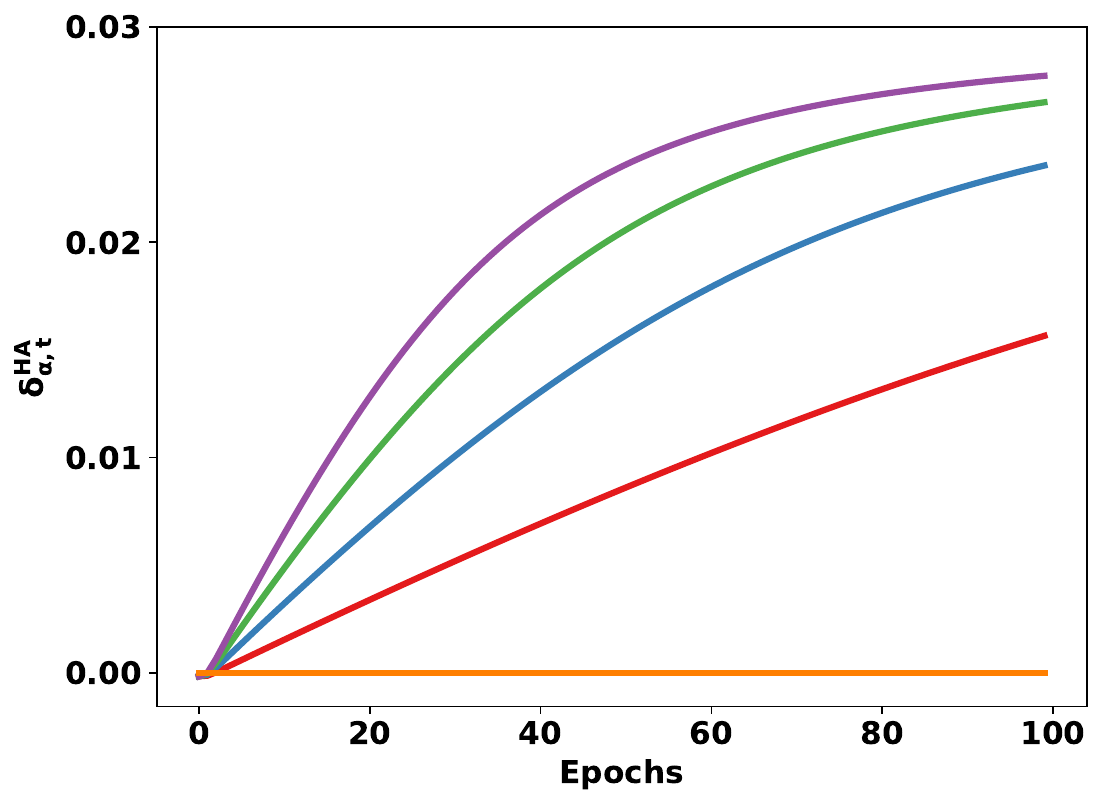}
       \caption{Linear-Orthogonal Hard attention}
        \label{fig3:subfig5}
    \end{subfigure}
    \begin{subfigure}[b]{0.32\textwidth}
       \centering
       \includegraphics[width=\textwidth]{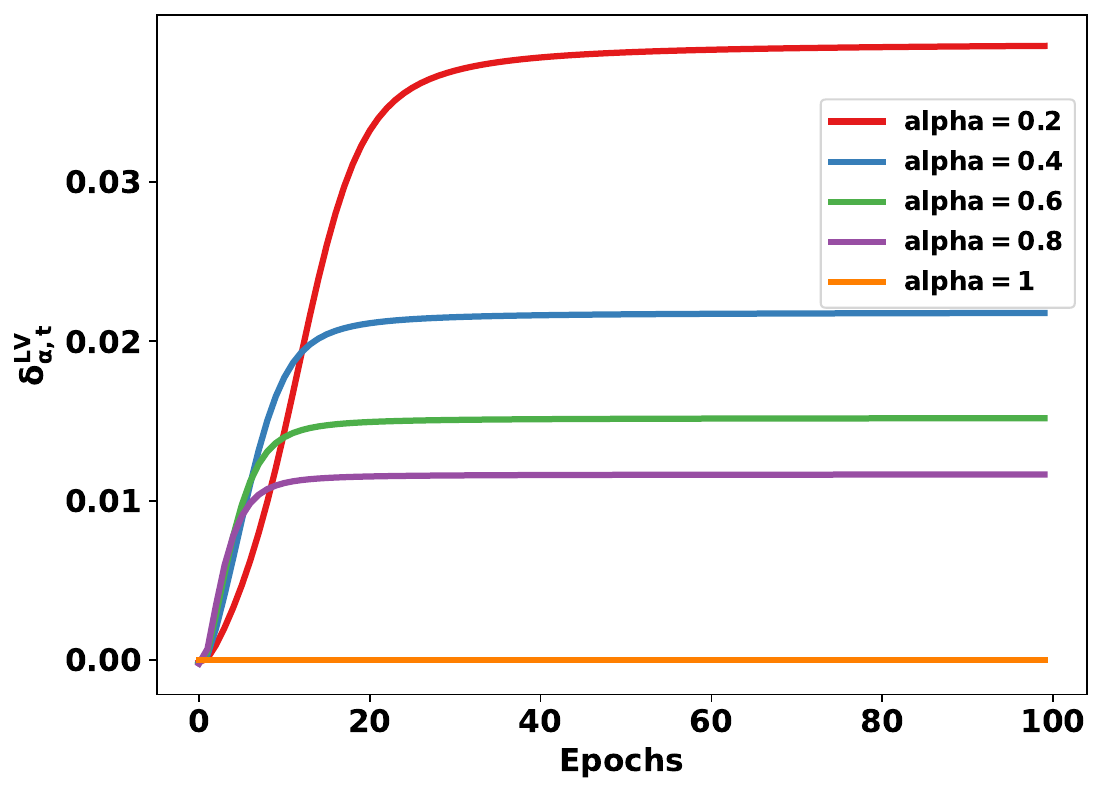}
       \caption{Linear-Orthogonal LVML }
        \label{fig3:subfig6}
    \end{subfigure}
    \caption{Focus Improvement Incentive curves for train data for SDC task under fixed focus setting. The top row contains CIFAR10 data for $n = 10000$ and $m = 5$ and the bottom row is for linear orthogonal setting for $C = 20$ and $m = 20$.  }
    \label{fig3:my_label}
\end{figure*}

\subsection{Focus Improvement Incentive Curves}
\label{subsec:focus-improvement-incentive}
In the previous section, we studied the evolution of the classification network, when the focus network $f$ is (somehow magically) fixed to give a score of $\alpha$ to the foreground and $\frac{1-\alpha}{m-1}$ the background segments. Note that this is trivially possible for $\alpha=\frac{1}{m}$ by setting the focus network to be identically equal to $0$. 

The prevailing intuition for learning dynamics of  attention models is the following. The classification network $\g$ can capture a non-negligible signal distinguishing the $C$ foreground classes even with this trivial focus network and improve. As $\g$ gets better, the incentive for the focus network to `improve' by giving higher scores to the foreground segment increases, and thereby making the optimisation problem faced by the classification network $\g$ easier and kick-starts a virtuous cycle.

We quantify the above phenomenon for the three attention paradigms, by defining the `incentive' for the focus model to improve as follows. For any given classification model $\g$, the focus improvement incentive at focus value $\alpha$ is given by $\E_{(\bX,y)} \left[ - \frac{d}{d \alpha} L^\text{FF, AP} (\alpha, \g, \bX, y) \right]$. This captures the reduction in the loss for a given classification model $\g$, as the focus model gets better. In particular, we study this quantity for the classifiers $\g^\text{AP}_{\alpha,t}$ obtained while minimising the fixed focus losses, where we approximate the derivative with a finite difference.

\begin{align*}
    \delta^\text{AP}_{\alpha,t} = L^\text{FF,AP}(g^\text{AP}_{\alpha,t},\bX,y,\alpha) - L^\text{FF,AP}(g^\text{AP}_{\alpha,t},\bX,y,\alpha')
\end{align*}
where, $\alpha' = \min(\alpha+0.01,1)$ and $\text{AP}$ takes values in $\{\text{SA}, \text{HA}, \text{LV}\}$. 

We make the following observations based on the focus improvement incentive curves in Figure \ref{fig3:my_label}.
\begin{itemize}
\setlength\itemsep{1em}
    \item The focus improvement incentive for the soft attention paradigm diminishes with increase in $\alpha$ (when considering $\g^\text{SA}_{\alpha,t}$ for moderately large $t$), and becomes quite negligible as $\alpha$ approaches $1$. (See Figure \ref{fig3:my_label}(a, d)).
    
    \item The focus improvement incentive for the hard attention paradigm increases with increase in $\alpha$, and remains quite small till very large $t$ for small $\alpha$ (See Figure \ref{fig3:my_label}(b, e)).

    \item The focus improvement incentive for the LVML paradigm also diminishes with increase in $\alpha$ similar to the soft attention paradigm, but the fall in incentive is not nearly as steep and remains bounded away from zero even for large values of $\alpha$ (See Figure \ref{fig3:my_label}(c, f)).
\end{itemize}

\subsection{Explanation for the Attention Paradigm Behaviour }
Based on the observations in Sections \ref{subsec:fixed-focus-loss-curves} and \ref{subsec:focus-improvement-incentive}, we now attempt an explanation of the behaviour of the attention paradigms when the entire model (i.e. both $f$ and $\g$) is optimised simultaneously. In particular, we explain the findings in Section \ref{sec:2.4} that comment on the results in Figure \ref{fig1:my_label}. 

The final soft attention Focus-prediction heat maps in Figure \ref{fig1:my_label} have a large number of mosaic instances with medium values for the focus score $a_z$, because the need for the focus network to be confident and correct is not present -- i.e. the soft attention loss $L^{\text{FF, SA}}_\alpha$ can be driven close to zero even for $\alpha < 1$ (See green, purple and orange curves in Figure \ref{fig2:my_label} (a,d)). The incentive for increasing $\alpha$ also falls rapidly with increase in $\alpha$ through the optimisation process (See Figure \ref{fig3:my_label} (a,d). 

The final hard attention model has a large fraction of instances where the focus score on the foreground segment $a_z$ is small ($<0.2$) or large ($>0.8$). This is likely due to the fact that the fixed focus loss curves are almost flat for small values of $\alpha$ (See red curve in Figure \ref{fig2:my_label}(b,d)) and the incentive for the focus network to improve is significantly lesser (see red curve in Figure \ref{fig3:my_label}(b,d)). However, the large incentive for the focus to improve when $\alpha$ is moderately large (see green, blue and red curves in Figure \ref{fig3:my_label}(b,d)) ensures that very few instances remain in the moderate $a_z$ region: i.e. the focus scores go all the way to $1$ or stay around $\frac{1}{m}$.

The latent variable marginal likelihood model seems to combine the best of both the soft and hard attention paradigms: a significant incentive for the focus network to improve for small $\alpha$ that does not decay too rapidly even for large $\alpha$, and hence it appears to achieve the best result while training FCAMs for the SDC task.

\begin{figure*}[!ht]
    \centering
        \begin{subfigure}[b]{0.24\textwidth}
       \centering
        \includegraphics[width=\textwidth]{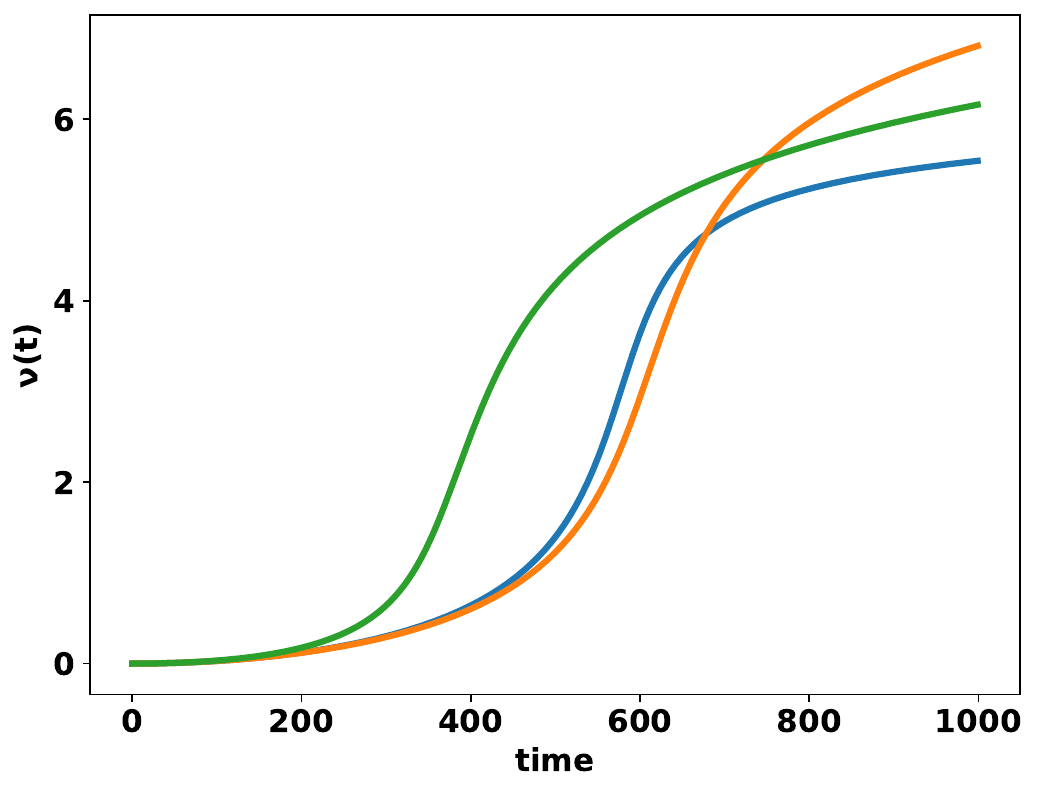}
       \caption{$\nu(t) : m=20 \text{, } C=20$ }
        \label{fig4:subfig1}
    \end{subfigure}
    \begin{subfigure}[b]{0.24\textwidth}
       \centering
        \includegraphics[width=\textwidth]{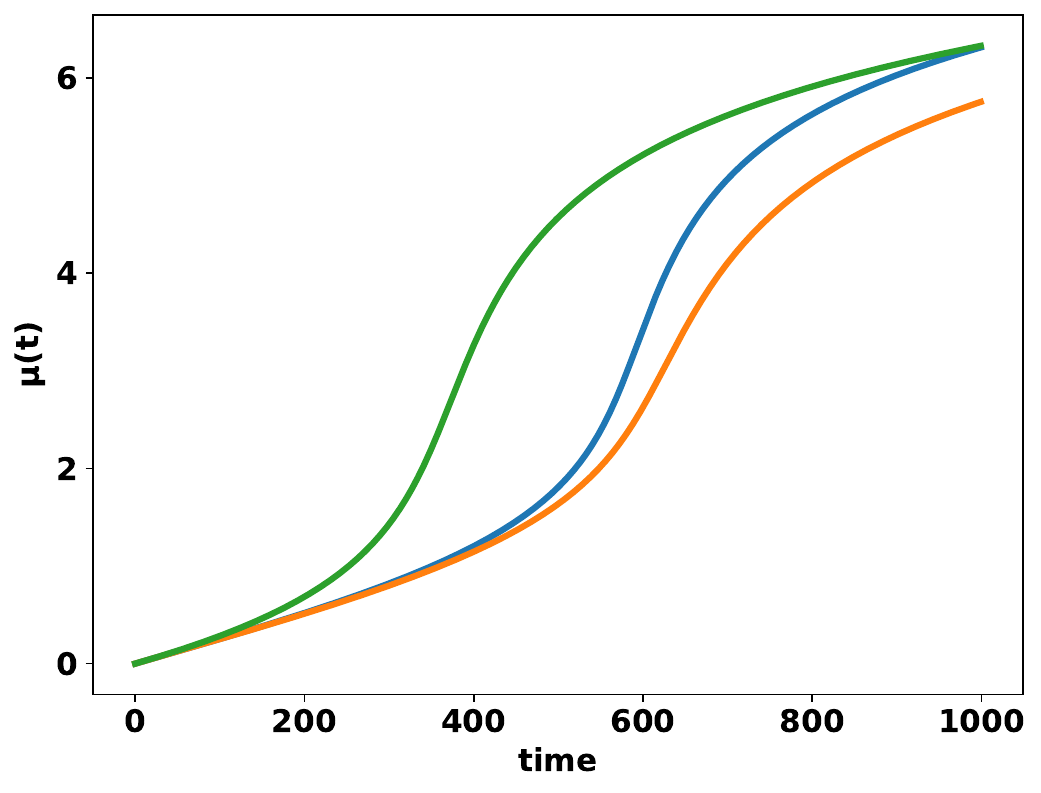}
       \caption{$\mu(t) : m=20 \text{, } C=20$}
        \label{fig4:subfig2}
    \end{subfigure}
    \begin{subfigure}[b]{0.24\textwidth}
       \centering
        \includegraphics[width=\textwidth]{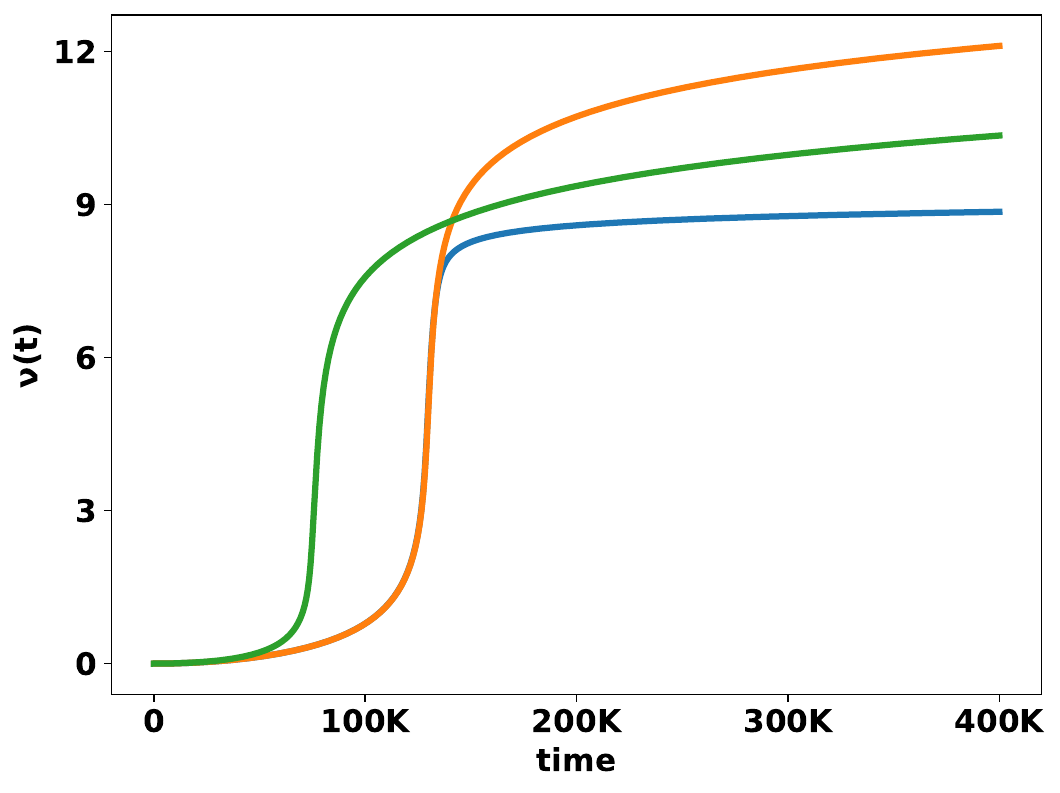}
       \caption{$\nu(t) : m=100 \text{, } C=1000$}
        \label{fig4:subfig3}
    \end{subfigure}
    \begin{subfigure}[b]{0.24\textwidth}
       \centering
        \includegraphics[width=\textwidth]{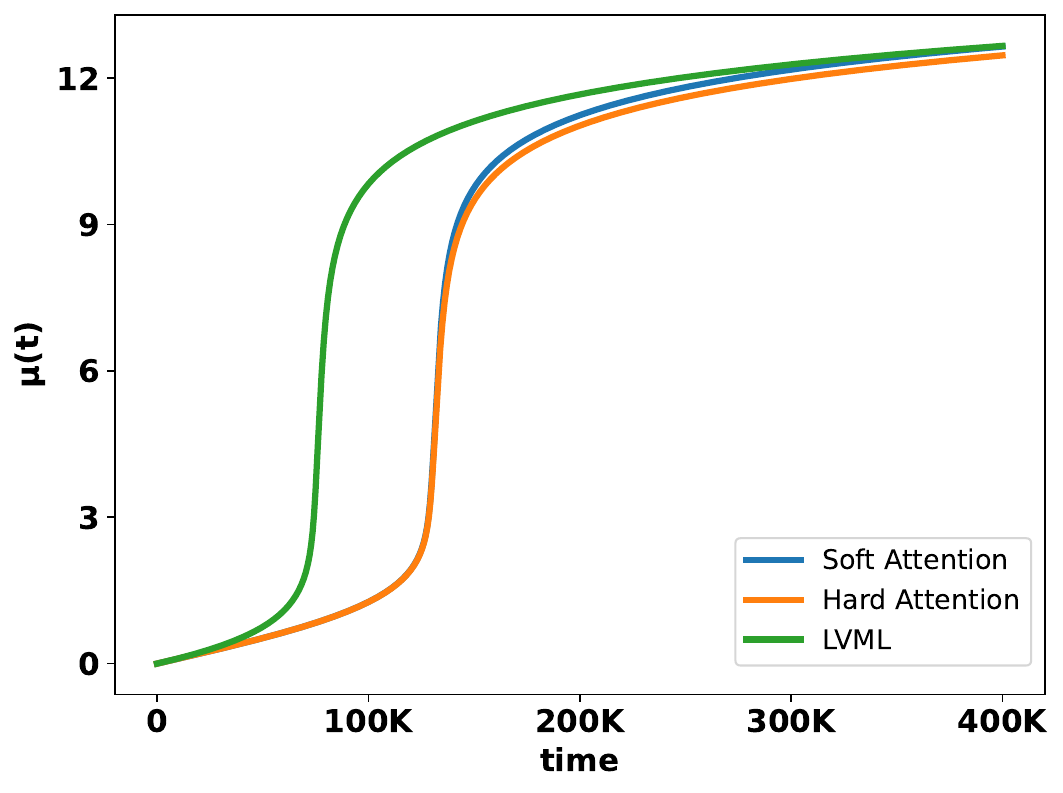}
       \caption{$\mu(t) : m=100 \text{, } C=1000$ }
        \label{fig4:subfig4}
    \end{subfigure}
    \caption{Evolution of the focus model ($\nu(t)$ in Theorem 2) and  classification model ($\mu(t)$ in Theorem 2) for linear orthogonal setting} 
    \label{fig4:my_label}
\end{figure*}

\begin{table*}[!ht]
    \centering
    \begin{tabular}{|c|c|c|c|c||c|c|c|c|}
         \hline
          \multicolumn{1}{|c|}{-}  & \multicolumn{4}{|c||}{SAIF (\%) } & \multicolumn{4}{|c|}{ Accuracy (\%)} \\
         \hline 
         Dataset & Hard  & Soft & Hybrid & LVML & Hard & Soft & Hybrid & LVML \\
         \hline
         CIFAR10 ($m=5$) & 37.8 & 37.1 & 59.0  &  62.0 & 63.3 & 75.3 & 76.6 & 77.9   \\
         \hline
         CIFAR10 ($m=20$) & 1.9 & 23.8 & 38.2 & 47.2 & 42.1 & 65.2 & 66.0 & 71.0  \\
         \hline
         CIFAR100 ($m=5$) & 13.6 & 79.2 & 83.2 & 77.5  & 46.6  & 83.2 & 80.0 & 79.7   \\
         \hline
         CIFAR100 ($m=20$) & 1.7 & 67.0 & 66.8 & 66.5 & 8.2 & 72.2 & 71.0  & 69.0 \\
         \hline
         HateXplain-1    & 2.7 & 4.6 & 21.6 & 22.6  & 54.3 & 51.6 & 54.6 & 54.6   \\
         \hline
         HateXplain-2  & 23.4  & 9.2  & 24.6 & 19.2 & 49.8 & 49.63 & 49.0 & 44.7    \\
         \hline
         MSCOCO (Bleu4 Score) & 11.1  & 23.2 & 24.2 & -  & 10.4 & 20.9 &  20.4 & -     \\
         \hline 
         
    \end{tabular}
    \caption{Strongly Accurate Interpretable Fraction and Accuracy for various datasets and attention paradigms}
    \label{table1:my_label}
\end{table*}

\section{The Linear Orthogonal Setting}
\label{sec:lin-ortho}

In this section, we define a simple (toy-like) SDC task for which the gradient flow with the expected population gradient can be computed in closed form. The generative model under this setting has following assumptions. 
\begin{assumption}
    The foreground distributions are all supported on a single point, i.e. $D_y= \delta_{\bbs_y}$, where $\bbs_1, \bbs_2, \ldots, \bbs_C$ are all orthogonal vectors in $\R^d$.
\end{assumption}

\begin{assumption}
    The background distribution $D_0$ has zero mean.
\end{assumption}

\begin{assumption}
    The support of the background distribution $D_0$ is orthogonal to the vectors $\bbs_1, \bbs_2, \ldots, \bbs_C$.
\end{assumption}


The FCAM model we use for this analysis is simply a linear model for $f:\R^d\>\R$ parameterised by a $d$-dimensional vector $\bbu$ and a linear model for $\g:\R^d\>\R^C$ parameterised by a $C\times d$ weight matrix $[\bbw_1 , \bbw_2, \ldots, \bbw_C]^\top$.  While these assumptions are clearly very restrictive, the simple nature of this setting allows us to clearly see the fundamental difference between the three paradigms.

The ideal solution to the above problem would be such that the focus model $f$ has the background subspace in its null-space thus ensuring that in every mosaic instance $\bX$, the foreground segment $\x_{j^*}$ would get a positive attention score $f(\x_{j^*})$ while all the background patches evaluate to $0$. The ideal classification model $\g$ would then be a model that classifies $\bbs_y$ as $y$ for all $y \in [C]$. We show that gradient flow on the population in all three paradigms do converge to the same ideal solutions (up to a scalar multiple) but at differing rates.

\subsection{Fixed focus losses}
In this section, we first derive the dynamics of the classification model $\g$, when the focus model is fixed to give a score of $\alpha$ to the foreground segment and $\frac{1-\alpha}{m-1}$to all the background segments.

\begin{theorem}
\label{th1}
In the linear-orthogonal setting, when the focus network $f$ is fixed to score the foreground segment with value $\alpha \in [0,1]$ the parameter $\bbw$ of the classification model $\g$ varies as follows under gradient flow from a zero initialisation:
 \[\bbw^{\textup{AP}}_{k}(t) = \mu^{\textup{AP}}(t)\left[ \bbs_k - \dfrac{1}{C}\sum_{k'=1}^C \bbs_{k'}\right] \]
 where $\textup{AP}$ takes values in $\{\textup{SA}, \textup{HA}, \textup{LV}\}$ and $k \in [C]$. The scalar $\mu^\textup{AP}(t)$ varies as follows: 

  \begin{align}
\odv{ \mu^{\textup{SA}}(t)}{t} &=  \dfrac{\alpha }{\exp( \alpha \mu^{\textup{SA}}(t))+C-1} \label{eq1} \\
\odv{ \mu^{\textup{HA}}(t)}{t} &= \dfrac{\alpha \beta^{\textup{HA}}(t)}{\exp(\mu^{\textup{HA}}(t))} \label{eq2} \\
\odv{ \mu^{\textup{LV}}(t)}{t} &=  \dfrac{\alpha (\beta^{\textup{LV}}(t))^2}{Z(t)\exp(\mu^{\textup{LV}}(t))} \label{eq3}
 \end{align}
 where  \mbox{$Z(t) =\alpha \beta^{\textup{LV}}(t)+ \dfrac{1-\alpha}{C} $},  
  \mbox{$\beta^{\textup{AP}}(t) = \dfrac{\exp(\mu^{\textup{AP}}(t))}{\exp(\mu^{\textup{AP}}(t))+C-1}$}

\end{theorem}
The losses of the corresponding classification model and the focus improvement incentive are illustrated in Figure \ref{fig2:my_label}(d,e,f) and Figure \ref{fig3:my_label} (d,e,f) respectively. The similarity of linear orthogonal setting plots with CIFAR10 plots in Figures \ref{fig2:my_label} and \ref{fig3:my_label} suggests that differences in behaviour among the attention paradigms is due to an intrinsic property of the paradigms and not an artifact of the data.

\subsection{Gradient flow trajectory for the full FCAM}

The extreme simplicity of the linear orthogonal setting allow for the fixed focus model used in the previous section to actually manifest during training. In fact, when $f$ and $\g$ are optimised simultaneously, the focus model exactly takes the trajectory of the fixed focus model with $\alpha$ that increases from $\frac{1}{m}$ to $1$. Based on this we can derive the joint trajectory of the focus network parameter $\bbu$ and classification network parameter $\bbw$ as follows.

 \begin{theorem}
 \label{th2}
 Under the linear-orthogonal setting, when the initial values of $\bbu$ and $\bbw$ are set to zero, the parameters evolve under population gradient flow as follows
 \[\bbw^{\textup{AP}}_{k}(t) = \mu^{\textup{AP}}(t)\left[ \bbs_k - \dfrac{1}{C}\sum_{k'=1}^C \bbs_{k'}\right] \]
 \[ \bbu^{\textup{AP}}(t) = \nu^{\textup{AP}}(t) \sum_{k=1}^C \bbs_k\]
 
where $\textup{AP}$ takes values in $\{\textup{SA}, \textup{HA}, \textup{LV}\}$ and $k \in [C]$. The scalar $\nu^\textup{AP}(t)$ varies as follows:

\begin{align}
\odv{\nu^{\textup{SA}}(t)}{t} &= \dfrac{ \mu^{\textup{SA}}(t)(C-1) (\alpha^{\textup{SA}}(t)-(\alpha^{\textup{SA}}(t))^2) }{C(\exp(\alpha^{\textup{SA}}(t) \mu^{\textup{SA}}(t))+C-1)} \label{eq4}  \\
\odv{\nu^{\textup{HA}}(t)}{t} &= \frac{\log[C \beta^{\textup{HA}}(t)]}{C}(\alpha^{\textup{HA}}(t)-(\alpha^{\textup{HA}}(t))^2) \label{eq5} \\
\odv{\nu^{\textup{LV}}(t)}{t} &= \frac{\alpha^{\textup{LV}}(t)}{C}\bigg[\dfrac{\beta^{\textup{LV}}(t) }{Z(t)} - 1\bigg] \label{eq6}
\end{align}
\textup{where,} \mbox{$Z(t) =\alpha^{\textup{LV}}(t) \beta^{\textup{LV}}(t) + \dfrac{1-\alpha^{\textup{LV}}(t)}{C}$ },

\mbox{$\alpha^{\textup{AP}}(t) =  \dfrac{\exp(\nu^{\textup{AP}}(t))}{\exp(\nu^{\textup{AP}}(t))+m-1}$},

\mbox{$\beta^{\textup{AP}}(t) = \dfrac{\exp(\mu^{\textup{AP}}(t))}{\exp(\mu^{\textup{AP}}(t))+C-1}$}, \textup{and}

$\odv{\mu^{\textup{AP}}(t)}{t}$ follows the expression in Theorem \ref{th1}.

  \end{theorem}

Theorem \ref{th1} and \ref{th2} implies that under linear-orthogonal setting, the three algorithms trajectory differs only in terms of scalar multiples $\mu(t)$ and $\nu(t)$. Figure \ref{fig4:my_label} shows the simulated evolution of $\mu(t)$ and $\nu(t)$ for two different values of $m,C$. The LVML model clearly converges faster than both soft and hard attention. The hard attention model converges to larger value for $\nu(t)$ corresponding to a more confident focus model. The proofs for theorems are in given in the appendix.

\section{A Hybrid Attention Paradigm and Experiments}

A natural algorithm design goal here is to require that the final FCAM be both accurate and interpretable -- which can be enforced by requiring that the predicted probability score of the true class and the focus score of the true foreground are both large. This reduces to requiring that the top right number in the Focus-Prediction heatmap (SAIF) be as large as possible. The LVML loss function corresponding to maximum likelihood has ideal properties for achieving this, but it suffers from computational issues when the number of segments $m$ is large. 

Based on the observations in Sections \ref{subsec:fixed-focus-loss-curves} and \ref{subsec:focus-improvement-incentive}, a simple algorithm is proposed that combines the properties of both soft and hard attention. Soft attention provides strong incentives for the focus model to improve at initialization but stagnates later on. In the Hard attention paradigm the incentive for focus model to improve is small at initialization but becomes larger once a non-trivial focus model is attained. This suggests a ‘Hybrid’ approach where an FCAM trained with soft attention is used as the initial step for hard attention learning. Previous works have shown hard attention to actually be a hybrid method, where the model randomly selects either soft or hard attention during each epoch, however, this hybrid nature was an after effect of the unintentional random selection \cite{pmlr-v37-xuc15}. Our work is different in terms of conscious selection between hard and soft attention based on empirical observation. 


We perform experiments on semi-synthetic (refer section \ref{sec:2.4}) and real-world datasets.
The ``HateXplain'' dataset \cite{Mathew_2021} is used to classify text into one of three categories: hate speech, offensive, or normal. The dataset also includes explanations for the assigned label in the form of ``rationales'', which highlight specific portions of the text. The focus score of the true foreground is calculated as  $a_{z^*}= \sum_{z \in z^* } a_z$ (where $z^*$ specifies the locations in input responsible for the label). Experiments were conducted using both a standard embedding layer and a self-attention embedding layer (1 and 2 respectively), as shown in Table \ref{table1:my_label}).

Experiments were also conducted on the MSCOCO (2014) dataset for image captioning using an encoder-decoder model with attention, as described in \cite{pmlr-v37-xuc15}. The definition of SAIF was adapted for sequence generation by measuring the overlap between the attention vector and the bounding box of an object when a word corresponding to that object is output by the decoder \cite{pmlr-v189-pandey23a}. Due to the large number of classes and vocabulary, as well as the large number of patches ($m=196$), the threshold used for SAIF was relaxed to $0.3$ instead of the $0.8$ used in other experiments. 


Table \ref{table1:my_label} gives the SAIF and accuracy results of the experiments across different paradigms with equal compute time. The hybrid approach seems to increase the SAIF performance over soft attention without affecting the accuracy.

 \section{Conclusion}
In this paper, we study the learning dynamics of different paradigms of attention models observing that, in terms of interpretability, soft attention performs well at initialization and hard attention perform well at later stages of training. We propose a hybrid approach and demonstrate an improvement without incurring the same cost as maximum likelihood methods. This approach is applicable to any task, where the class label depends on a small but unknown segment of the input, and this location information is absent in the training data. However evaluating the improvement in interpretability requires access to this location information -- e.g. datasets such as HateXplain, TVQA+ \cite{lei-etal-2020-tvqa}.

\bibliography{ecai}

\clearpage
\onecolumn

\appendix
\section{Appendix}

\subsection{Codes for Reproducing Results} 
\label{c1}

All the datasets and codes are available \href{https://github.com/VASHISHT-RAHUL/On-the-Learning-Dynamics-of-Attention-Networks/tree/main}{\textbf{here}}.
\subsection{Experimental Details}
\label{experiment details}
\subsubsection{CIFAR10 Experiments}
\textbf{Soft Attention:} \\  We used a CNN based network for Focus as well as Classification module with 3 convolution layers and 4 linear layers{ Network architecture is same for all the algorithms for cifar10 }.We used SGD optimizer with momentum with learning rate of $0.01$ for setings $(m=5,n=10k;~~m=20,n=10k;)$ and tuned learning rate over search space of $[0.0005,0.001,0.002,0.004,0.01,0.02,0.05,0.1,0.2]$. We have used the $12$ random seed for parameter initialization for all the experiments.\\

\textbf{Hard Attention:}  \\  We used SGD optimizer with momentum with learning rate of $0.005$ for setings $(m=5,n=10k;~~m=20,n=10k;)$ and tuned learning rate over search space of $[0.0005,0.001,0.002,0.005,0.01,0.02,0.05,0.1,0.2]$. We have used the $12$ random seed for parameter initialization for all the experiments. \\

\textbf{LVML:}  \\  We used SGD optimizer with momentum with learning rate of $0.01$ for setings $(m=5,n=10k;~m=20,n=10k;$ and tuned learning rate over search space of $[0.0005,0.001,0.002,0.004,0.01,0.02,0.05,0.1,0.2]$. We have used the $12$ random seed for parameter initialization for all the experiments.

\subsubsection{CIFAR100 Experiments} 
\textbf{Soft Attention:} \\
We used a CNN based network for Focus module with 2 convolution layers and 2 linear layers and 3 convolution layers and 3 linear layers for classification module{ Network architecture is same for all the algorithms for cifar100}. We used SGD optimizer with momentum with learning rate of $0.005$ for setings $(m=5,n=50k;~~m=20,n=50k)$ respectively and tuned learning rate over search space of $[0.0005,0.001,0.002,0.005,0.01,0.02,0.05,0.1,0.2]$. We have used the $12$ random seed for parameter initialization for all the experiments.\\

\textbf{Hard Attention:} \\ We used SGD optimizer with momentum with learning rate of $0.001 $ and $0.007$ for setings  $(m=5,n=50k;~~m=20,n=50k;)$ respectively and tuned learning rate over search space of $[0.0005,0.001,0.002,0.005,0.01,0.02,0.05,0.1,0.2]$. We have used the $12$ random seed for parameter initialization for all the experiments. \\

\textbf{LVML:} \\ We used SGD optimizer with momentum with learning rate of $0.01$ for all setings $(m=5,n=50k;~~m=20,n=50k;)~~$ and tuned learning rate over search space of $[0.0005,0.001,0.002,0.004,0.01,0.02,0.05,0.1,0.2]$. We have used the $12$ random seed for parameter initialization for all the experiments.\\

\textbf{Fixed Focus Setting:} \\ For all the fixed focus settings for both CIFAR10 and CIFAR100, we have used the same tuned learning rate as above.
The classification network architecture is also same as mentioned above.

\subsection{HateXplain Experiments}
\textbf{Standard Embedding layer}
We use an embedding layer with vocabulary size $28041 $ and embedding size $100$. The focus network has one hidden layer with $200$ units and Relu activation. The classification network has one hidden layer with 200 units and Relu activation. We SGD optimizer with momentum and  learning rate of $0.1$ for soft attention, hard attention, and $0.3$ for LVML Model. We tuned the learning rate over the search space of $[0.5,0.3,0.1,0.05,0.01]$. We have used the $12$ random seed for parameter initialization for all the experiments. For the hybrid model, we use a learning rate of $0.1$ for soft attention and $0.01$ for hard attention.
 
\textbf{Self-Attention based Embedding}

We use an embedding layer with vocabulary size $28041 $ and embedding size $100$. Here we consider a multi-head self-attention model with positional encoding and 2-heads based on the transformer model. The embeddings from the self-attention model are fed into the focus model for cross-attention. The focus network has one hidden layer with $50$ units and Relu activation. The classification network  has one hidden layer with 200 units and Relu activation. We SGD optimizer with momentum and  learning rate of $0.1$ for soft attention, hard attention, and $0.3$ for LVML Model. We tuned the learning rate over the search space of $[0.5,0.3,0.1,0.05,0.01]$. We have used the $12$ random seed for parameter initialization for all the experiments. For the hybrid model, we use a learning rate of $0.1$ for soft attention and $0.01$ for hard attention.

\subsection{MSCOCO2014 Experiments}
For the MSCOCO dataset, we use the vocabulary of size $8733$ after removing words that have a frequency less than $5$. We use a VGG19-based encoder with features extracted ($14\times14 \times 512$) from the lower convolutional layer. For attention, we use the hidden size of  512 units for the encoder as well as decoder projection. The decoder is an LSTM model with one hidden layer of $512$ size. We train the soft attention model with Adam Optimizer and a teacher-forcing ratio of  $1$ for $25$ epochs. The initial learning rate for the encoder is $1\mathrm{e}{-5}$  and $1\mathrm{e}-3$ for the decoder. We change the learning rate of the decoder after $6$ epochs to $1\mathrm{e}-4$.

The hard attention model is trained with Adam optimizer with an initial learning rate of $1\mathrm{e}-5$ and $1\mathrm{e}-4$ for the encoder and decoder respectively and the teacher-forcing ratio $0.7$. We update the learning  rate for the decoder to $1\mathrm{e}-5$ after 10 epochs and the teacher-forcing ratio to $1$. The same configuration is used for the hybrid model for both soft and hard attention models. We use $s_y = \max_{t} s_{y}^{t}$, where t is the number of generated words. Each image has multiple categories of objects, for which vocabulary has multiple words, we use different words for a category, as specified in \cite{https://doi.org/10.48550/arxiv.2212.14776} as a word-categorization table. Similarly for $a_z = \sum_{z^* \in z} a_{z}^{t}$, where $t =\argmax_t{s_{y}^{t}}$.

\subsection{Illustration of SDC task}
\label{sdc}
\begin{figure}[ht]
 \centering 
    \includegraphics[width=0.22\columnwidth]{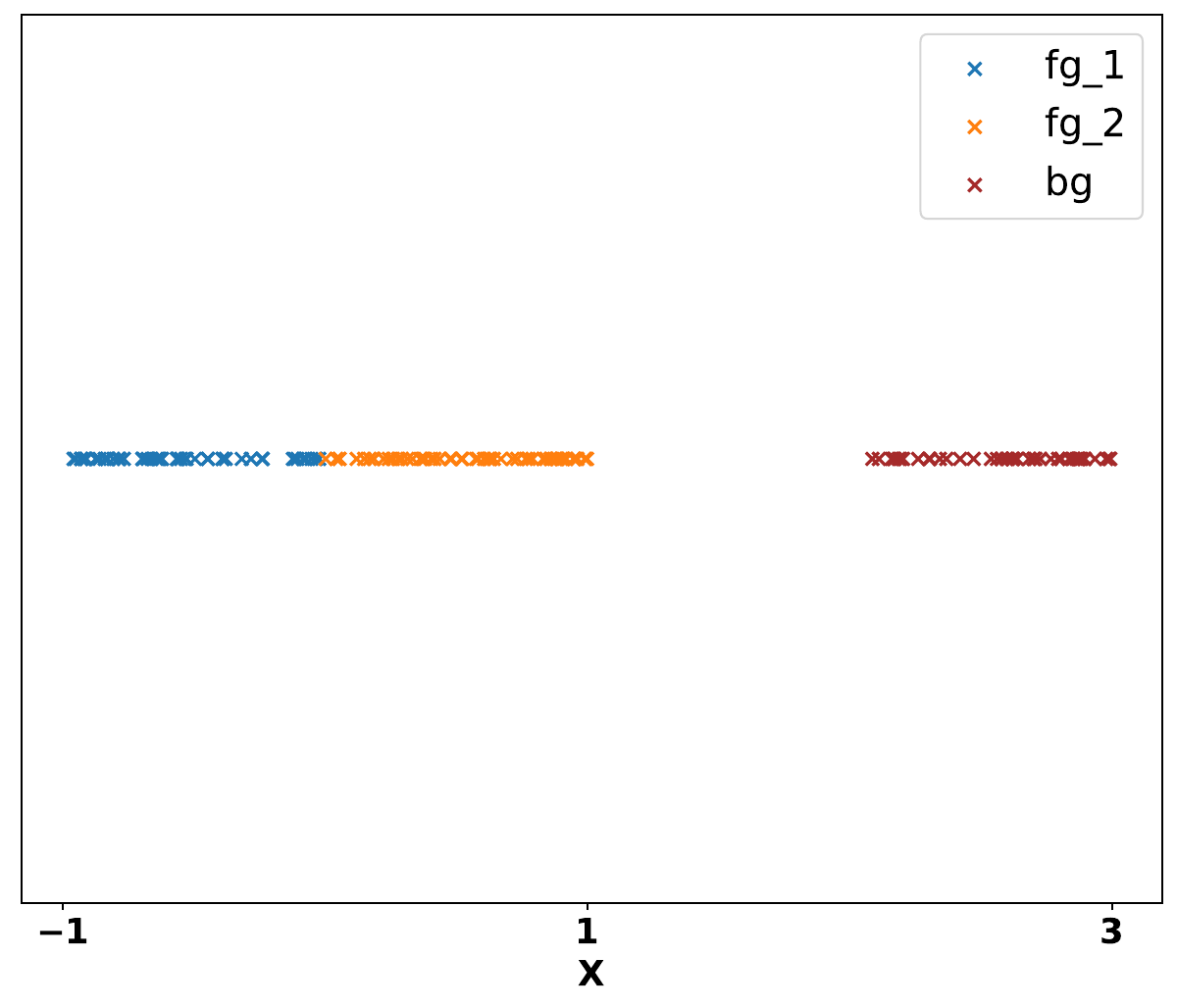} 
    \includegraphics[width=0.24\columnwidth]{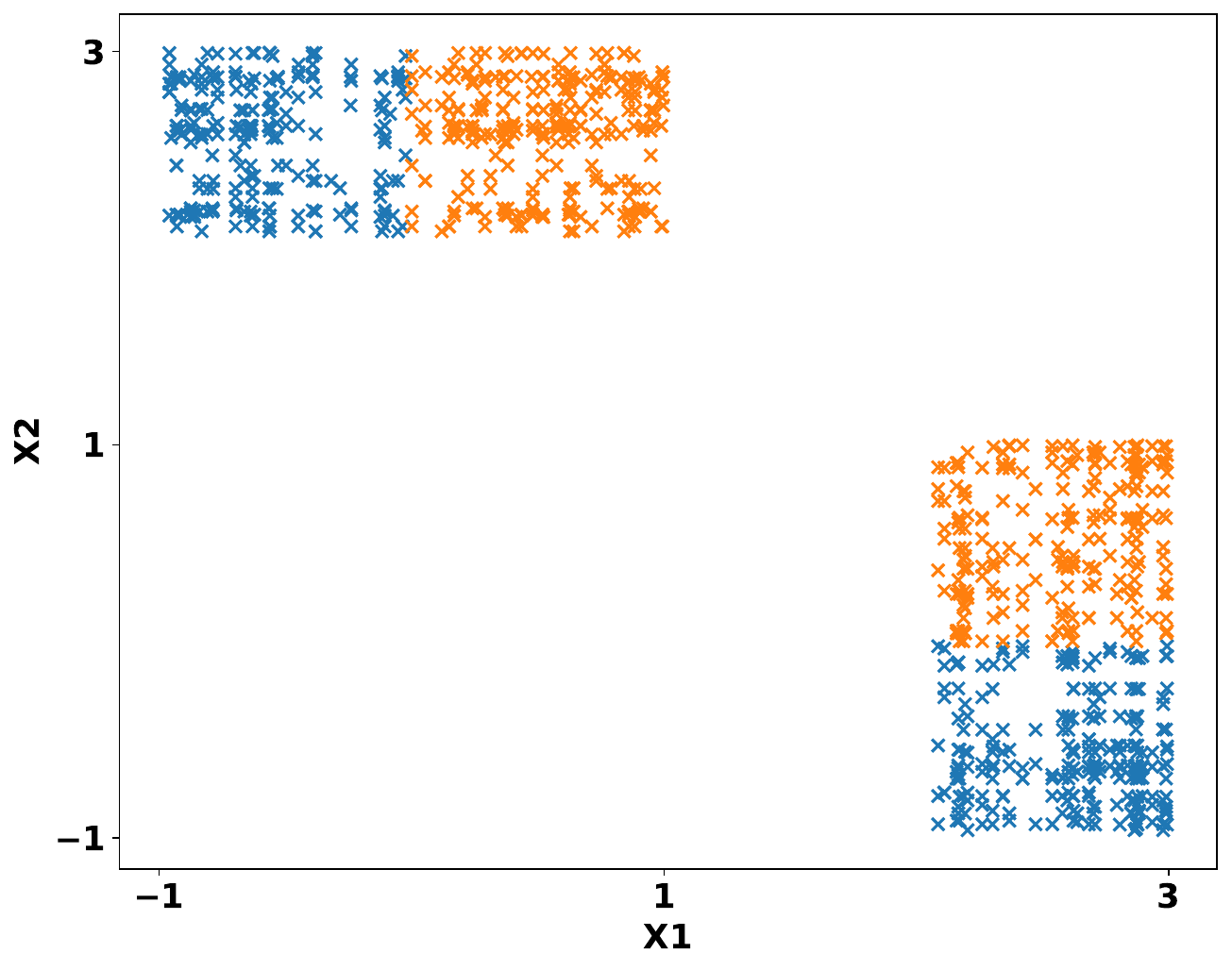} 
    \caption{(\textbf{left}) Sampled data from $D_0$(brown), $D_1$(blue), $D_2$(orange). (\textbf{right}) Mosaic instances.}
    \label{fig Appdx:1}
\end{figure}

We illustrate the SDC task using a $1$-dimensional base distribution with two foreground classes  and one background class. For $m=2$, it results in a mosaic distribution as specified in Figure \ref{fig Appdx:1}. Note the symmetric structure in the scatter plot for the mosaic data,  is due to the swap symmetry, i.e. the foreground segment can be either the first or the second segment \cite{https://doi.org/10.48550/arxiv.2212.14776}.

\subsection{Gradient of Soft Attention Loss with respect to $\bbw_k$ and $\bbu$ }

\begin{align*}
    -L(W,\bbu) &= \frac{1}{C} \sum_{y=1}^{C} \log [ \sigma_y(W\widetilde{\x})], ~~ \textup{where} \widetilde{\x} = \sum_{j=1}^{m}\sigma_j(\bX \bbu)\x_j \\
    -\odv{ L }{ \bbw_k} &=  \frac{1}{C} \sum_{y=1}^{C}\odv{[ \log(\sigma_y(W\widetilde\x))]}{ \bbw_k} \\
-\odv{ L }{ \bbw_k} &=  \frac{1}{C} \sum_{y=1}^{C} \frac{1}{\sigma_y(W\widetilde\x )} \odv{\sigma_y(W\widetilde\x)}{\bbw_k}\\
    -\odv{ L }{ \bbw_k} &=  \frac{1}{C} \sum_{y=1}^{C} \sum_{j=1}^{m}\sigma_j(\bX  \bbu) \left[ \1[y=k] -\sigma_k(W\widetilde\x)  \right]\x_{j}^\top
\end{align*}

Now for gradient with respect to $\bbu$,
\begin{align*}
    -\odv{L}{\bbu} &= -\odv{ L }{ \widetilde\x}  \odv{ \widetilde\x }{ \bbu} \\ 
    -\odv{L}{\widetilde{\x}} &= \dfrac{1}{\sigma_y(W\widetilde{\x})} \odv{\sigma_y(W\widetilde{\x})}{\widetilde{\x}}\\
     -\odv{L}{\widetilde{\x}} &=\frac{1}{C} \sum_{y=1}^{C} \left[ \bbw_{y}^{\top} - \sum_k \sigma_k(W\widetilde{\x})\bbw_{k}^\top \right]\\
     \textup{and, } ~~~ \odv{\widetilde{\x}}{\bbu} &= \odv{\sigma_j(\bX \bbu)}{\bbu} \x_{j}^\top \\
     \odv{\widetilde{\x}}{\bbu} &=  \sum_j \sigma_j(\bX \bbu) [\x_j - \widetilde{\x}] \x_{j}^\top\\
    \textup{Thus} ~~~ -\odv{L}{\bbu} &= \frac{1}{C} \sum_{y=1}^{C}\left[ \bbw_{y}^{\top} - \sum_k \sigma_k(W\widetilde{\x})\bbw_{k}^\top \right] \left[\sum_j \sigma_j(\bX \bbu) [\x_j - \widetilde{\x}] \x_{j}^\top\right] 
\end{align*}

\begin{align}
    -\nabla_{\bbw_k} L &=  \frac{1}{C} \sum_{y=1}^{C} \sum_{j=1}^{m}\sigma_j(\bX  \bbu) \left[ \1[y=k] -\sigma_k(W\widetilde\x )  \right]\x_{j} \label{aeq:1} \\
    -\nabla_{\bbu} L &= \frac{1}{C} \sum_{y=1}^{C}\left[\sum_j \sigma_j(\bX \bbu) \x_j[\x_j - \widetilde{\x}]^\top\right] \left[ \bbw_{y} - \sum_k \sigma_k(W\widetilde{\x} )\bbw_{k} \right]  \label{aeq:2}
\end{align}

\subsection{Gradient of Hard Attention Loss with respect to $\bbw_k$ and $\bbu$ }

\begin{align*}
    -L(W,\bbu) &= \frac{1}{C} \sum_{y=1}^{C}  \sum_{j=1}^{m} \sigma_j(\bX \bbu)\log [ \sigma_y(W \x_j)] 
\end{align*}

\begin{align*}
-\odv{ L }{ \bbw_k} &= \frac{1}{C} \sum_{y=1}^{C} \sum_{j=1}^{m} \sigma_j(\bX \bbu) \frac{1}{\sigma_y(W\x_j)}\odv{\sigma_y(W\x_j)}{\bbw_k}\\
-\odv{ L }{ \bbw_k} &=\frac{1}{C} \sum_{y=1}^{C} \sum_{j=1}^{m} \sigma_j(\bX  \bbu) \left[ \1[y=k] -\sigma_k(W\x_j)  \right]\x_{j}^\top\\
\end{align*}

Now for gradient with respect to $\bbu$,

\begin{align*}
  -\odv{ L }{ \bbu} &=  \frac{1}{C} \sum_{y=1}^{C}  \sum_{j=1}^{m} \log [ \sigma_y(W \x_j)] \odv{\sigma_j(\bX \bbu)}{\bbu}\\
  -\odv{ L }{ \bbu} &= \frac{1}{C} \sum_{y=1}^{C}  \sum_{j=1}^{m} \log [ \sigma_y(W \x_j)] \sigma_j(\bX \bbu) (\x_j-\widetilde\x)^\top
\end{align*}

\begin{align}
    -\nabla_{\bbw_k} L &=  \frac{1}{C} \sum_{y=1}^{C} \sum_{j=1}^{m}\sigma_j(\bX  \bbu) \left[ \1[y=k] -\sigma_k(W\x_j)  \right]\x_{j} \label{aeq:3} \\
    -\nabla_{\bbu} L &= \frac{1}{C} \sum_{y=1}^{C}\sum_{j=1}^{m} \log [ \sigma_y(W \x_j)] \sigma_j(\bX \bbu) (\x_j-\widetilde\x) \label{aeq:4}
\end{align}

\subsection{Gradient of LVML with respect to $\bbw_k$ and $\bbu$ }

\text{Proof:}
\begin{align*}
    -L(W,\bbu) &= \frac{1}{C} \sum_{y=1}^{C} \log \sum_{j=1}^{m}\sigma_j(\bX \bbu)[ \sigma_y(W\x_j)] 
\end{align*}

\begin{align*}
-\odv{ L }{ \bbw_k} &=  \frac{1}{C} \sum_{y=1}^{C} \dfrac{1}{\sum_j \sigma_j(\bX \bbu) \sigma_y(W\x_j)} \sum_{j=1}^{m} \sigma_j(\bX \bbu) \odv{\sigma_y(W\x_j)}{\bbw_k} \\
-\odv{ L }{ \bbw_k} &=  \frac{1}{C} \sum_{y=1}^{C} \dfrac{\sum_j \sigma_j(\bX \bbu) \sigma_y(W\x_j)   \left[ \1[y=k] -\sigma_k(W\x_j)  \right]\x_{j}^\top }{\sum_{j'} \sigma_{j'}(\bX \bbu) \sigma_y(W\x_{j'}) }\\
-\odv{ L }{ \bbw_k} &=  \frac{1}{C} \sum_{y=1}^{C} \sum_{j=1}^{m} \gamma_j \left[ \1[y=k] -\sigma_k(W\x_j)  \right]\x_{j}^\top
\end{align*}

Now for gradient with respect to $\bbu$,
\begin{align*}
-\odv{ L }{ \bbu} &=  \frac{1}{C} \sum_{y=1}^{C} \dfrac{1}{\sum_j \sigma_j(\bX \bbu) \sigma_y(W\x_j)} \sum_{j=1}^{m} \sigma_y(W\x_j) \odv{\sigma_j(\bX \bbu)}{\bbu}\\
-\odv{ L }{ \bbu} &= \frac{1}{C} \sum_{y=1}^{C}  \dfrac{\sum_j \sigma_y(W\x_j) \sigma_j(\bX \bbu)(\x_j - \widetilde\x)^\top}{\sum_j \sigma_j(\bX \bbu) \sigma_y(W\x_j)}\\
-\odv{ L }{ \bbu} &= \frac{1}{C} \sum_{y=1}^{C} \sum_j \gamma_j(\x_j - \widetilde\x)^\top, \textup{where} ~~ \gamma_j = \dfrac{\sigma_j(\bX \bbu)\sigma_y(W\x_j)}{\sum_{j'} \sigma_{j'}(\bX \bbu)\sigma_y(W\x_{j'})}
\end{align*}

\begin{align}
    -\nabla_{\bbw_k} L &=  \frac{1}{C} \sum_{y=1}^{C} \sum_{j=1}^{m}\gamma_j \left[ \1[y=k] -\sigma_k(W\x_j)  \right]\x_{j}\label{aeq:5} \\
    -\nabla_{\bbu} L &= \frac{1}{C} \sum_{y=1}^{C}\sum_{j=1}^{m}  \gamma_j(\x_j - \widetilde\x) \label{aeq:6}
\end{align}

\subsection{Update Expression for Focus and Classification Module Parameters for Different Attention Paradigms}
\label{p1}
\begin{lemma} Classification module parameter in soft attention follows the equation \ref{eq1} 
   \[\odv{ \mu^{\textup{SA}}(t)}{t} =  \dfrac{\alpha }{\exp( \alpha \mu^{\textup{SA}}(t))+C-1}\] 
\end{lemma}

\text{Proof:}
Without loss of generality,  assuming first patch to be foreground. We apply the orthogonality condition and consider background mean to be zero, equation \eqref{aeq:1} simplies to
Consider $\bbw_k$ and $\bbu$ at some arbitrary time step t, for fixed $\alpha$
\[ -\nabla_{\bbw_k[t]} L = \frac{\alpha}{C}[\bbs_k - \sum_{k'} \sigma_k (W(t)\alpha \bbs_{k'})\bbs_{k'}] \]
Considering the assumption
\[ \bbw_k[t] = \mu^{\textup{SA}}(t) [\bbs_k - \dfrac{1}{C} \sum_{k'} \bbs_{k'} ]\]
Now using the orthogonality condition for background and foreground segment we get,
\[W(t)\bbs_k = \mu^{\textup{SA}}(t) [\e_k-\dfrac{1}{C}\1]\]

now using softmax property, $\sigma_j(z+c) = \sigma_j(z)$ and above equation, we get

\[ -\nabla_{\bbw_k[t]} L = \frac{\alpha}{C}[\bbs_k - \sum_{k'} \sigma_k (\alpha\mu^{\textup{SA}}(t)\e_{k'})\bbs_{k'}]\]

Simplifying this we get,
\[ -\nabla_{\bbw_k[t]} L = \dfrac{C\alpha}{C \exp(\alpha\mu^{\textup{SA}}(t))+C-1}[\bbs_k - \dfrac{1}{C} \sum_{k'} \bbs_{k'} ] \]

Thus,

\[\dfrac{\partial \mu^{\textup{SA}}(t)}{\partial t} = \dfrac{\alpha}{\exp(\alpha\mu^{\textup{SA}}(t))+C-1}\] \\

For Simultaneous updates $\alpha^{\textup{SA}}(t)$ follows, 

  Now using assumption $\bbu[t] = \nu^{\textup{SA}}(t) \sum_{k=1}^c \bbs_k$, we can write 
  \[\sigma_1(\bX \bbu[t] )  = \dfrac{\exp(\nu^{\textup{SA}}(t) \lvert\rvert s_y \lvert\rvert^{2})}{\exp(\nu^{\textup{SA}}(t)\lvert\rvert s_y \lvert\rvert^{2}) +m-1} \]
  
  \[\sigma_1(\bX \bbu[t] )  = \dfrac{\exp(\nu^{\textup{SA}}(t) )}{\exp(\nu^{\textup{SA}}(t)) +m-1} = \alpha^{\textup{SA}}(t)\]
\begin{align*}
    \alpha^{\textup{SA}}(t) = \dfrac{\exp(\nu^{\textup{SA}}(t))}{\exp(\nu^{\textup{SA}}(t)) + m-1}
\end{align*}

\begin{lemma} Focus module parameter in soft attention follows the equation \ref{eq4} 
   \[ \odv{\nu^{\textup{SA}}(t)}{t} = \dfrac{ \mu^{\textup{SA}}(t)(C-1) (\alpha^{\textup{SA}}(t)-(\alpha^{\textup{SA}}(t))^2) }{C(\exp(\alpha^{\textup{SA}}(t) \mu^{\textup{SA}}(t))+C-1)}  \] 
\end{lemma}

\text{Proof:}
Consider equation \eqref{aeq:2} for $\nabla_U L$, Without loss of generality, we assume first patch to be foreground patch. 
\begin{align*}
    -\nabla_{\bbu} L &=\frac{1}{C} \sum_{y=1}^{C} \left[\sum_j \sigma_j(\bX \bbu) \x_j[\x_j - \widetilde{\x}]^\top\right] \left[ \bbw_{y} - \sum_k \sigma_k(W\widetilde{\x})\bbw_{k} \right]  
\end{align*}

 Simplifying $W_y - \sum_k \sigma_k(W\widetilde{\x} +b  )W_k$, we get
 \[ \dfrac{C W_y}{\exp(\sigma_1(\bX U)\mu^{\textup{SA}}(t))+C-1}-\sum_k\dfrac{W_k}{\exp(\sigma_1(\bX U)\mu^{\textup{SA}}(t))+C-1}\]
 
 Now combing this with simplified above term and applying orthogonality assumption, we get,
 \begin{align*}
 -\nabla_\bbu L  &= \frac{1}{C}\sum_{y=1}^{C} [\sigma_1(\bX \bbu) - \sigma_1^2(\bX \bbu) ] \bbs_y \bbs_y^\top  \Bigg[ \dfrac{C \bbw_y}{\exp(\sigma_1(\bX \bbu)\mu^{\textup{SA}}(t))+C-1}  -  \sum_k\dfrac{\bbw_k}{\exp(\sigma_1(\bX \bbu)\mu^{\textup{SA}}(t))+C-1} \Bigg]
 \end{align*}

 Considering the assumptions 
 \[ \bbw_k(t) = \mu^{\textup{SA}}(t) [\bbs_k - \dfrac{1}{C} \sum_{k'} 
 \bbs_{k'}]\]

  \begin{align*}
    -\nabla_\bbu L  &= \frac{1}{C}\sum_{y=1}^{C} [\sigma_1(\bX \bbu) - \sigma_1^2(\bX \bbu) ] \bbs_y \bbs_y^\top  \Bigg[ \dfrac{C\mu^{\textup{SA}}(t) [\bbs_y - \dfrac{1}{C} \sum_{k'} \bbs_{k'}]}{\exp(\sigma_1(\bX \bbu)\mu^{\textup{SA}}(t))+C-1}  -\sum_k\dfrac{\mu^{\textup{SA}}(t) [\bbs_k - \dfrac{1}{C} \sum_{k'} \bbs_{k'}]}{\exp(\sigma_1(\bX \bbu)\mu^{\textup{SA}}(t))+C-1} \Bigg] 
  \end{align*}

  \[ -\nabla_\bbu L = \frac{1}{C}\sum_{y=1}^{C} [\sigma_1(\bX U) - \sigma_1^2(\bX U) ] \bbs_y \bbs_y^\top \Bigg[ \dfrac{C\mu^{\textup{SA}}(t) [\bbs_y - \dfrac{1}{C} \sum_{k'} \bbs_{k'}]}{\exp(\sigma_1(\bX U)\mu^{\textup{SA}}(t))+C-1} \Bigg] \]

  \[-\nabla_\bbu L =  \dfrac{ \sigma_1(\bX \bbu) - \sigma_1^2(\bX \bbu)  \mu^{\textup{SA}}(t)}{\exp(\sigma_1(\bX \bbu )\mu^{\textup{SA}}(t))+c-1}\sum_{y=1}^{c}  \bbs_y \bbs_y^\top \Bigg[ [\bbs_y - \dfrac{1}{c} \sum_{k'} \bbs_{k'}] \Bigg] \]

  \[ -\nabla_\bbu L = \dfrac{ \sigma_1(\bX \bbu) - \sigma_1^2(\bX \bbu)  \mu^{\textup{SA}}(t)}{\exp(\sigma_1(\bX \bbu)\mu^{\textup{SA}}(t))+C-1}\sum_{y=1}^{C}  \bbs_y \Bigg[ [\bbs_y^\top  \bbs_y - \dfrac{1}{C} \bbs_y^\top \sum_{k'} \bbs_{k'}] \Bigg] \]

  Using the assumption that $s_y^\top s_y  =1$ and $s_y^\top s_{y'}  =0$ $ \forall y,y'$,we get 
  
  \[ -\nabla_\bbu L =  \dfrac{ (C-1)\sigma_1(\bX \bbu) - \sigma_1^2(\bX \bbu) \mu^{\textup{SA}}(t)}{C(\exp(\sigma_1(\bX \bbu)\mu^{\textup{SA}}(t))+C-1)}\sum_{y=1}^{C}  \bbs_y \]

  Now using assumption $\bbu[t] = \nu^{\textup{SA}}(t) \sum_{k=1}^c \bbs_k$, we can write 
  \[\sigma_1(\bX \bbu[t] )  = \dfrac{\exp(\nu^{\textup{SA}}(t) \lvert\rvert s_y \lvert\rvert^{2})}{\exp(\nu^{\textup{SA}}(t)\lvert\rvert s_y \lvert\rvert^{2}) +m-1} \]
  
  \[\sigma_1(\bX \bbu[t] )  = \dfrac{\exp(\nu^{\textup{SA}}(t) )}{\exp(\nu^{\textup{SA}}(t)) +m-1} = \alpha^{\textup{SA}}(t)\]
  
  \[ \therefore -\nabla_\bbu L = \dfrac{\alpha^{\textup{SA}}(t)(C-1)(1-\alpha^{\textup{SA}}(t)) \mu^{\textup{SA}}(t)}{C(\exp(\alpha^{\textup{SA}}(t)\mu^{\textup{SA}}(t)))+C-1}\sum_{y=1}^{C}  \bbs_y \]

  \[  \odv{ \nu^{\textup{SA}}(t) }{t} =   \dfrac{\alpha^{\textup{SA}}(t)(C-1)(1-\alpha^{\textup{SA}}(t)) \mu^{\textup{SA}}(t)}{C(\exp(\alpha^{\textup{SA}}(t)\mu^{\textup{SA}}(t)))+C-1}\]

  \begin{lemma} Classification module parameter in hard attention follows the equation \ref{eq2} 
   \[ \odv{ \mu^{\textup{HA}}(t)}{t} = \dfrac{\alpha \beta^{\textup{HA}}(t)}{\exp(\mu^{\textup{HA}}(t))}  \] 
\end{lemma}
  \text{Proof:}

  Consider equation \eqref{aeq:3} for $\nabla_{\bbw_k} L$ for fixed $\alpha$, Without loss of generality, we assume first patch to be foreground patch. Applying orthogonality assumption, we get,
\[ -\nabla_{\bbw_k[t]} L = \frac{\alpha}{C}[\bbs_k - \sum_{k'} \sigma_k (W(t)\bbs_{k'})\bbs_{k'}] \]

By assumption, we have
\[ \bbw_k[t] = \mu^{\textup{HA}}(t) [\bbs_k - \dfrac{1}{C} \sum_{k'} \bbs_{k'} ]\]

Now using the orthogonality condition for background and foreground segment we get,
\[W(t)\bbs_k = \mu^{\textup{HA}}(t) [\e_k-\dfrac{1}{C}\1]\]

\[ \therefore -\nabla_{\bbw_k[t]} L=\frac{\alpha}{C}[\bbs_k - \sum_{k'} \sigma_k(\mu^{\textup{HA}}(t) [e_{k'} - \frac{1}{C}\1])\bbs_{k'}]  \]

now using softmax property, $\sigma_j(z+c) = \sigma_j(z)$ and above equation, we get 
\[   -\nabla_{\bbw_k[t]} L=\frac{\alpha}{C}[\bbs_k - \sum_{k'} \sigma_k(\mu^{\textup{HA}}(t) e_{k'})\bbs_{k'}] \]

After simplifying we can write this as follow
\[-\nabla_{\bbw_k[t]} L =  \frac{\alpha}{C} \left[\bbs_k  \bigg[ \dfrac{C}{\exp(\mu^{\textup{HA}}(t))+C-1} \bigg] - \sum_{k'}\dfrac{1}{\exp(\mu^{\textup{HA}}(t))+C-1}\bbs_{k'} \right])\]

\[-\nabla_{\bbw_k[t]} L =  \frac{\alpha C}{C\exp(\mu^{\textup{HA}}(t))+C-1} \left[\bbs_k  - \frac{1}{C}\sum_{k'}\bbs_{k'} \right]\]

Hence,
\begin{align*}
 \dfrac{\partial \mu^{\textup{HA}}(t) }{\partial t}  &= \dfrac{\alpha }{\exp(\mu^{\textup{HA}}(t))+C-1}  \\
  \dfrac{\partial \mu^{\textup{HA}}(t)}{\partial t} &= \dfrac{\alpha \beta^{\textup{HA}}(t) }{\exp(\mu^{\textup{HA}}(t))}
  \end{align*}
 
 where, $\beta^{\textup{HA}}(t) = \dfrac{\exp(\mu^{\textup{HA}}(t))}{\exp(\mu^{\textup{HA}}(t))+C-1}$. \\
 For Simultaneous updates $\alpha^{\textup{HA}}(t)$ follows, 
\begin{align*}
    \alpha^{\textup{HA}}(t) = \dfrac{\exp(\nu^{\textup{HA}}(t))}{\exp(\nu^{\textup{HA}}(t)) + m-1}
\end{align*}

  \begin{lemma} Focus module parameter in hard attention follows the equation \ref{eq5} 
   \[  \odv{\nu^{\textup{HA}}(t)}{t} = \frac{\log[C \beta^{\textup{HA}}(t)]}{C}(\alpha^{\textup{HA}}(t)-(\alpha^{\textup{HA}}(t))^2)\] 
\end{lemma}

\text{Proof:}

   Consider equation \eqref{aeq:3} for $\nabla_{\bbu} L$, Without loss of generality, we assume first patch to be foreground patch. 
  
\begin{align*}
    -\nabla_U L =  \frac{1}{C} \sum_{y=1}^{C} \sum_{j=1}^{m} 
    \log [\sigma_y(W \x_j)] \sigma_j(\bX U)(\x_j - \widetilde{\x})
\end{align*}
Simplifying it using orthogonality assumption and condition that background mean is zero, 

We get 

\begin{align*}
 -\nabla_U L = \frac{1}{C} \sum_{y=1}^{C} \log[\sigma_y(W\x_1)] \sigma_1(\bX U) \Bigg[1 -\sigma_1(\bX U) \Bigg] \x_1 - \sigma_1{(\bX U)} \log[{\frac{1}{C}}]  (1-\sigma_1(\bX U))\x_1
\end{align*}
\begin{align*}
 \sigma_y(W\x_1) &= \dfrac{\exp(\mu^{\textup{HA}}(t))}{\exp(\mu^{\textup{HA}}(t))+m-1} =\beta^{\textup{HA}}(t) \\
\sigma_1(\bX U) &= \dfrac{\exp(\nu^{\textup{HA}}(t))}{\exp(\nu^{\textup{HA}}(t))+m-1} =\alpha^{\textup{HA}}(t)
\end{align*}

we get, 
\begin{align*}
     -\nabla_U L =\frac{1}{C} \log[\beta^{\textup{HA}}(t)] \alpha^{\textup{HA}}(t) &-  
     \alpha^{\textup{HA}}(t)  \Bigg[   \log[\beta^{\textup{HA}}(t)] \alpha^{\textup{HA}}(t)  + \log[\frac{1}{C}] (1-\alpha^{\textup{HA}}(t))\Bigg] \sum_{y=1}^{C} \bbs_y
\end{align*}

Hence, 
\begin{align*}
 \frac{\partial{\nu^{\textup{HA}}(t)}}{\partial{t}} =  \frac{1}{C} \Bigg[ \log[\beta^{\textup{HA}}(t)] \alpha^{\textup{HA}}(t) &-  
     \alpha^{\textup{HA}}(t)  \Bigg[   \log[\beta^{\textup{HA}}(t)] \alpha^{\textup{HA}}(t)  + \log[\frac{1}{C}] (1-\alpha^{\textup{HA}}(t))\Bigg] \Bigg]
\end{align*}
\begin{align*}
\dfrac{\partial \nu^{\textup{HA}}(t)}{\partial t} = \frac{\alpha^{\textup{HA}}(t)(1-\alpha^{\textup{HA}}(t))}{C} \bigg[ \log[\beta^{\textup{HA}}(t)] + \log[C] \bigg]
\end{align*}

\begin{align*}
\dfrac{\partial \nu^{\textup{HA}}(t)}{\partial t} = \frac{\log[\beta^{\textup{HA}}(t) C] \alpha^{\textup{HA}}(t)(1-\alpha^{\textup{HA}}(t))}{C} \
\end{align*}

  \begin{lemma} Classification module parameter in latent variable model for attention follows the equation \ref{eq3} 
   \[ \odv{ \mu^{\textup{LV}}(t)}{t} =  \dfrac{\alpha (\beta^{\textup{LV}}(t))^2}{Z(t)\exp(\mu^{\textup{LV}}(t))} \] 
\end{lemma}

\text{Proof:}

  Consider equation \eqref{aeq:6} for $\nabla_{\bbw_k} L$ for fixed $\alpha$, Without loss of generality, we assume first patch to be foreground patch. Applying orthogonality assumption, we get,
\[ -\nabla_{\bbw_k(t)} L =\frac{1}{C} \left[ \gamma_1 \bbs_k -\sum_{y=1}^C \sigma_k(W\bbs_y) \gamma_1 \bbs_y\right] \]

where, 
\begin{align*}
 \gamma_1 &=   \dfrac{\alpha \sigma_y(W\bbs^y)}{\alpha \sigma_y(W\bbs^y)+\dfrac{1-\alpha}{c}} = \dfrac{\alpha \sigma_1(\mu(t)\e^c_1)}{\alpha \sigma_1(\mu(t)\e^c_1)+\dfrac{1-\alpha}{c}} =   \dfrac{\alpha \sigma_1(\mu(t)\e^c_1)}{Z(t)} 
\end{align*}

By assumption, we have
\[ \bbw_k[t] = \mu^{\textup{LV}}(t) [\bbs_k - \dfrac{1}{C} \sum_{k'} \bbs_{k'} ]\]

\[\therefore  W[t]\bbs^y = \mu^{\textup{LV}}(t) [e_y -\frac{1}{C}\1] \]

\[  -\nabla_{W_k[t]} L = \frac{\alpha}{CZ(t)} \Bigg[ \sigma_k(W\bbs_k) \bbs_k -\sum_{y=1}^{c} \sigma_y(W\bbs_y) \sigma_k(W\bbs_y) \bbs_y  \Bigg]  \]


Now Simplifying and using the softmax property $\sigma_j(\z+c) = \sigma_j(\z)$, we have

\begin{align*}  -\nabla_{\bbw_k(t)} L = \dfrac{\alpha}{CZ(t)} \Bigg[ \sigma_k(\mu^{\textup{LV}}(t) e_k ) \bbs_k - \sum_{y=1}^{c} \sigma_y(\mu^{\textup{LV}}(t) e_y )  \sigma_k(\mu^{\textup{LV}}(t) e_y ) \bbs_y  \Bigg] \end{align*}

\begin{align*}  -\nabla_{\bbw_k(t)} L = \dfrac{\alpha}{CZ(t)} \Bigg[ \sigma_k(\mu^{\textup{LV}}(t) e_k ) \bbs_k -\sigma_k(\mu^{\textup{LV}}(t) e_k ) \sigma_k(\mu^{\textup{LV}}(t) e_k ) \bbs_k -  \sum_{y=1,y\neq k}^{c} \sigma_y(\mu^{\textup{LV}}(t) e_y )  \sigma_k(\mu^{\textup{LV}}(t) e_y ) \bbs_y  \Bigg] \end{align*}

Now consider,

\[\sigma_k(\mu^{\textup{LV}}(t)e_k) = \dfrac{\exp(\mu^{\textup{LV}}(t))}{\exp(\mu^{\textup{LV}}(t))+C-1} = \beta^{\textup{LV}}(t) \]

\begin{align*}
     = \dfrac{\alpha}{CZ(t)} \Bigg[\beta^{\textup{LV}}(t) \bbs_k -  (\beta^{\textup{LV}}(t))^2 \bbs_k  -   \sum_{y=1,y\neq k}^{c}\dfrac{\beta^{\textup{LV}}(t)}{\exp(\mu^{\textup{LV}}(t))+C-1} \bbs_y\Bigg]
\end{align*}

\begin{align*}
    = \dfrac{\alpha}{CZ(t)} \Bigg[ \beta^{\textup{LV}}(t) \dfrac{C-1}{\exp(\mu^{\textup{LV}}(t))+C-1} \bbs_k  -  \sum_{y=1,y\neq k}^{c}\dfrac{\beta^{\textup{LV}}(t)}{\exp(\mu^{\textup{LV}}(t))+C-1} \bbs_y\Bigg]
\end{align*}

\begin{align*}
     -\nabla_{\bbw_{k(t)}} L = \dfrac{\alpha}{CZ(t)} \Bigg[\dfrac{C \beta^{\textup{LV}}(t)\bbs_k}{\exp(\mu^{\textup{LV}}(t))+C-1} - \dfrac{\beta^{\textup{LV}}(t)\bbs_k}{\exp(\mu^{\textup{LV}}(t))+C-1} -   \sum_{y=1,y\neq k}^{c}\dfrac{\beta^{\textup{LV}}(t)}{\exp(\mu^{\textup{LV}}(t))+C-1} \bbs_y\Bigg]
\end{align*}

\begin{align*}
     -\nabla_{\bbw_{k(t)}} L= \dfrac{\alpha}{CZ(t)} \Bigg[\dfrac{C \beta^{\textup{LV}}(t)\bbs_k}{\exp(\mu^{\textup{LV}}(t))+C-1} - \sum_{y=1}^{c}\dfrac{\beta^{\textup{LV}}(t)}{\exp(\mu^{\textup{LV}}(t))+C-1} \bbs_y\Bigg]
\end{align*}

\begin{align*}
   -\nabla_{\bbw_{k(t)}} L = \dfrac{\alpha}{CZ(t)} \Bigg[ \dfrac{ \beta^{\textup{LV}}(t)}{\exp(\mu^{\textup{LV}}(t))+C-1} \Bigg[ C\bbs_k -  \sum_{y=1}^{C} \bbs_y\Bigg]\Bigg]
\end{align*}

\begin{align*}
 -\nabla_{W_k[t]} L =  \dfrac{\alpha}{CZ(t)} \Bigg[\dfrac{ C \beta^{\textup{LV}}(t)}{\exp(\mu^{\textup{LV}}(t))+C-1} \Bigg[ \bbs_k -  \frac{1}{C}\sum_{y=1}^{C} \bbs_y\Bigg]\Bigg]
\end{align*}

\begin{align*}
\therefore
   \dfrac{\partial \mu(t) }{\partial t} =   \dfrac{  \alpha(\beta^{\textup{LV}}(t))^2}{Z(t)\exp(\mu^{\textup{LV}}(t))} 
\end{align*}

where $Z(t) = \alpha\beta^{\textup{LV}}(t) + \dfrac{1-\alpha}{C}$ \\

For Simultaneous updates $\alpha^{\textup{LV}}(t)$ follows, 
\begin{align*}
    \alpha^{\textup{LV}}(t) = \dfrac{\exp(\nu^{\textup{LV}}(t))}{\exp(\nu^{\textup{LV}}(t)) + m-1}
\end{align*}

  \begin{lemma} Focus module parameter in latent variable model for attention follows the equation \ref{eq6} 
   \[ \odv{\nu^{\textup{LV}}(t)}{t} = \frac{\alpha^{\textup{LV}}(t)}{C}\bigg[\dfrac{\beta^{\textup{LV}}(t) }{Z(t)} - 1\bigg] \] 
\end{lemma}

\text{Proof:}
   Consider equation \eqref{aeq:5} for $\nabla_{\bbu} L$, Without loss of generality, we assume first patch to be foreground patch $\bbs_y$. we get, 
\begin{align*} 
-\nabla_U  L = \frac{1}{C}\sum_{y=1}^C\sigma_1(\bX U) \Bigg[\dfrac{\sigma_y(W\x_1)}{Z} - 1 \Bigg]\x_1  +\sum_{y=1}^C \dfrac{1-\sigma_1(\bX U)}{m-1}\Big(\dfrac{1}{CZ} -1\Big) \sum_{j=2}^m \x_j \end{align*}

where, $Z = \sum_{j'} \sigma_{j'}(\bX \bbu)\sigma_y(W\x_{j'})$

Using the condition background mean is zero, and orthogonality assumption we get,

\[-\nabla_U  L = \frac{1}{C}\sum_{y=1}^C \sigma_1(\bX U) \Bigg[\dfrac{\sigma_y(W\bbs_y)}{Z} - 1 \Bigg]\bbs_y  \]

\[-\nabla_U  L = \frac{\sigma_1(\bX U)}{C}  \sum_{y=1}^C  \Bigg[\dfrac{\sigma_y(\mu^{\textup{LV}}(t)[\e_y - \frac{1}{C}\1])}{Z} - 1 \Bigg]\bbs_y  \]

now using softmax property, $\sigma_j(z+c) = \sigma_j(z)$ and above equation, we get 

\[-\nabla_U  L =  \frac{\sigma_1(\bX U)}{C}  \sum_{y=1}^C  \Bigg[\dfrac{\sigma_y(\mu^{\textup{LV}}(t)\e_y )}{Z} - 1 \Bigg]\bbs_y  \]

\[-\nabla_U  L =  \frac{\sigma_1(\bX U)}{C}  \sum_{y=1}^C  \Bigg[\dfrac{\exp(\mu^{\textup{LV}}(t)) }{Z(\exp(\mu^{\textup{LV}}(t))+C-1)} - 1 \Bigg]\bbs_y  \]

Using the assumption, $U(t) = \nu^{\textup{LV}}(t)\sum_{k=1}^{c}s_k$ Hence, 
\[ \sigma_1(\bX U) = \dfrac{\exp(\nu^{\textup{LV}}(t))}{\exp(\nu^{\textup{LV}}(t))+m-1} = \alpha^{\textup{LV}}(t),~~~ 
 \dfrac{\exp(\mu^{\textup{LV}}(t))}{\exp(\nu^{\textup{LV}}(t))+C-1} = \beta^{\textup{LV}}(t)
\]

\[-\nabla_U  L =  \frac{ \alpha^{\textup{LV}}(t)}{C} \sum_{y=1}^c  \Bigg[\dfrac{\beta^{\textup{LV}}(t) }{Z(t)} - 1 \Bigg]\bbs_y  \]

\[-\nabla_U  L = \frac{ \alpha^{\textup{LV}}(t)}{C} \Bigg[\dfrac{\beta^{\textup{LV}}(t) }{Z(t)} - 1 \Bigg] \sum_{y=1}^C \bbs_y  \]

\[\therefore \odv{ \nu^{\textup{LV}}(t)}{ t} =\frac{ \alpha^{\textup{LV}}(t)}{C} \Bigg[\dfrac{\beta^{\textup{LV}}(t) }{Z(t)} - 1 \Bigg] \]

where $Z(t) = \alpha(t)\beta(t) + \dfrac{1-\alpha(t)}{C}$

\subsection{Focus Prediction Heat maps for different Settings}
\label{more_results}
Heat maps for all the settings for CIFAR10, CIFAR100, HateXplain, and MSCOCO dataset  are given in this section.

 \begin{figure*}[!ht]
  \centering
    \begin{subfigure}[b]{0.32\textwidth}
      \centering
      \includegraphics[width=\textwidth]{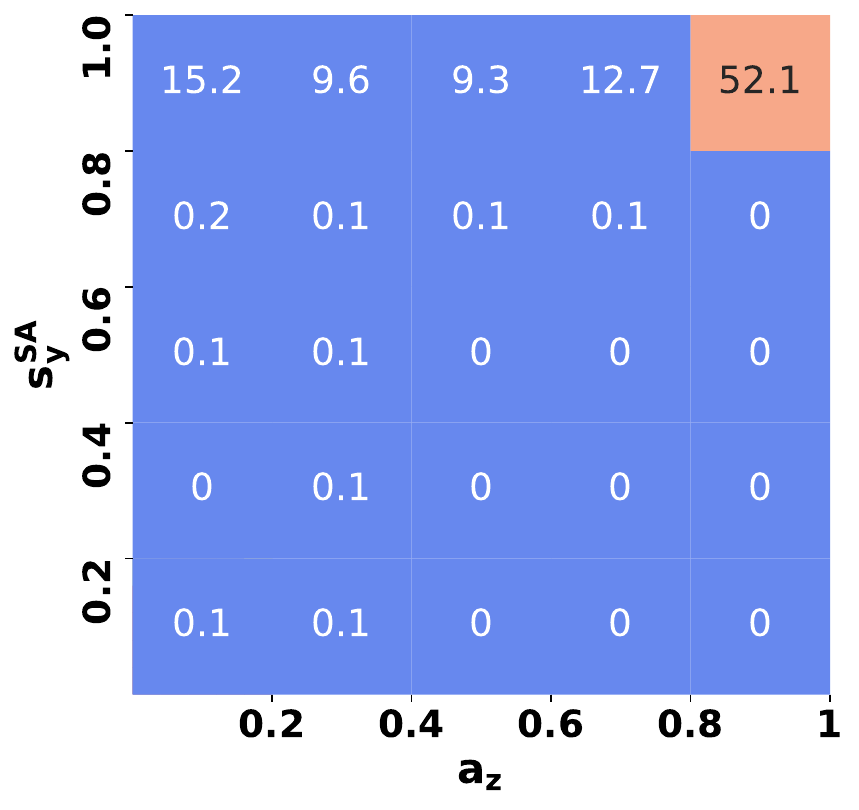}
      \caption{Soft Attention train data }
        \label{fig5:subfig1}
    \end{subfigure}
    \begin{subfigure}[b]{0.32\textwidth}
      \centering
    \includegraphics[width=\textwidth]{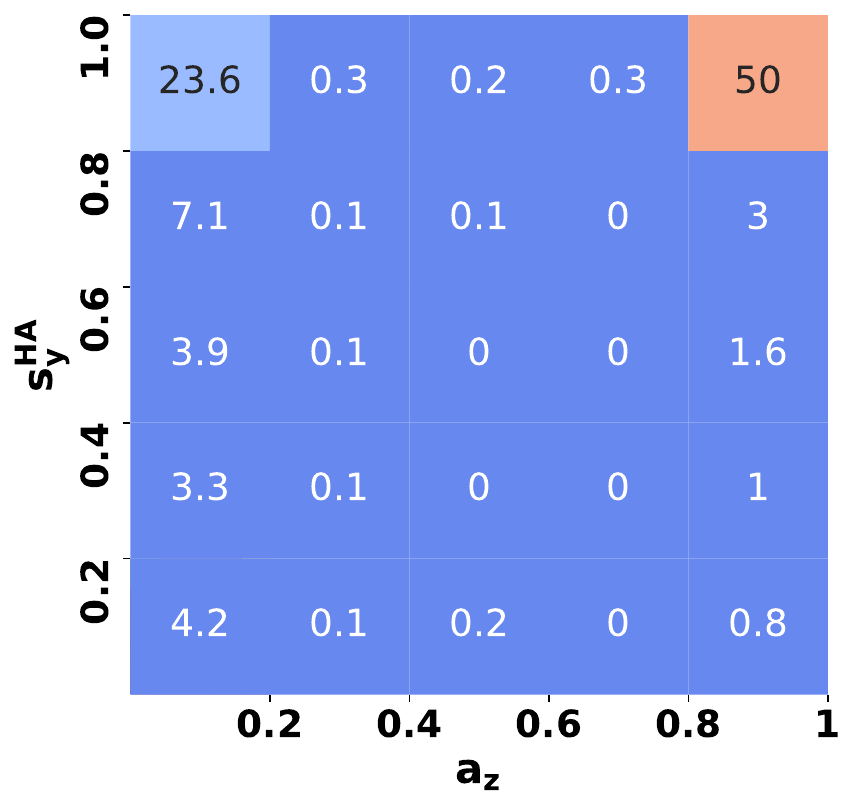}
      \caption{Hard Attention train data}
        \label{fig5:subfig2}
    \end{subfigure}
  \begin{subfigure}[b]{0.32\textwidth}
      \centering
    \includegraphics[width=\textwidth]{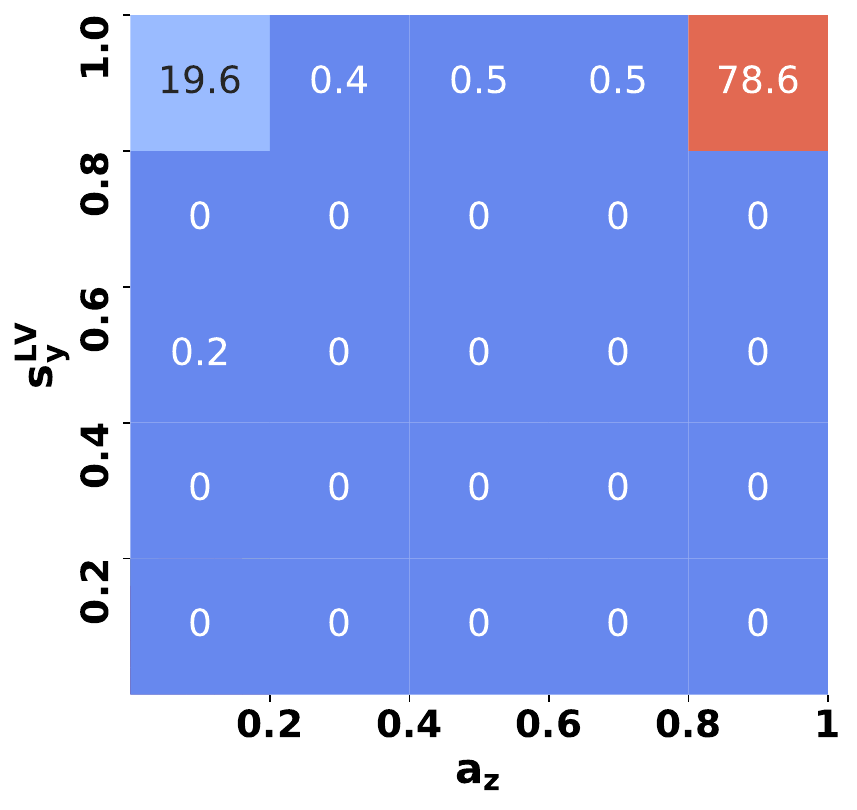}
      \caption{LVML train data}
        \label{fig5:subfig3}
    \end{subfigure}
    \caption{Focus-Prediction heat maps of train data for the three attention paradigms on the CIFAR10 SDC task with $n=10000$ and $m=5$.}
    \label{fig5:my_label}
\end{figure*}

 \begin{figure*}[!ht]
  \centering
    \begin{subfigure}[b]{0.32\textwidth}
      \centering
      \includegraphics[width=\textwidth]{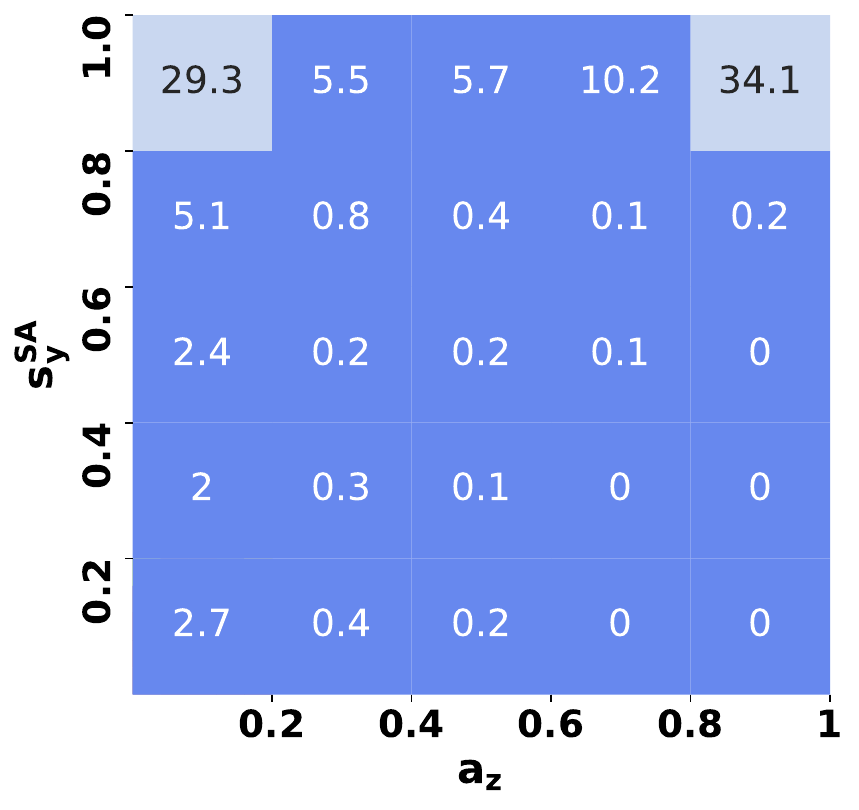}
      \caption{Soft Attention train data }
        \label{fig6:subfig1}
    \end{subfigure}
    \begin{subfigure}[b]{0.32\textwidth}
      \centering
    \includegraphics[width=\textwidth]{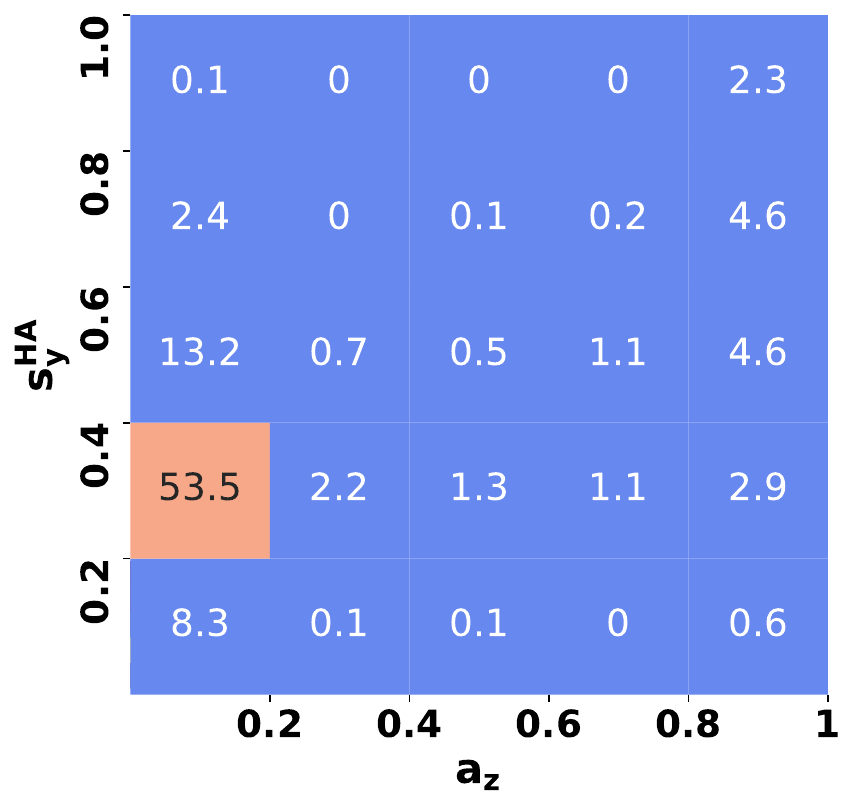}
      \caption{Hard Attention train data}
        \label{fig6:subfig2}
    \end{subfigure}
  \begin{subfigure}[b]{0.32\textwidth}
      \centering
    \includegraphics[width=\textwidth]{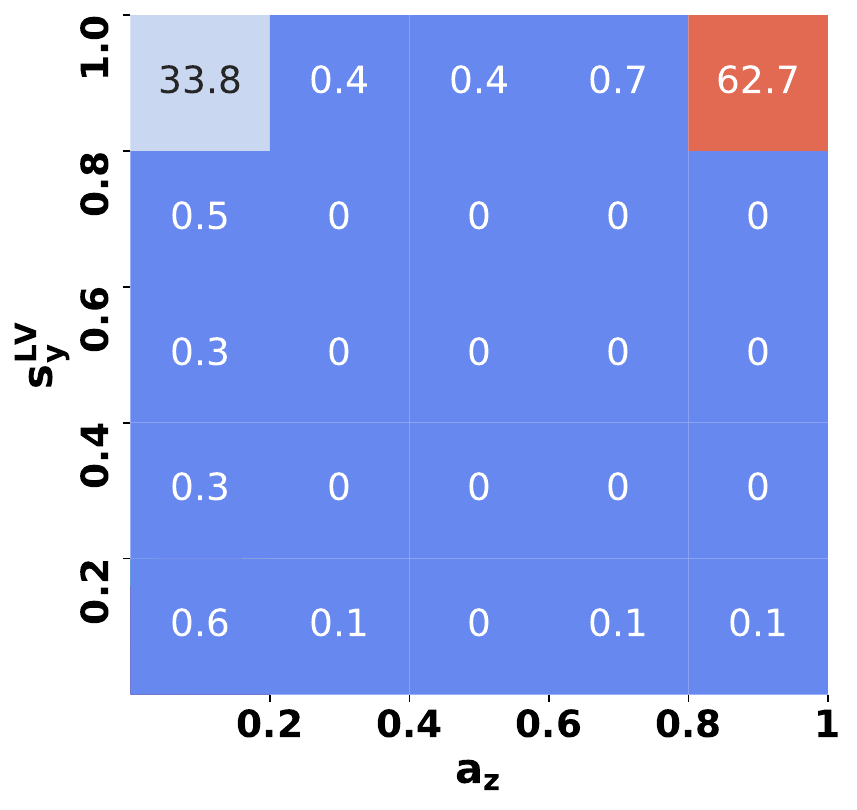}
      \caption{LVML train data}
        \label{fig6:subfig3}
    \end{subfigure}
    \begin{subfigure}[b]{0.32\textwidth}
      \centering
    \includegraphics[width=\textwidth]{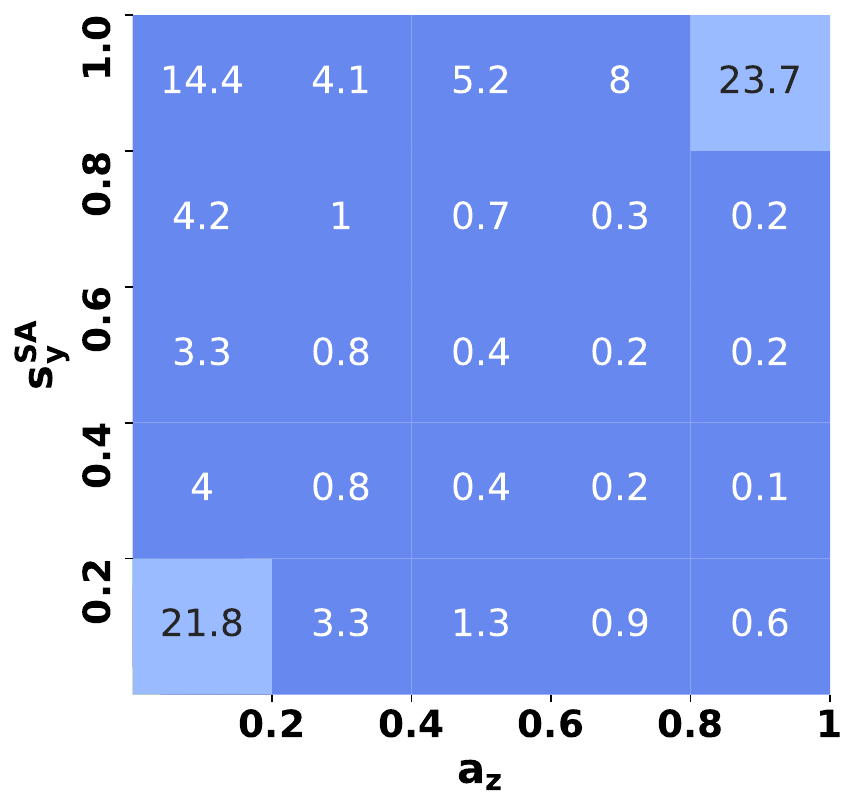}
      \caption{Soft Attention test data}
        \label{fig6:subfig4}
    \end{subfigure}
    \begin{subfigure}[b]{0.32\textwidth}
      \centering
    \includegraphics[width=\textwidth]{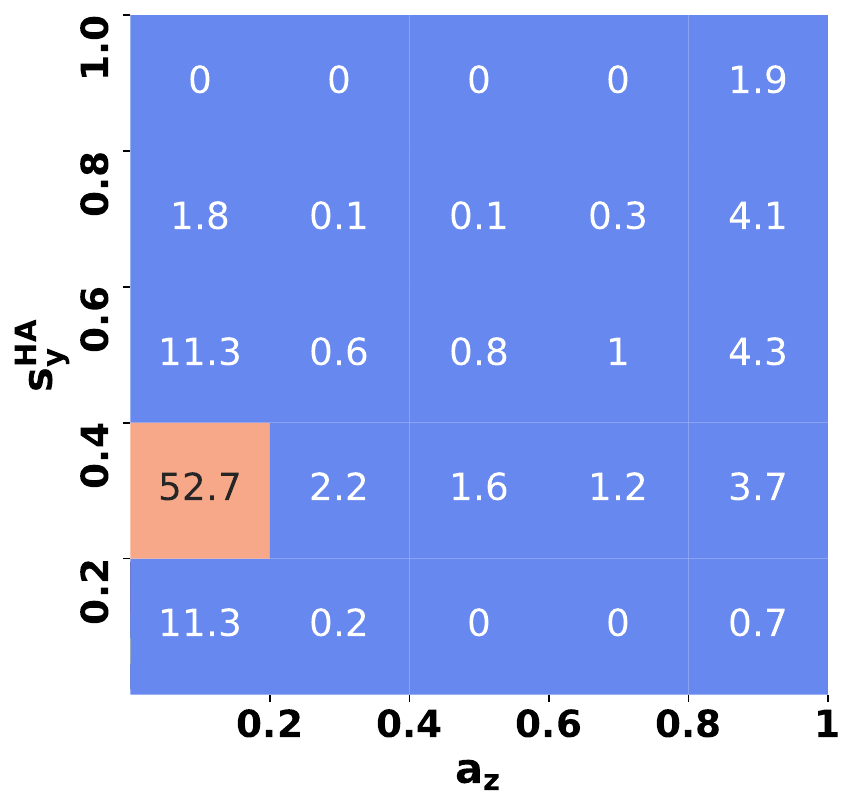}
      \caption{Hard Attention test data}
        \label{fig6:subfig5}
    \end{subfigure}
    \begin{subfigure}[b]{0.32\textwidth}
      \centering
    \includegraphics[width=\textwidth]{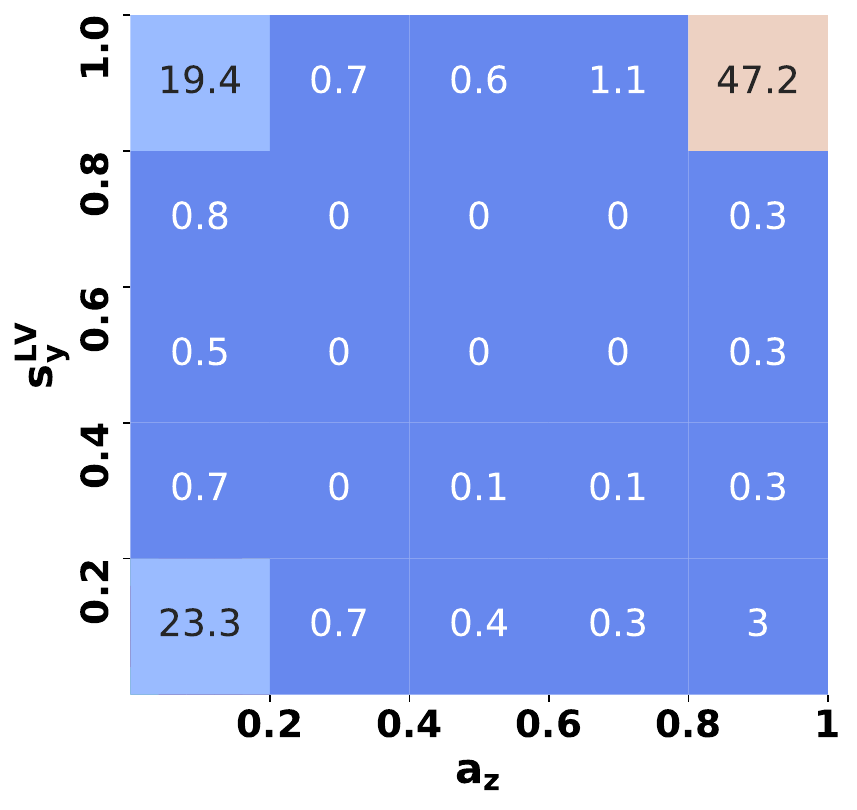}
      \caption{LVML test data}
        \label{fig6:subfig6}
    \end{subfigure}
    \caption{Focus-Prediction heat maps for the three attention paradigms on the CIFAR10 SDC task with $n=10000$ and $m=20$.  The top row contains results on train data and bottom row gives results on test data. }
    \label{fig6:my_label}
\end{figure*}

\begin{figure*}[!ht]
  \centering
    \begin{subfigure}[b]{0.32\textwidth}
      \centering
      \includegraphics[width=\textwidth]{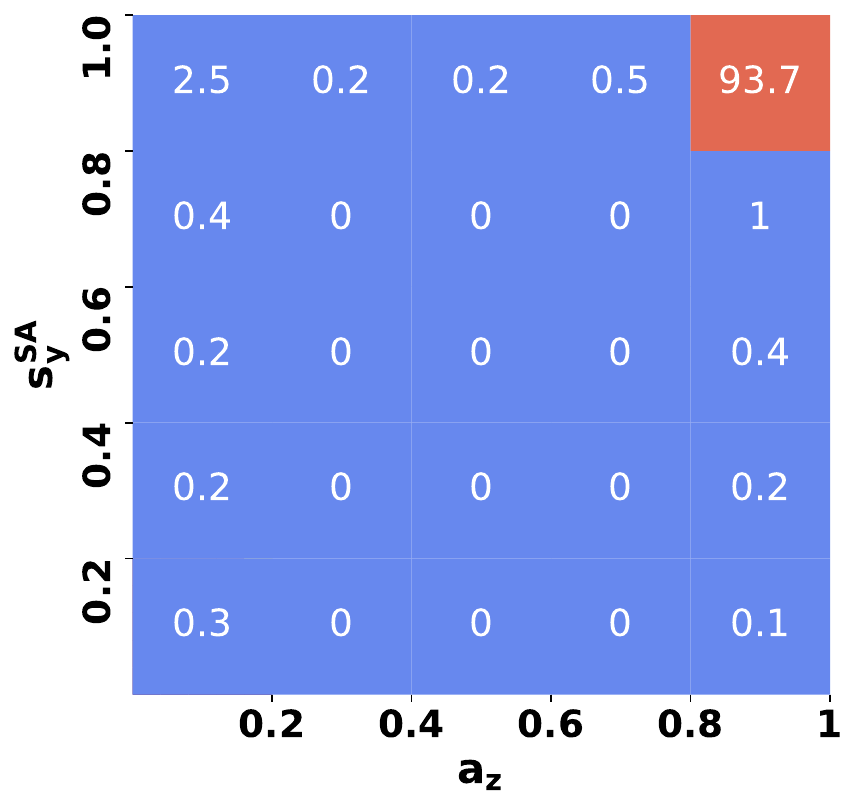}
      \caption{Soft Attention train data }
        \label{fig7:subfig1}
    \end{subfigure}
    \begin{subfigure}[b]{0.32\textwidth}
      \centering
    \includegraphics[width=\textwidth]{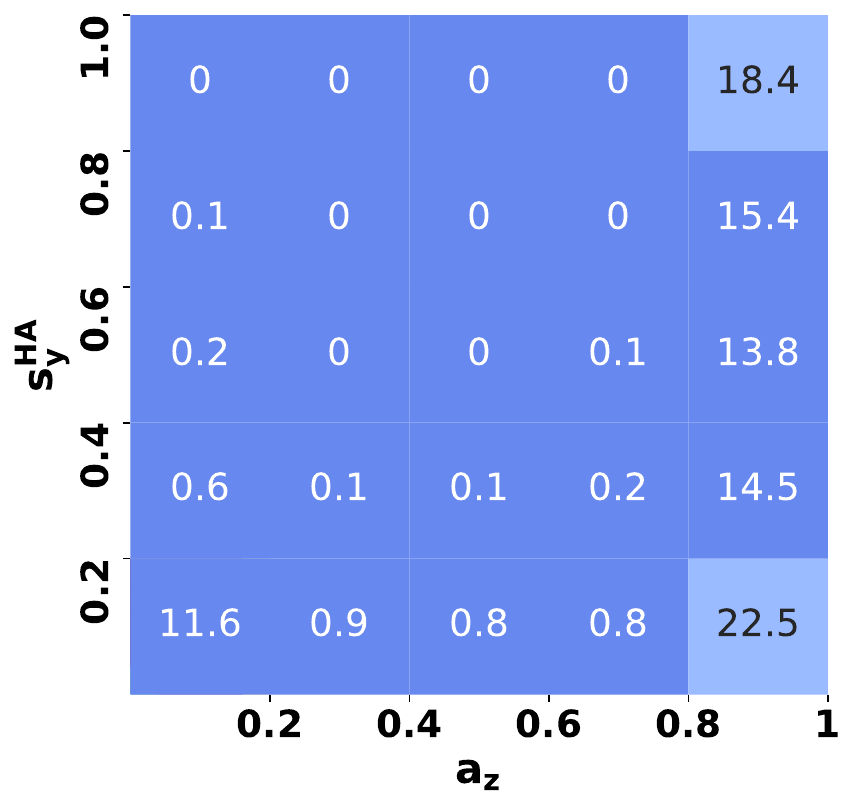}
      \caption{Hard Attention train data}
        \label{fig7:subfig2}
    \end{subfigure}
  \begin{subfigure}[b]{0.32\textwidth}
      \centering
    \includegraphics[width=\textwidth]{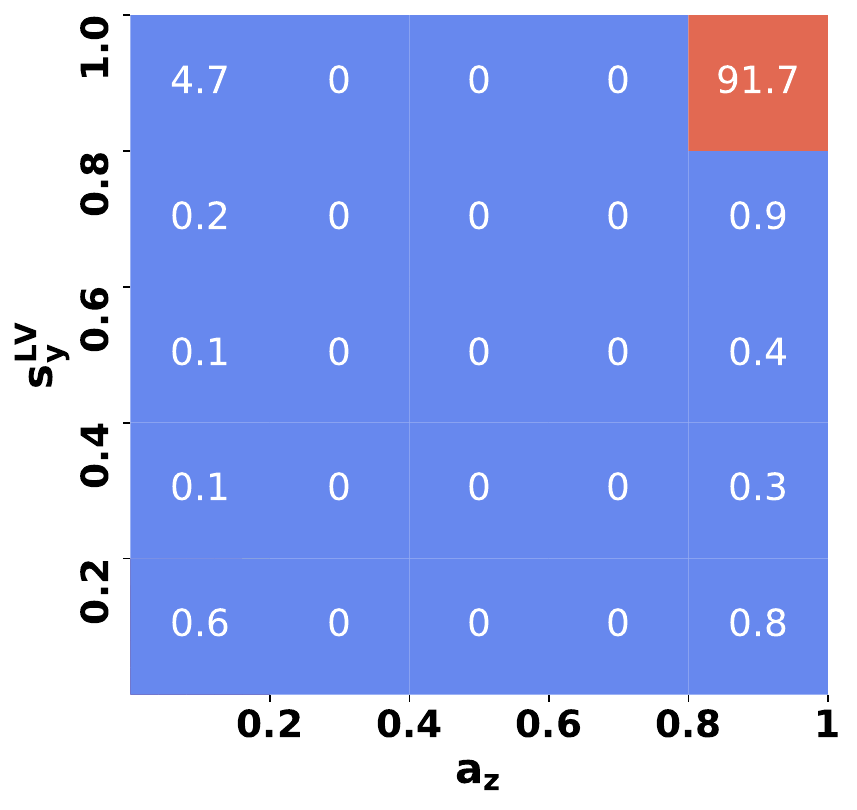}
      \caption{LVML train data}
        \label{fig7:subfig3}
    \end{subfigure}
    \begin{subfigure}[b]{0.32\textwidth}
      \centering
    \includegraphics[width=\textwidth]{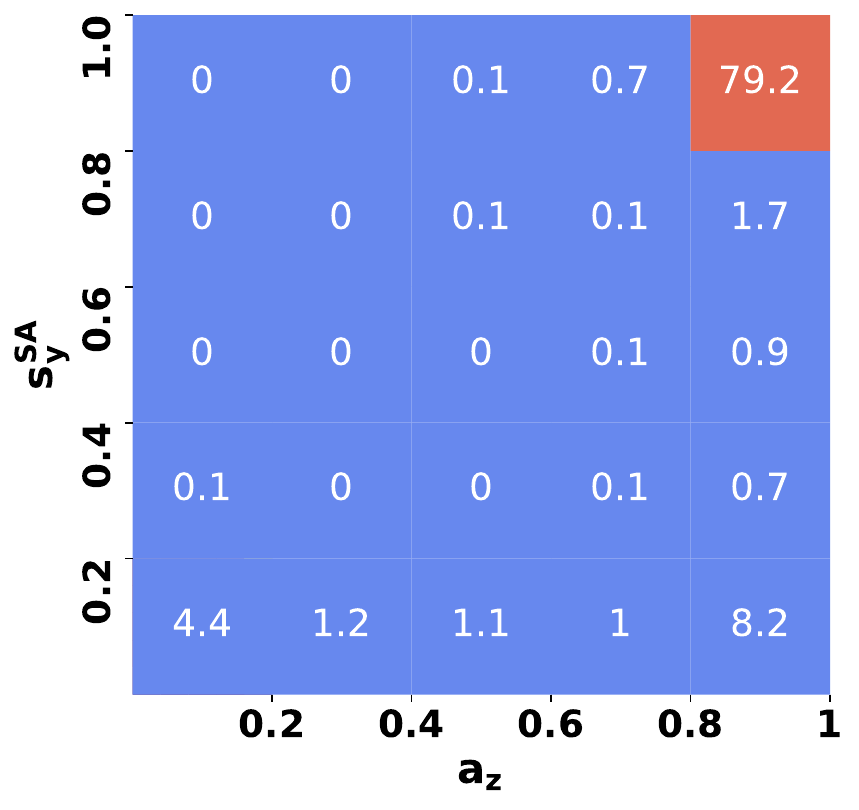}
      \caption{Soft Attention test data}
        \label{fig7:subfig4}
    \end{subfigure}
    \begin{subfigure}[b]{0.32\textwidth}
      \centering
    \includegraphics[width=\textwidth]{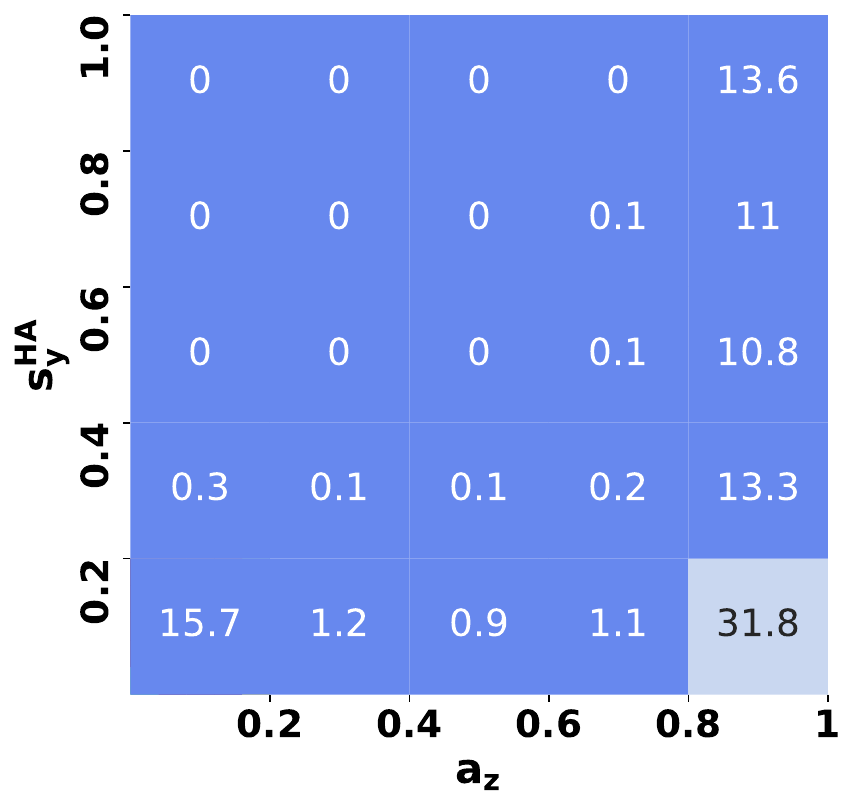}
      \caption{Hard Attention test data}
        \label{fig7:subfig5}
    \end{subfigure}
    \begin{subfigure}[b]{0.32\textwidth}
      \centering
    \includegraphics[width=\textwidth]{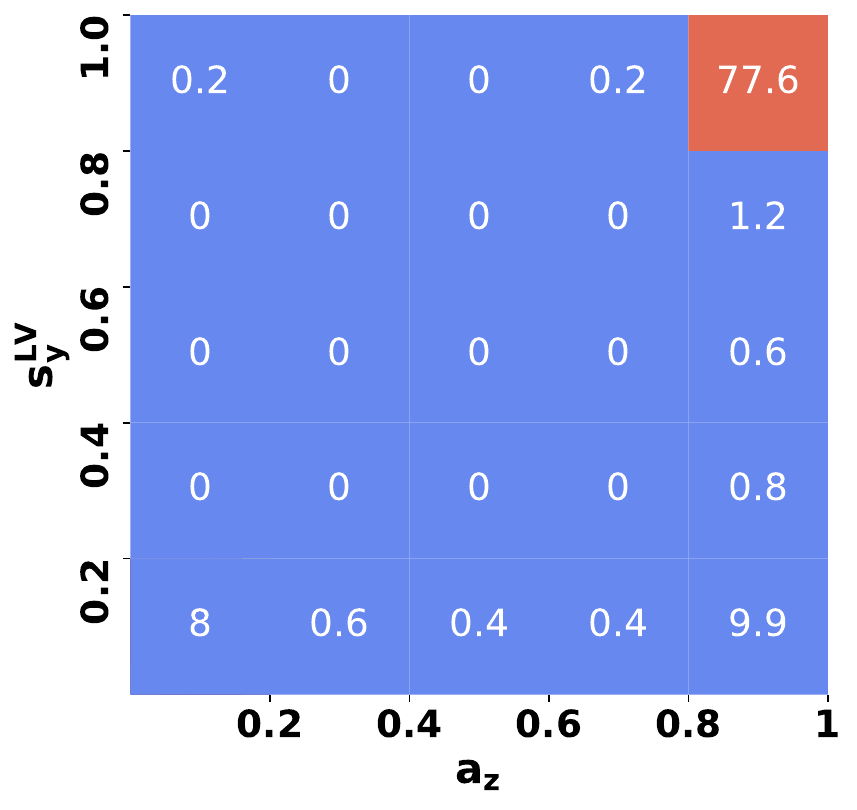}
      \caption{LVML test data}
        \label{fig7:subfig6}
    \end{subfigure}
    \caption{Focus-Prediction heat maps for the three attention paradigms on the CIFAR100 SDC task with $n=50000$ and $m=5$.  The top row contains results on train data and bottom row gives results on test data. . }
    \label{fig7:my_label}
\end{figure*}

\begin{figure*}[!ht]
  \centering
    \begin{subfigure}[b]{0.32\textwidth}
      \centering
      \includegraphics[width=\textwidth]{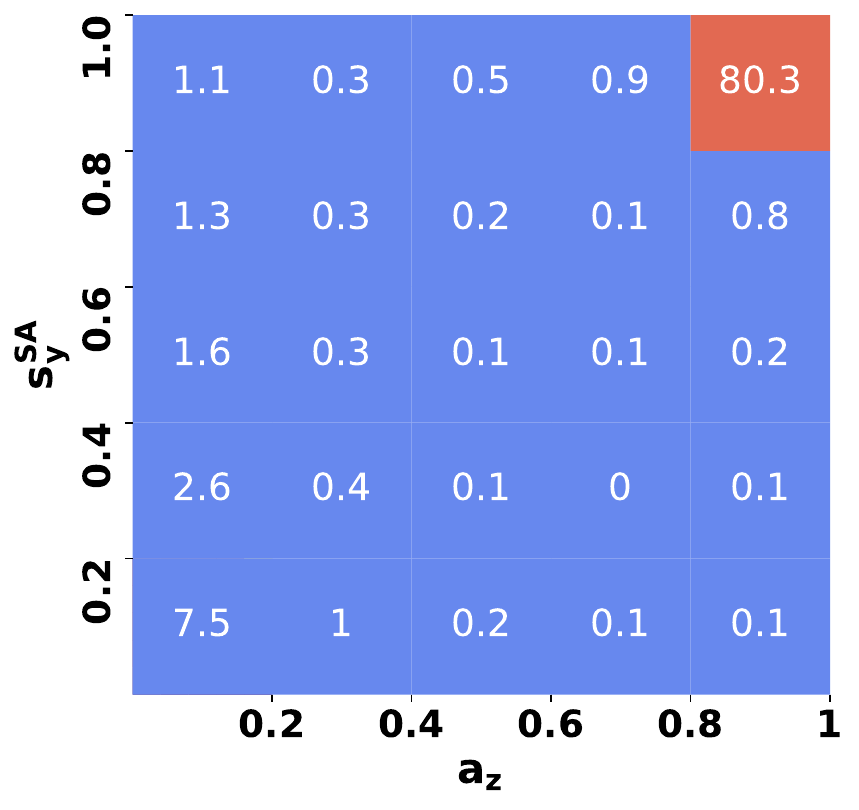}
      \caption{Soft Attention train data }
        \label{fig8:subfig1}
    \end{subfigure}
    \begin{subfigure}[b]{0.32\textwidth}
      \centering
    \includegraphics[width=\textwidth]{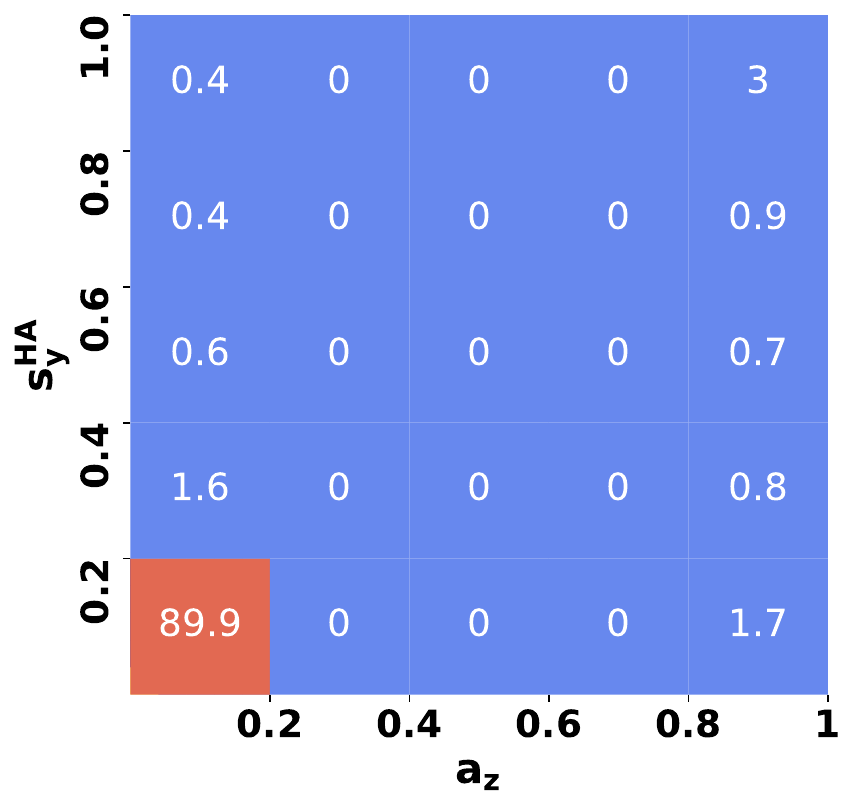}
      \caption{Hard Attention train data}
        \label{fig8:subfig2}
    \end{subfigure}
  \begin{subfigure}[b]{0.32\textwidth}
      \centering
    \includegraphics[width=\textwidth]{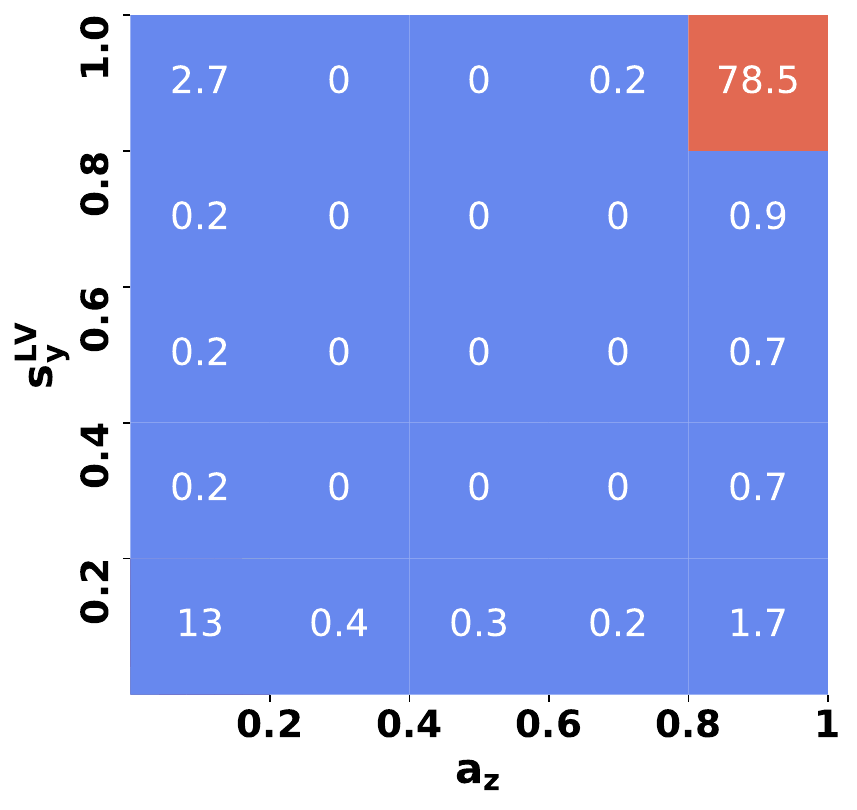}
      \caption{LVML train data}
        \label{fig8:subfig3}
    \end{subfigure}
    \begin{subfigure}[b]{0.32\textwidth}
      \centering
    \includegraphics[width=\textwidth]{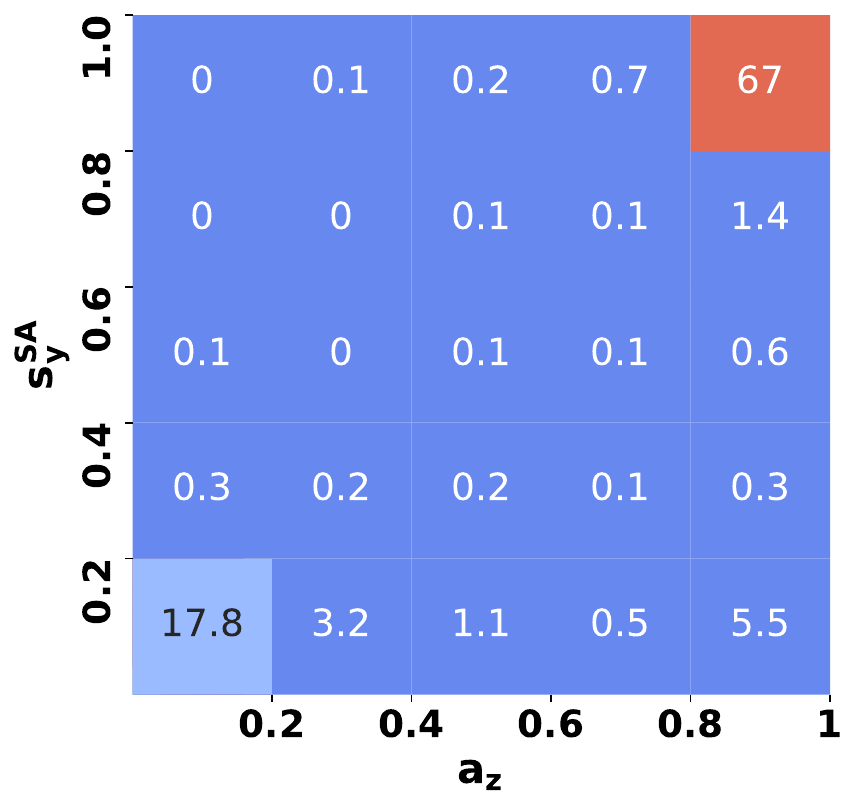}
      \caption{Soft Attention test data}
        \label{fig8:subfig4}
    \end{subfigure}
    \begin{subfigure}[b]{0.32\textwidth}
      \centering
    \includegraphics[width=\textwidth]{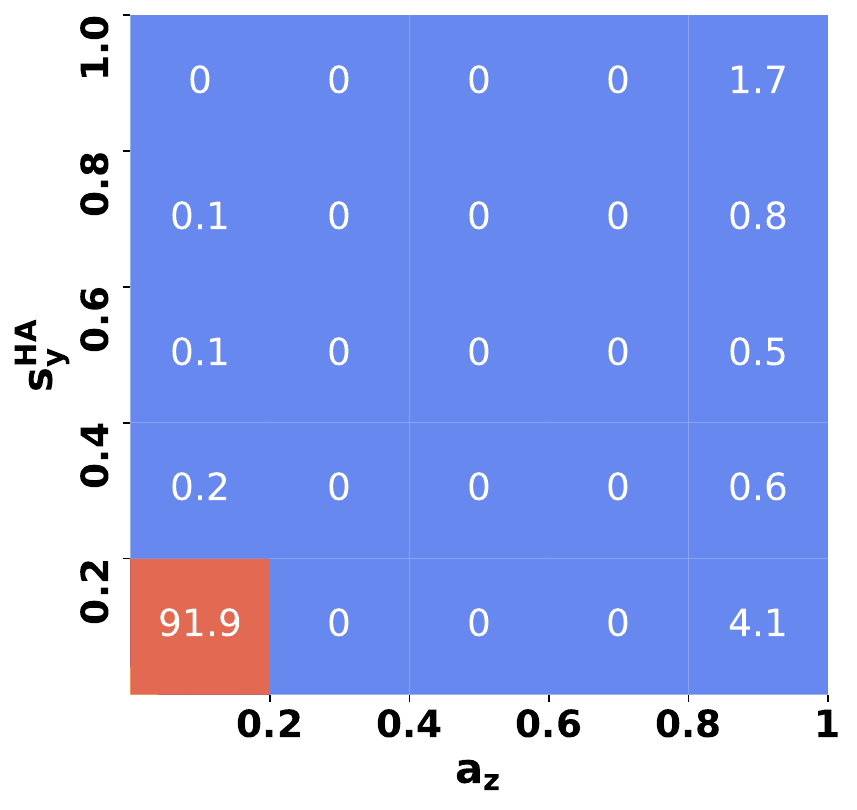}
      \caption{Hard Attention test data}
        \label{fig8:subfig5}
    \end{subfigure}
    \begin{subfigure}[b]{0.32\textwidth}
      \centering
    \includegraphics[width=\textwidth]{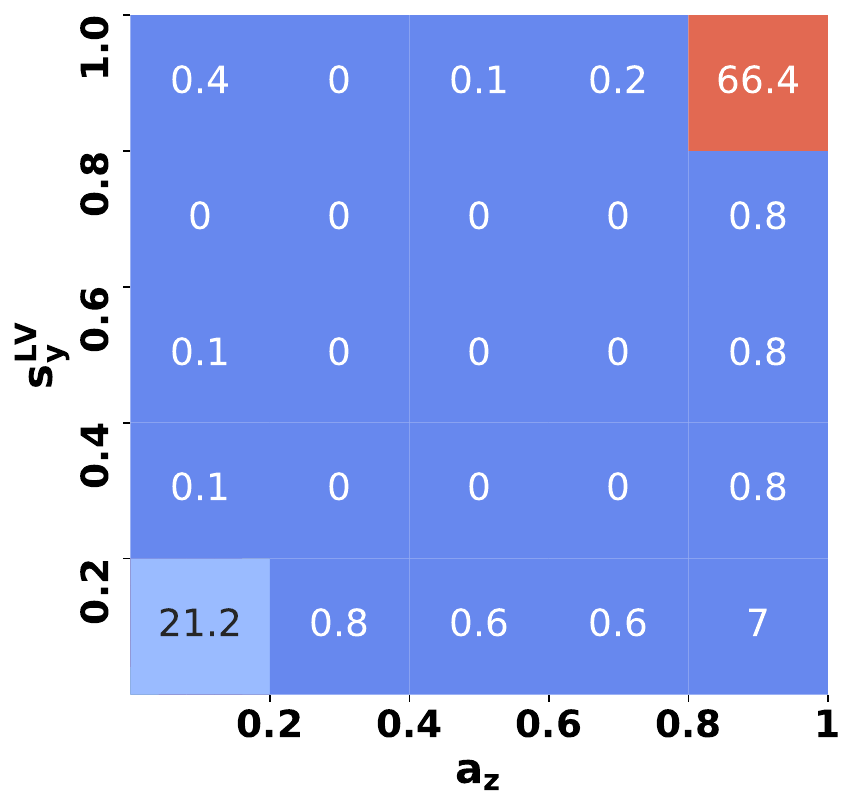}
      \caption{LVML test data}
        \label{fig8:subfig6}
    \end{subfigure}
    \caption{Focus-Prediction heat maps for the three attention paradigms on the CIFAR100 SDC task with $n=50000$ and $m=20$.  The top row contains results on train data and bottom row gives results on test data.  }
    \label{fig8:my_label}
\end{figure*}


\begin{figure*}[!ht]
  \centering
    \begin{subfigure}[b]{0.32\textwidth}
      \centering
      \includegraphics[width=\textwidth]{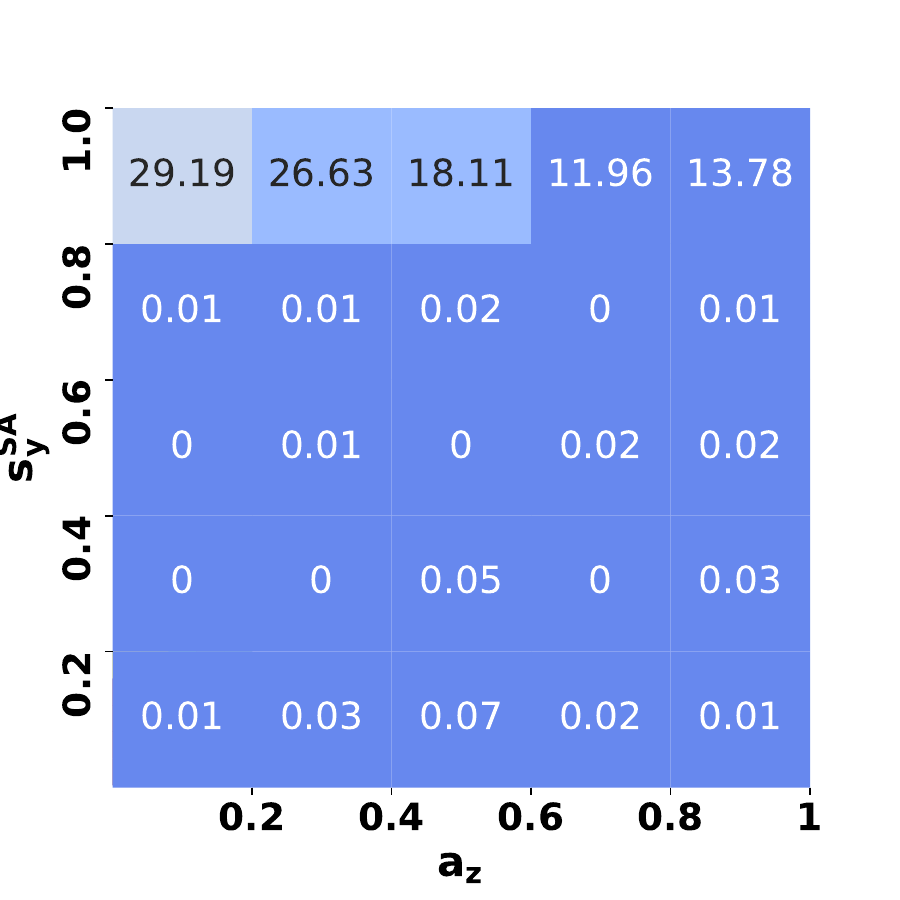}
      \caption{Soft Attention train data }
        \label{fig9:subfig1}
    \end{subfigure}
    \begin{subfigure}[b]{0.32\textwidth}
      \centering
    \includegraphics[width=\textwidth]{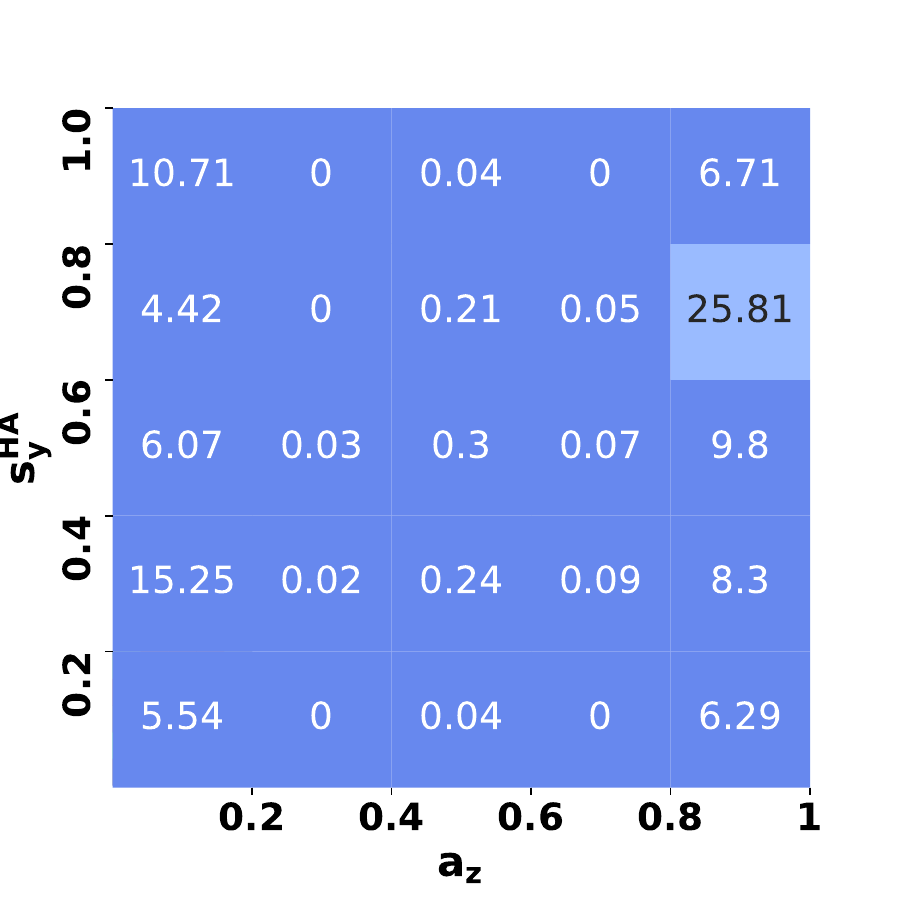}
      \caption{Hard Attention train data}
        \label{fig9:subfig2}
    \end{subfigure}
  \begin{subfigure}[b]{0.32\textwidth}
      \centering
    \includegraphics[width=\textwidth]{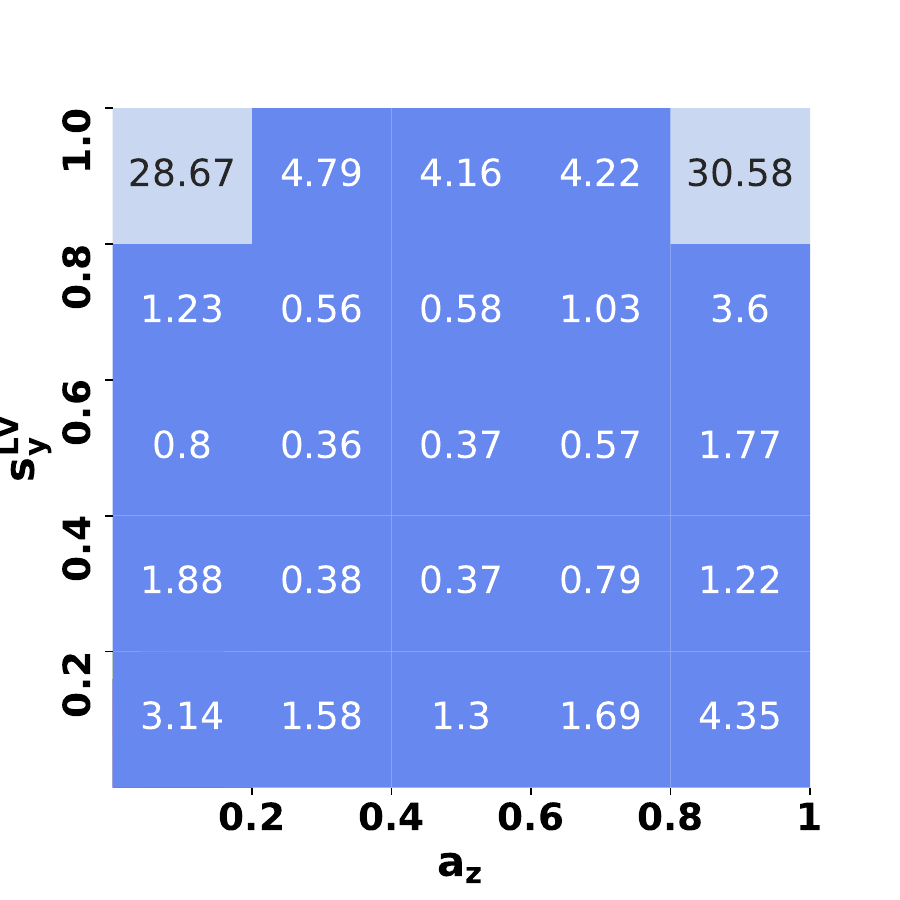}
      \caption{LVML train data}
        \label{fig9:subfig3}
    \end{subfigure}
    \begin{subfigure}[b]{0.32\textwidth}
      \centering
    \includegraphics[width=\textwidth]{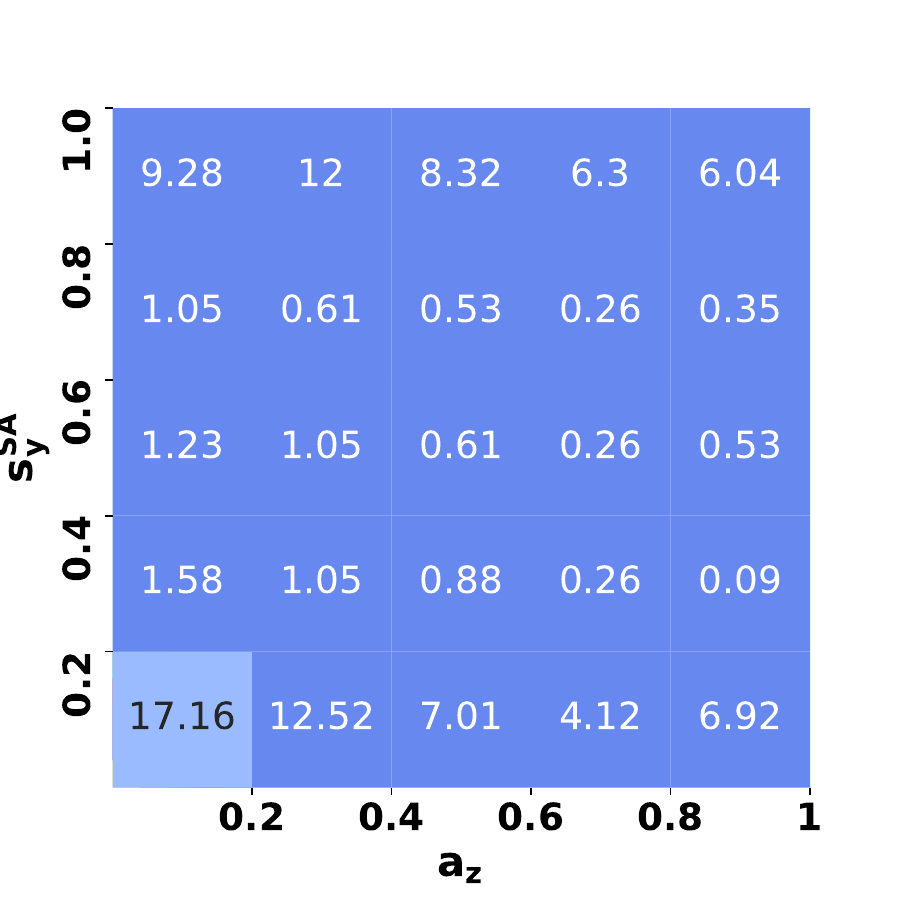}
      \caption{Soft Attention test data}
        \label{fig9:subfig4}
    \end{subfigure}
    \begin{subfigure}[b]{0.32\textwidth}
      \centering
    \includegraphics[width=\textwidth]{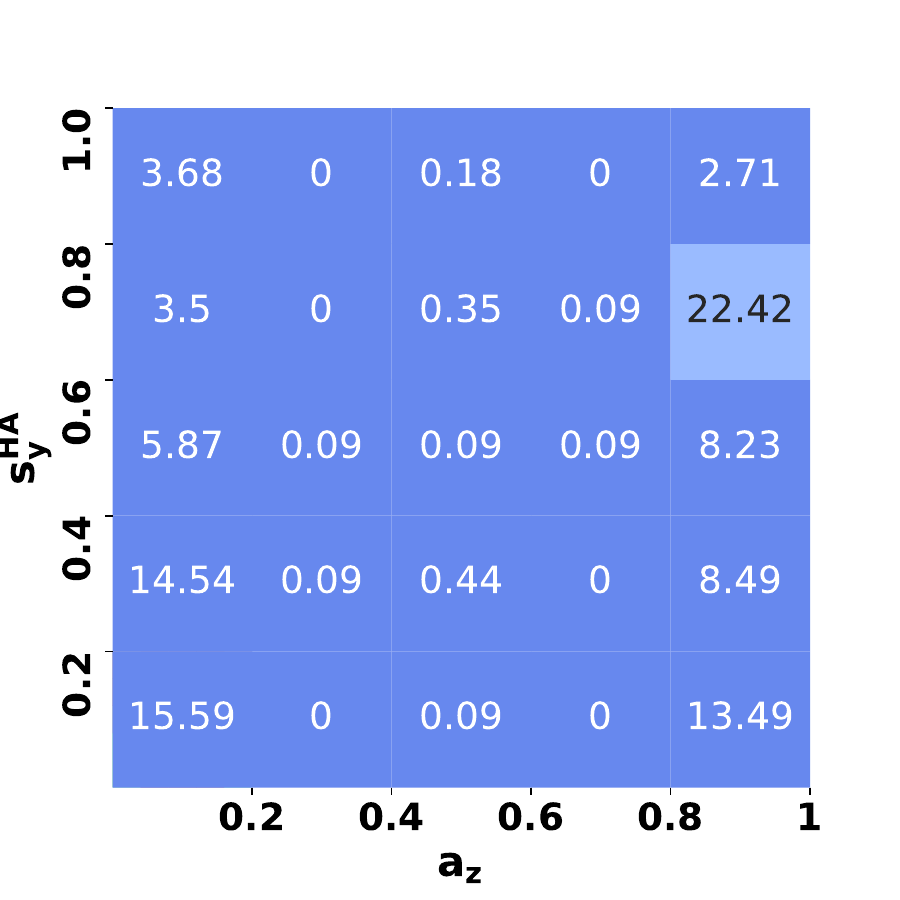}
      \caption{Hard Attention test data}
        \label{fig9:subfig5}
    \end{subfigure}
    \begin{subfigure}[b]{0.32\textwidth}
      \centering
    \includegraphics[width=\textwidth]{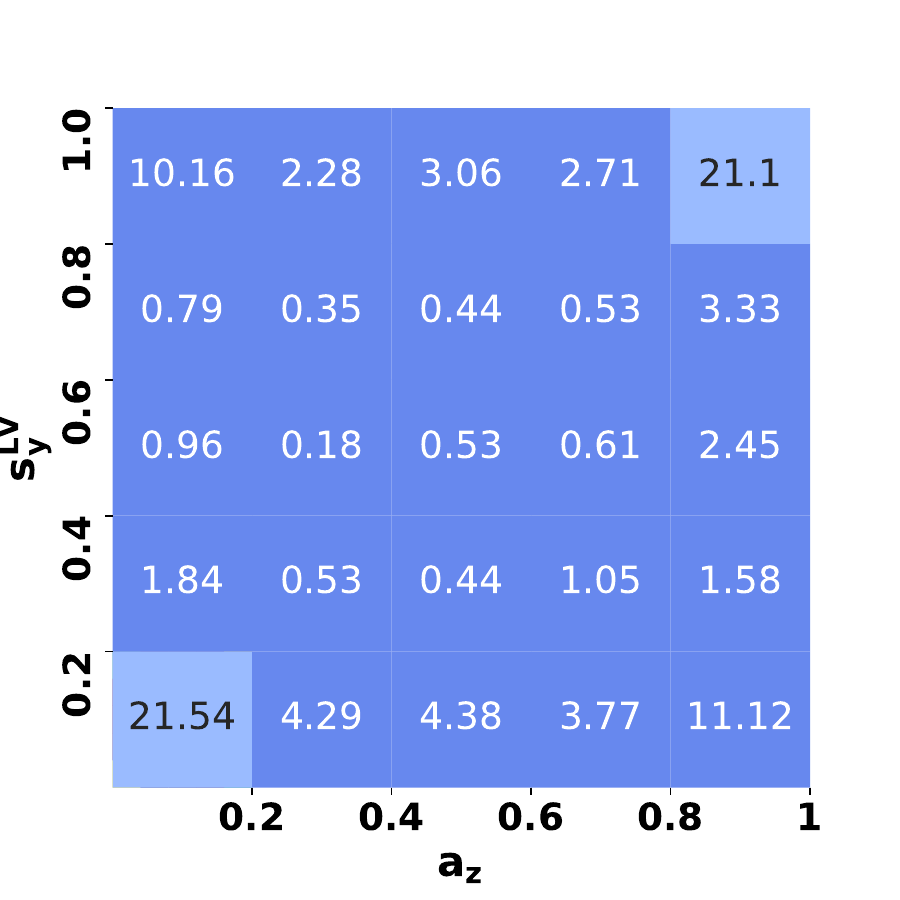}
      \caption{LVML test data}
        \label{fig9:subfig6}
    \end{subfigure}
    \caption{Focus-Prediction heat maps for the three attention paradigms on the HateXplain data.  The top row contains results on train data and bottom row gives results on test data for model.  }
    \label{fig9:my_label}
\end{figure*}

\begin{figure*}[!ht]
  \centering
    \begin{subfigure}[b]{0.32\textwidth}
      \centering
      \includegraphics[width=\textwidth]{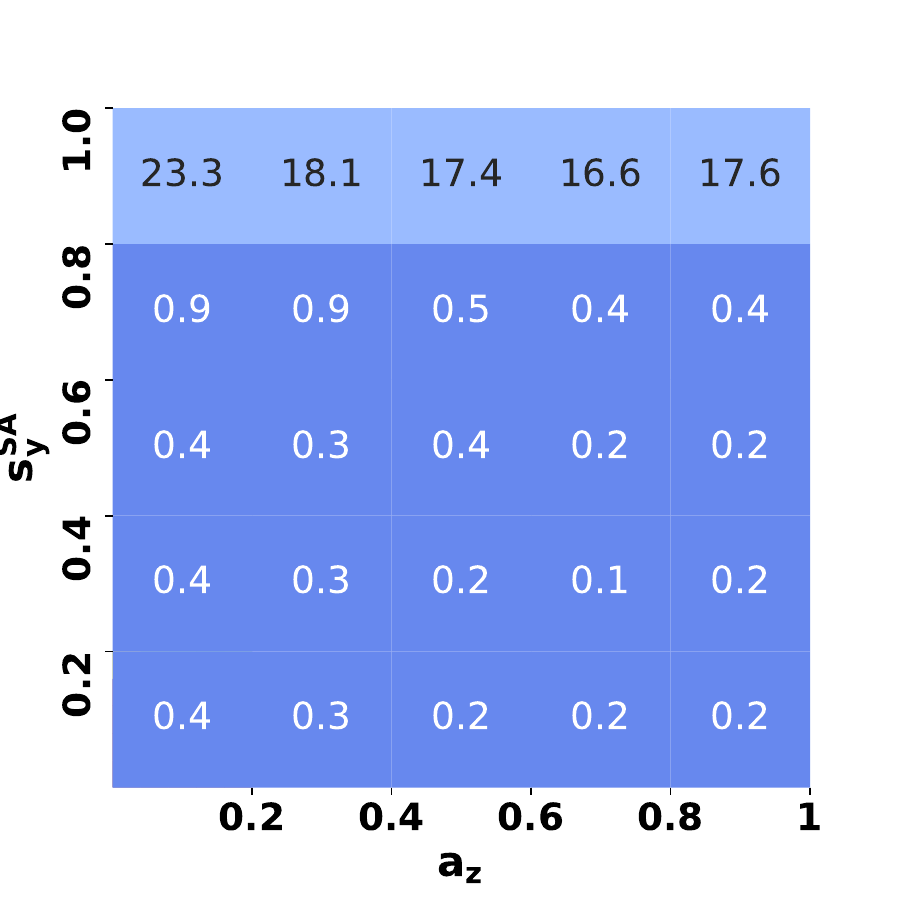}
      \caption{Soft Attention train data }
        \label{fig10:subfig1}
    \end{subfigure}
    \begin{subfigure}[b]{0.32\textwidth}
      \centering
    \includegraphics[width=\textwidth]{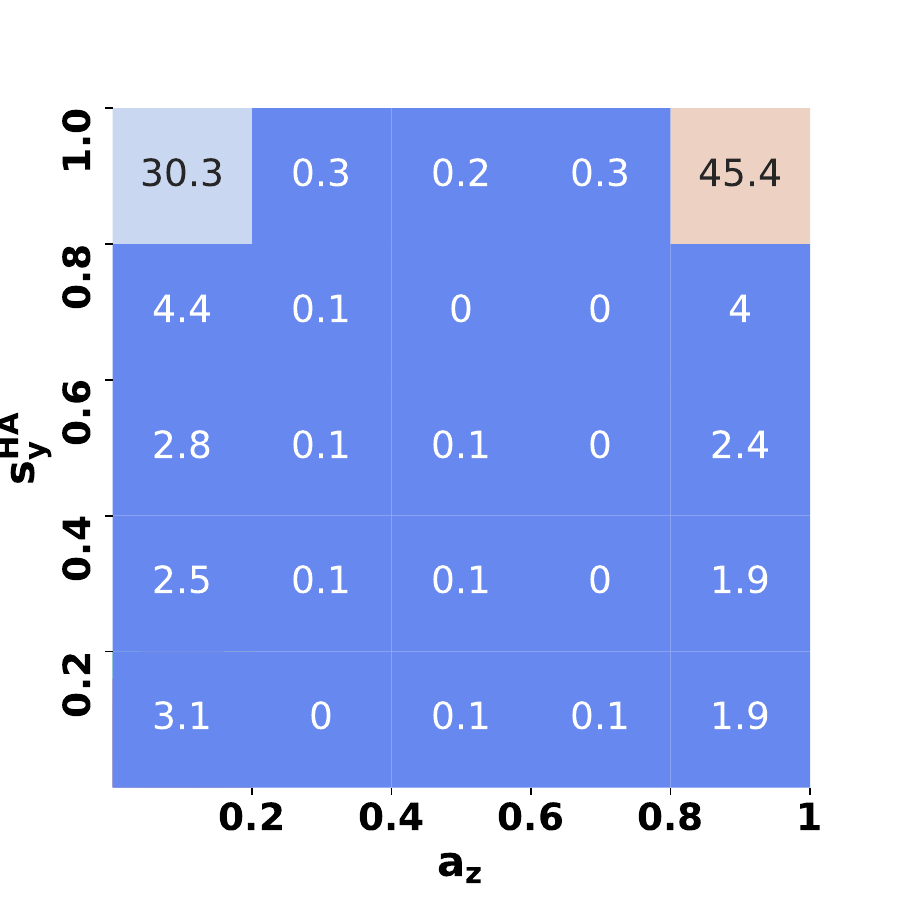}
      \caption{Hard Attention train data}
        \label{fig10:subfig2}
    \end{subfigure}
  \begin{subfigure}[b]{0.32\textwidth}
      \centering
    \includegraphics[width=\textwidth]{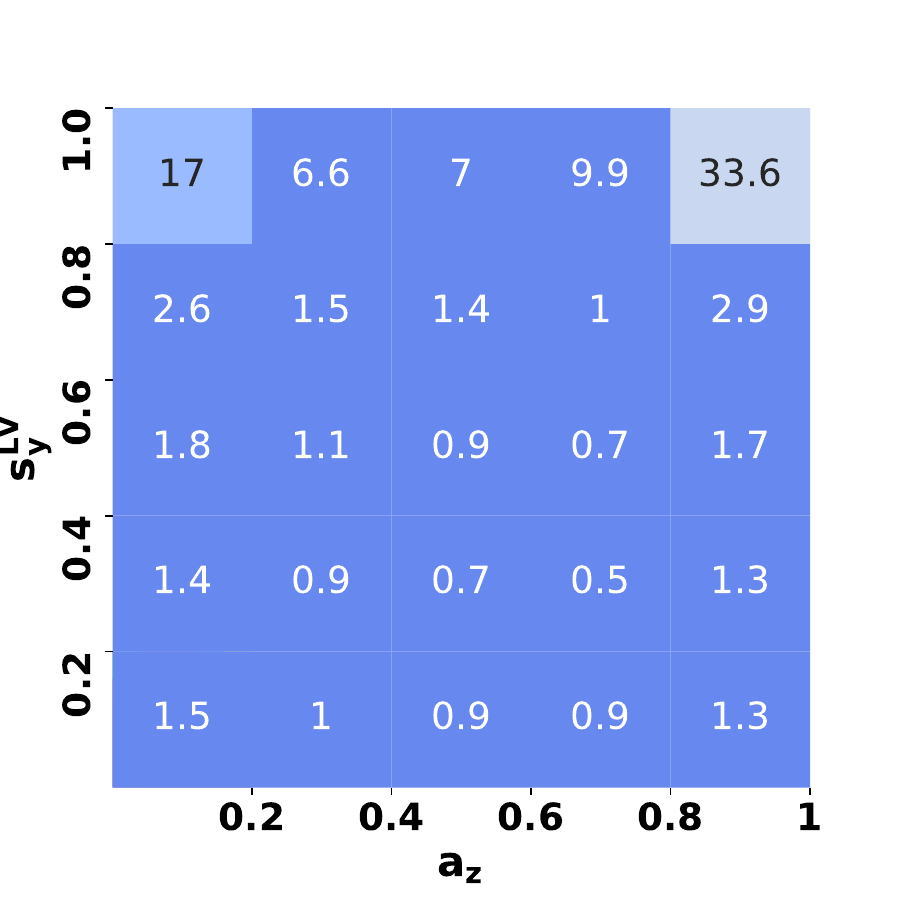}
      \caption{LVML train data}
        \label{fig10:subfig3}
    \end{subfigure}
    \begin{subfigure}[b]{0.32\textwidth}
      \centering
    \includegraphics[width=\textwidth]{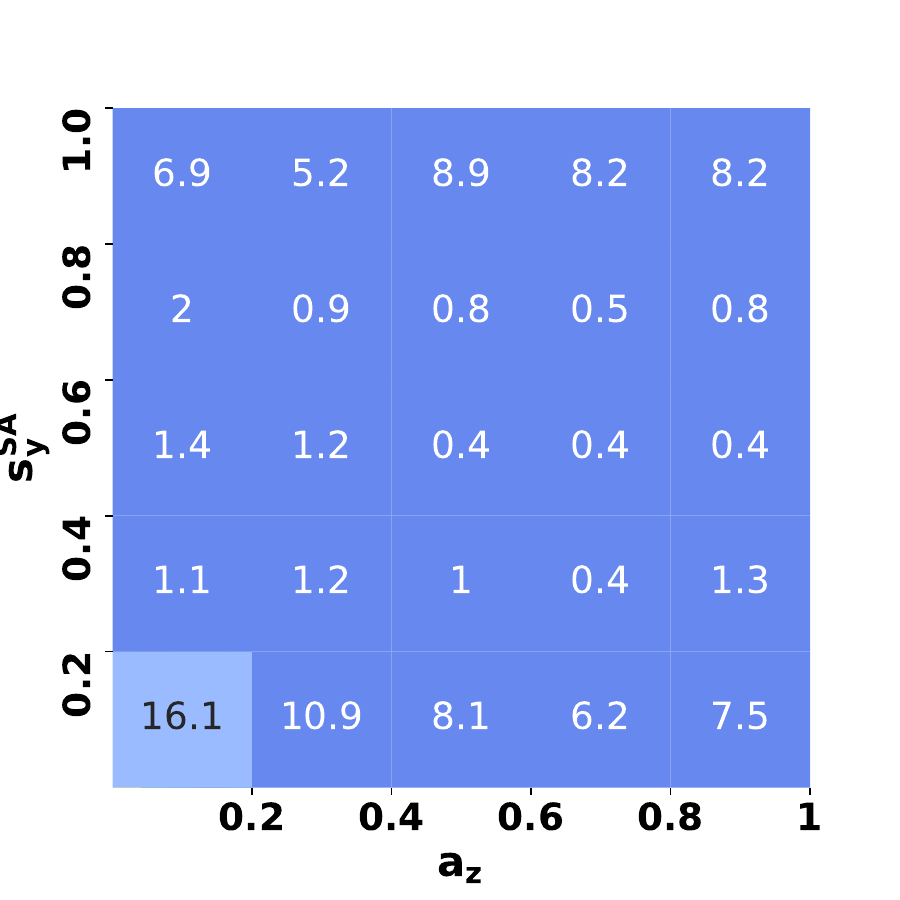}
      \caption{Soft Attention test data}
        \label{fig10:subfig4}
    \end{subfigure}
    \begin{subfigure}[b]{0.32\textwidth}
      \centering
    \includegraphics[width=\textwidth]{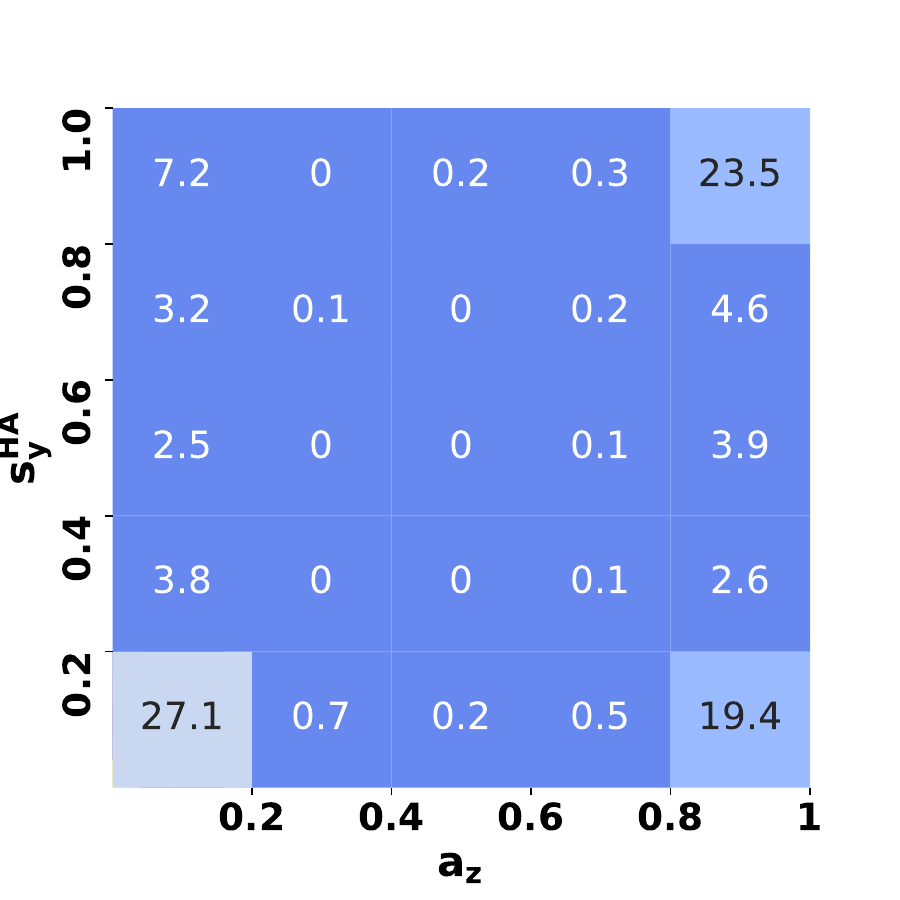}
      \caption{Hard Attention test data}
        \label{fig10:subfig5}
    \end{subfigure}
    \begin{subfigure}[b]{0.32\textwidth}
      \centering
    \includegraphics[width=\textwidth]{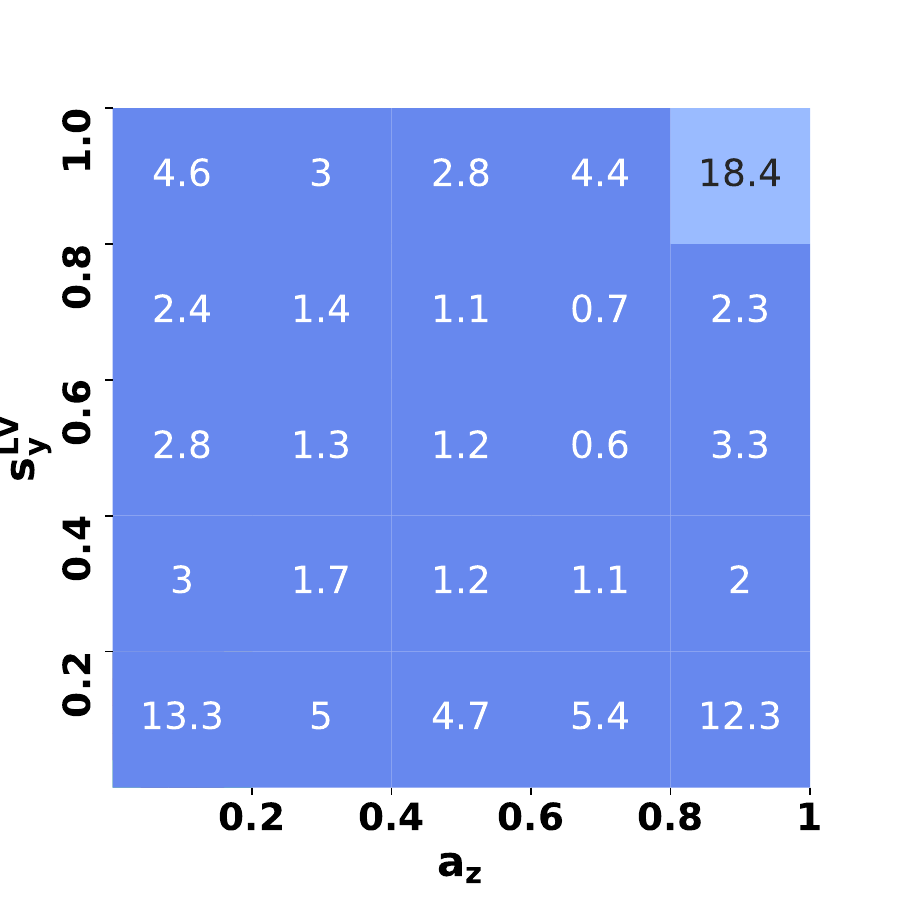}
      \caption{LVML test data}
        \label{fig10:subfig6}
    \end{subfigure}
    \caption{Focus-Prediction heat maps for the three attention paradigms on the HateXplain data.  The top row contains results on train data and bottom row gives results on test data for model with self-attention.  }
    \label{fig10:my_label}
\end{figure*}


\begin{figure*}[!ht]
  \centering
      \begin{subfigure}[b]{0.49\textwidth}
      \centering
      \includegraphics[width=\textwidth]{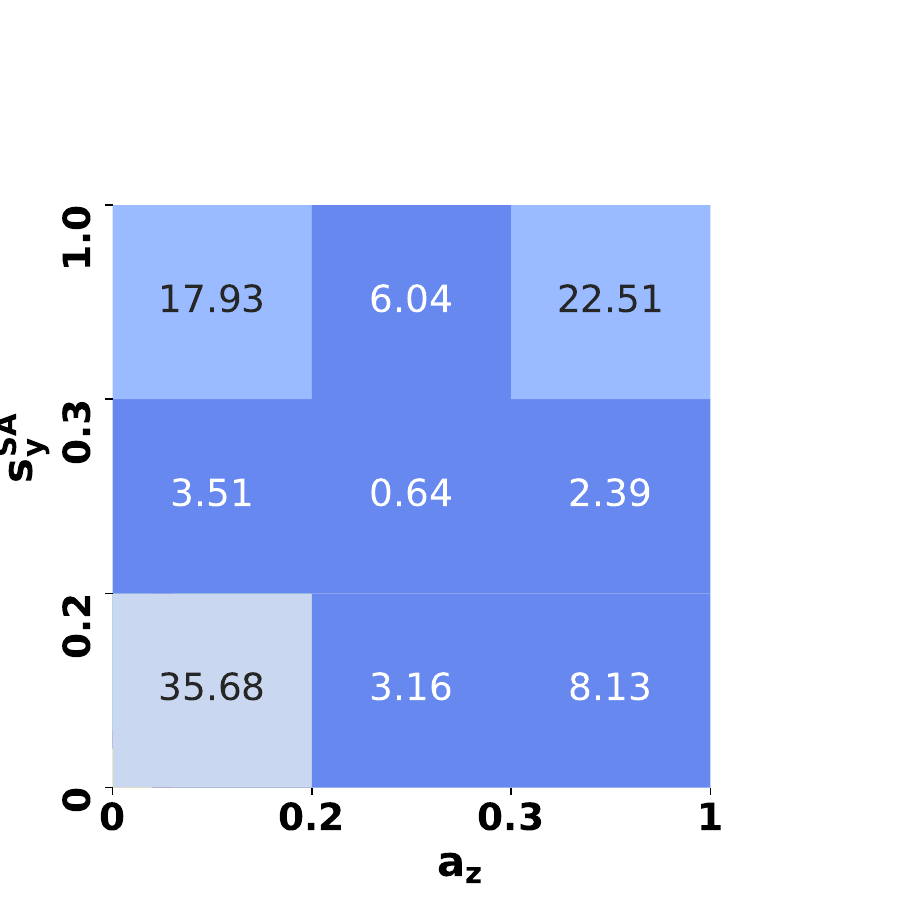}
      \caption{Soft Attention }
        \label{fig11:subfig1}
    \end{subfigure}
    \begin{subfigure}[b]{0.49\textwidth}
      \centering
    \includegraphics[width=\textwidth]{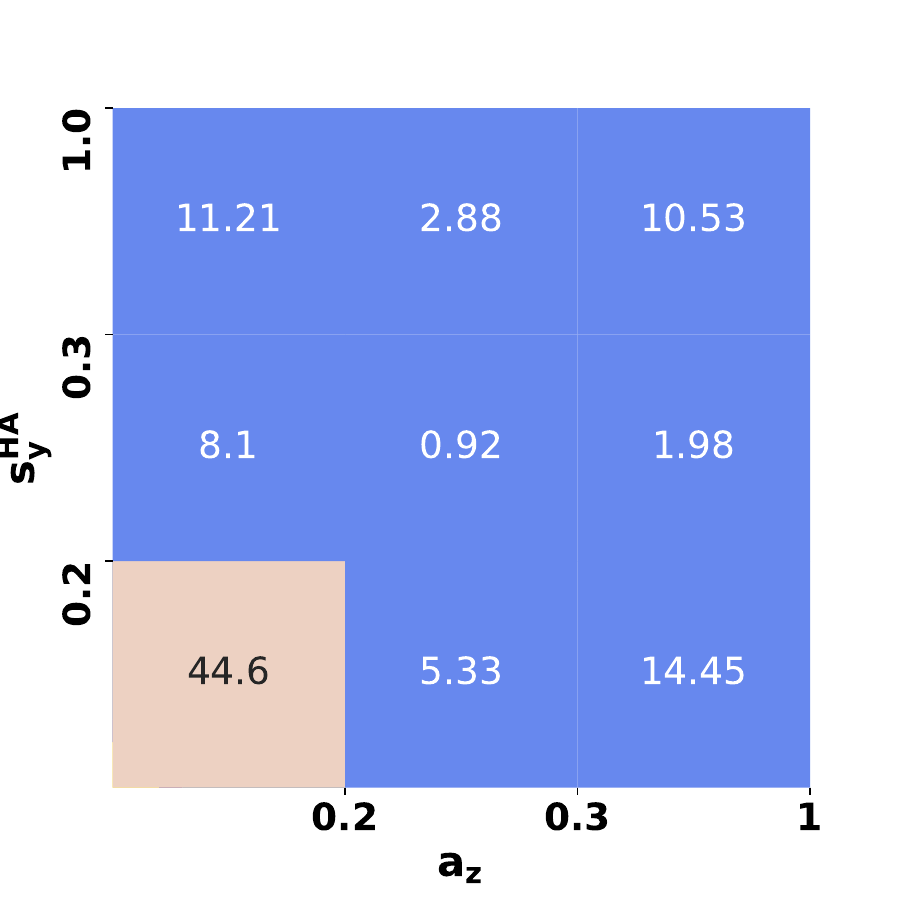}
      \caption{Hard Attention }
        \label{fig11:subfig2}
    \end{subfigure}
    \begin{subfigure}[b]{0.49\textwidth}
      \centering
      \includegraphics[width=\textwidth]{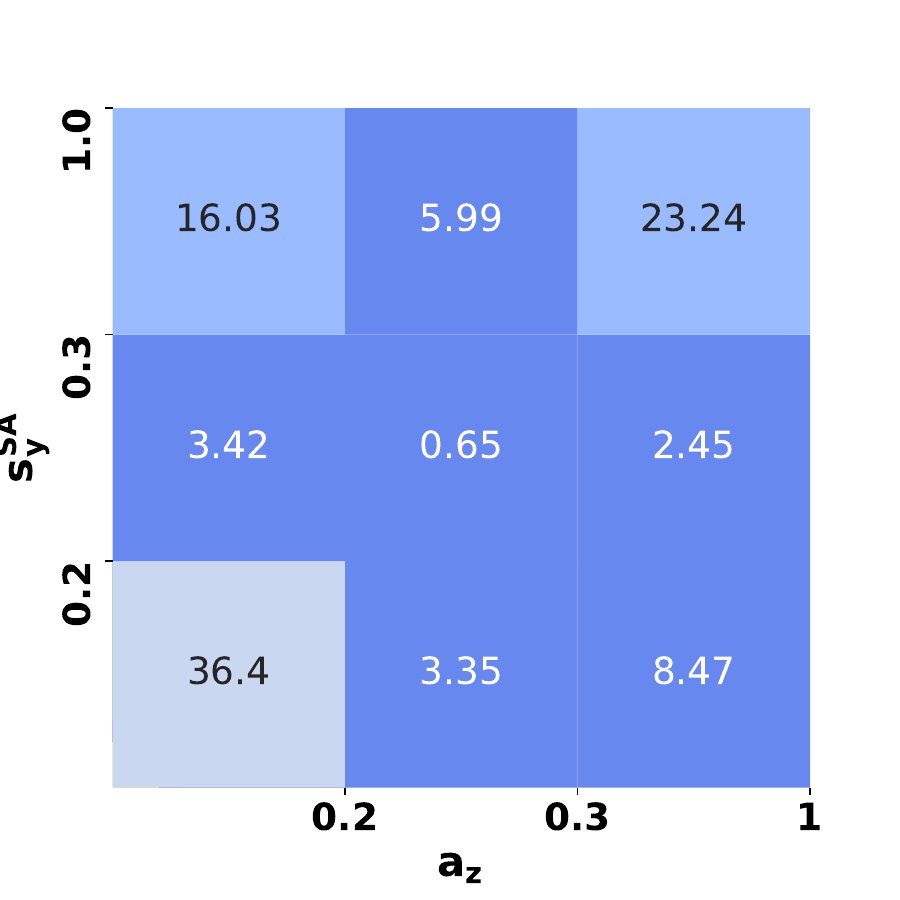}
      \caption{Soft Attention }
        \label{fig11:subfig3}
    \end{subfigure}
    \begin{subfigure}[b]{0.49\textwidth}
      \centering
    \includegraphics[width=\textwidth]{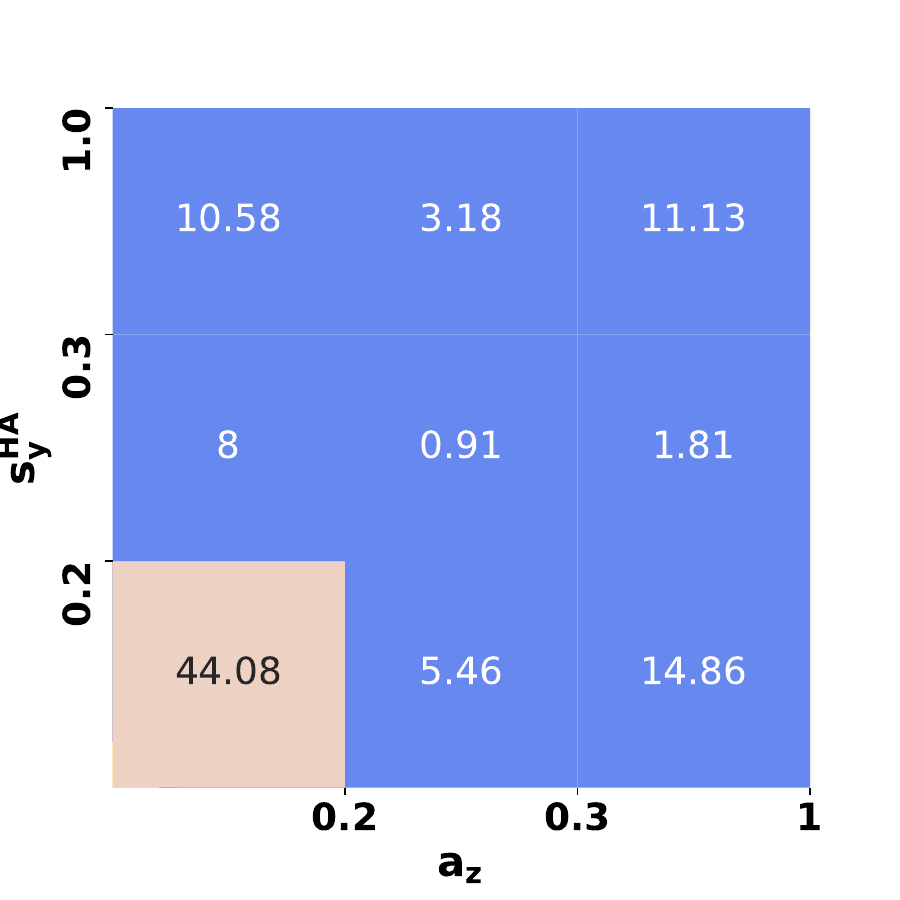}
      \caption{Hard Attention }
        \label{fig11:subfig4}
    \end{subfigure}

    \caption{Focus-Prediction heat maps for the attention paradigms on the MSCOCO dataset for train and validation data. Top row is for train data and bottom row is for validation data. }
    \label{fig11:my_label}
\end{figure*}

\end{document}